\documentclass[journal]{IEEEtran}
\ifCLASSINFOpdf
\else
\fi

\usepackage{microtype}
\usepackage{graphicx}
\usepackage{subfigure}
\usepackage{booktabs}
\usepackage{hyperref}
\usepackage{bm}
\usepackage{amsmath,amssymb}
\usepackage{nccmath}
\usepackage{subfigure}
\usepackage{xcolor}
\DeclareMathOperator{\E}{\mathbb{E}}
\makeatletter
\newcommand*\bigcdot{\mathpalette\bigcdot@{.5}}
\newcommand*\bigcdot@[2]{\mathbin{\vcenter{\hbox{\scalebox{#2}{$\m@th#1\bullet$}}}}}
\makeatother

\begin{document}
%
\title{Skill Transfer in Deep Reinforcement Learning under Morphological Heterogeneity}
%
%
%

\author{Yang~Hu
        and~Giovanni~Montana
\thanks{The authors are with Warwick Manufacturing Group (WMG), University of Warwick, United Kindom. E-mail: \{yang.hu.1, g.montana\}@warwick.ac.uk.}
\thanks{\textcircled{c} 2019 IEEE. Personal use of this material is permitted. Permission from IEEE must be obtained for all other uses, in any current or future media, including reprinting/republishing this material for advertising or promotional purposes, creating new collective works, for resale or redistribution to servers or lists, or reuse of any copyrighted component of this work in other works.}
}

\maketitle

\begin{abstract}
Transfer learning methods for reinforcement learning (RL) domains facilitate the acquisition of new skills using previously acquired knowledge. The vast majority of existing approaches assume that the agents have the same design, e.g. same shape and action spaces. In this paper we address the problem of transferring previously acquired skills amongst morphologically different agents (MDAs). For instance, assuming that a bipedal agent has been trained to move forward, could this skill be transferred on to a one-leg hopper so as to make its training process for the same task more sample efficient? We frame this problem as one of subspace learning whereby we aim to infer latent factors representing the control mechanism that is common between MDAs. We propose a novel paired variational encoder-decoder model, PVED, that disentangles the control of MDAs into shared and agent-specific factors. The shared factors are then leveraged for skill transfer using RL. Theoretically, we derive a theorem indicating how the performance of PVED depends on the shared factors and agent morphologies. Experimentally, PVED has been extensively validated on four MuJoCo environments. We demonstrate its performance compared to a state-of-the-art approach and several ablation cases, visualize and interpret the hidden factors, and identify avenues for future improvements.
\end{abstract}


%
\IEEEpeerreviewmaketitle

\section{Introduction}

Reinforcement learning (RL)~\cite{RLIntroBook2nd_Sutton_2018} with deep neural network-based policies, known as deep reinforcement learning (DRL), 
has recently attained impressive performance on a wide range of tasks such as
Atari games~\cite{DRLNature_Mnih_2015}, Go~\cite{DRLGo_Silver_2016} and robot control~\cite{DDPG_Lillicrap_2015, DeepVisuomotorPolicies_Levine_2016}. To improve the sample efficiency of DRL, significant research has focused on transfer learning, 
exploiting the previously acquired knowledge to 
efficiently learn new skills~\cite{TransferLearningSurvey_Taylor_2009}. Existing studies have investigated several skill transfer methodologies including learning a task from previous tasks~\cite{ProgressiveNet_Rusu_2016, TaskTransfer_Pinto_2017, TaskTransfer_Barreto_2017, TaskTransfer_Oh_2017, TaskTransfer_Hu_2018, PolicyArchitectureTransfer_Czarnecki_2018, TaskTransfer_Tirinzoni_2019}  and transferring tasks between environments with different dynamics~\cite{DynamicTransfer_Rajeswaran_2017, SimToReal_Sadeghi_2017, SimToReal_Peng_2018, DynamicTransfer_Yu_2019}.

Many of these existing studies typically assume that the learning agents share the same shape and operational functionalities; for instance, a robotic arm with fixed links and joints is trained on different tasks. In this work we consider a setting whereby an agent (e.g. a robot) has initially been trained on a particular task through RL, and we would like a second, but morphologically different agent (MDA), to acquire the same skills by leveraging the existing knowledge. This problem, which we refer to as {\it learning under morphological heterogeneity} (LMH), has emerged in recent years~\cite{ManifoldAlignment_Bocsi_2013, ManifoldAlignment_Ammar_2015, LearningProjection_Raimalwala_2016, ModularDRLTrasfer_Devin_2017, InvariantSubspaceLearning_Gupta_2017, HardwareEncoding_Chen_2018, GNNPolicy_Wang_2018, TransferLearningApprentice_Joshi_2018, HierarchicalRL_Tirumala_2019}. One application domain where this could be advantageous is robotics where different manufacturers produce robots with different designs and specifications for the same task~\cite{SurveyRobotManufacturing_Chu_2010,SurveyRobotManufacturing_Smith_2012}. Enabling skill transfer between MDAs would introduce significant sample efficiencies thus reducing the time required to train different robots. Another practical benefit would be the quick adaption of policies when parts of the robot are damaged in field work. Despite these initial efforts, LMH in the context of DRL is still deemed an open problem and existing methods present a number of limitations. For instance, a significant amount of experience is often required from multiple agents and/or multiple tasks to enable skill transfer ~\cite{HardwareEncoding_Chen_2018, ModularDRLTrasfer_Devin_2017, InvariantSubspaceLearning_Gupta_2017}. Agent-related and agent-agnostic states need to be manually defined to learn hierarchical control~\cite{HierarchicalRL_Tirumala_2019}. Moreover, most approaches involving MDAs consider similar morphologies and operational functionalities, for instance locomotion agents with similarly shaped legs ~\cite{GNNPolicy_Wang_2018}. The problem of transferring skills using MDAs with larger differences is more complex and has not yet been fully explored. For instance, learning to {\it move forward} would require different control strategies for a one-leg and a bipedal agents despite the outcome being the same.

In this paper we frame the LMH problem as one of subspace learning~\cite{InvariantSubspaceLearning_Gupta_2017}. Our aim is to leverage existing knowledge about the observed control mechanisms in two different MDAs in order to infer shared latent factors enabling skill transfer. The intuition we develop here is that there must exist both {\it shared} (i.e. morphology-invariant) factors providing an abstract representation of the common control mechanism as well as {\it individual} (i.e. morphology-dependent) factors capturing the detailed controls specific to each agent. For example, when a bipedal robot and a one-leg hopping robot learn how to move forward, the shared factors would capture the mechanisms controlling the stages of motion like initial contact, balance keeping and swinging, whereas the individual factors would capture agent-specific actions like walking and hopping. Being able to disentangle shared and individual factors then yields a simple strategy for skill transfer as the shared factors can be leveraged within a deep reinforcement learning algorithm. 

The technical contributions of this work are two-fold. First, we propose a novel paired variational encoder-decoder (PVED) that jointly models shared and individual factors within a unified framework. PVED uses an agent's states and actions as a joint representation of the control mechanism, and provides a fully probabilistic model for subspace learning to support skill transfer. Second, we perform a theoretical analysis for the proposed PVED model. We derive a theorem indicating how the performance of PVED depends on the shared factors and agent morphologies. This also provides a performance guarantee for the proposed method. Empirically, using four environments consisting of robot arms and locomotion robots with various morphologies, we demonstrate that PVED can handle MDAs with significantly different morphology and operational functionalities. Furthermore, skill transfer can be achieved with training data acquired from only a single training task and two agents. Other than performance analysis, by visualizing the subspaces learned by PVED, we show that PVED also yields interpretable subspaces for both shared and individual factors.

\section{Problem statement}
\label{sec:problem_statement}

We consider a setting with two MDAs: a {\it source} agent and a {\it target} agent. We assume that the source agent has learned to perform a {\it target task}, and aim to transfer the skills required to perform the target task from the source agent to the target agent. Our only requirement is that the pair of agents has learned a different and simpler {\it training task}, for which some data (i.e. trajectories) has been generated. Accordingly, we consider two domains with distinct input/output distributions and reward structures: the {\it source domain}, denoted by $D_S$, and the {\it target domain}, denoted by $D_T$. Each domain  is modelled with a Markov decision process (MDP), i.e. $D_S = \left (  \mathcal{S}_S, \mathcal{A}_S, \mathcal{T}_S, R_S \right )$ and $D_T = \left (  \mathcal{S}_T, \mathcal{A}_T, \mathcal{T}_T, R_T \right )$, whereby $\mathcal{S}_S$, $\mathcal{S}_T$ are the state spaces, $\mathcal{A}_S$, $\mathcal{A}_T$ are the action spaces, $\mathcal{T}_S$, $\mathcal{T}_T$ are the dynamics or transition functions, and $R_T$, $R_S$ are the reward functions. Due to the morphological heterogeneity, the state and action spaces may be different in the source and target spaces, and the dynamics may also differ. For the same task, we assume some similarities exist in $R_T$ and $R_S$, as the two agents share similar final goals. 

\section{Paired variational model for skill transfer} \label{sec:method}

\subsection{Representing the control mechanism} \label{sec:method_repre_operation_patterns}

Let $s \in \mathcal{S}$, $a \in \mathcal{A}$. We represent the control of both agents using the joint distributions of states and actions: $p \left ( s_S, a_S \right )$ and $p \left ( s_T, a_T \right )$. The rationale is that both states and actions are important to define the control of the agents. Intuitively, the actions define the control parameters and the states reflect the condition under which specific control actions are used. For example, two robot arms perform the same actions with different joint angles (i.e. state differences) could result different control in the end effector; two robot arms with similar joint angles, but moving towards different directions (i.e. action differences) also reflect different controls. As in previous work~\cite{InvariantSubspaceLearning_Gupta_2017}, we assume $\left ( s_S, a_S \right )$ and $\left ( s_T, a_T \right )$ can be paired in the model learning stage, i.e. a correspondence between them can be established; see Section~\ref{sec:experiment_settings} for further details.

\subsection{Decomposing the control mechanism} \label{sec:method_assumption}

We assume the existence of two factors determining the control of the two MDAs: shared and individual factors. The shared factors are meant to capture high-level similarities that are expected to exist in the control stages for a particular task despite the morphological differences. For example, for a button pushing task executed by two very different robot arms, the shared factors would represent the control stages requiring to approach and push the button whereas the individual factors would capture the specific control parameters required by each arm in each stage. We model the two factors as shared and individual subspaces, denoted by $\mathcal{C}_S$ and $\mathcal{Y}_S$, respectively, for the source agent, and $\mathcal{C}_T$ and $\mathcal{Y}_T$ for the target agent.  Let $c \in \mathcal{C}$, $y \in \mathcal{Y}$. The two subspaces are defined by probability distributions $p (c_S,y_S|s_S,a_S)$ and $p (c_T,y_T|s_T,a_T)$, for the source and target agents, respectively. We associate the two subspaces with source and target MDPs: $D_S = \left (  \mathcal{S}_S, \mathcal{A}_S, \mathcal{T}_S, R_S, \mathcal{C}_{S}, \mathcal{Y}_{S} \right )$ and $D_T = \left (  \mathcal{S}_T, \mathcal{A}_T, \mathcal{T}_T, R_T, \mathcal{C}_{T}, \mathcal{Y}_{T} \right )$. 

\subsection{A paired variational encoder-decoder (PVED)} \label{sec:method_model}

We require three ideal properties to enable skill transfer: {\bf(1) Stochasticity}, full probabilistic model for subspace learning is required to account for the uncertainty due to finite sample estimation. {\bf(2) Transferability}, the subspaces must  only capture the information transferable between $D_S$ and $D_T$; this includes the shared factors and the individual factors that can be projected between the source and target domains. This reduces the redundancy of the information in the subspaces for the seeking of the shared factors  {\bf(3) Consistency}, the marginal distributions of $c_S$ and $c_T$ must be consistent, in the sense of having high similarity, given the observed pairs of $(s_S,a_S)$ and $(s_T,a_T)$.

We model the two subspaces stochastically using generative models, and specifically variational autoencoders (VAEs)~\cite{VAE_Kingma_2014, VAE_Rezende_2014}. We view the shared and individual factors as the latent generative factors for states and actions. For the target agent, an observation $\left ( s_T, a_T \right )$ can be thought as being generated by first sampling $\left ( c_T, y_T \right )$ from a prior distribution $p\left ( c_T, y_T \right )$, and then generating $\left ( s_T, a_T \right )$ from a conditional distribution,  $p\left ( s_T, a_T | c_T, y_T \right )$. We assume that $p\left ( c_T, y_T \right )$ and $p\left ( s_T, a_T | c_T, y_T \right )$ come from probability distribution parameterized by ${\theta}_T$, i.e. $p_{{\theta}_{T}}\left ( c_T, y_T \right )$ and $p_{{\theta}_{T}}\left ( s_T, a_T | c_T, y_T \right )$. 

To estimate the parameters ${\theta}_{T}$, we maximize the marginal log-likelihood, i.e. $\log p_{{\theta}_{T}}\left ( s_T, a_T \right )$, where $
p_{{\theta}_{T}}\left ( s_T, a_T \right ) = \iint p_{{\theta}_{T}}(s_T,a_T|c_T,y_T)p_{{\theta}_{T}}(c_T,y_T) \,d{c_T} \, d{y_T}$. 
Practically, the integration over $c_T$ and $y_T$ is intractable. Thus, a recognition model $q_{{\phi}_T}(c_T,y_T|s_T,a_T)$ parameterized by ${\phi}_T$ is introduced to approximate the true posterior $p_{{\theta}_T}(c_T,y_T|s_T,a_T)$. Accordingly, we optimize a variational lower-bound of $\log p_{{\theta}_{T}}\left ( s_T, a_T \right )$ to learn both ${\theta}_T$ and ${\phi}_T$~\cite{VAE_Kingma_2014, VAE_Rezende_2014}:
\begin{equation}
\begin{split}
\log\, &p_{{\theta}_{T}}\left ( s_T, a_T \right ) \geq \E_{q_{{\phi}_T}}\left [ \log p_{{\theta}_T}\left ( s_T,a_T|c_T,y_T \right ) \right ] \\ 
&- D_{KL}\left ( q_{{\phi}_T}\left ( c_T,y_T|s_T,a_T \right )||p_{{\theta}_T}\left ( c_T,y_T \right ) \right )
\label{eqn:1}
\end{split}
\end{equation}
where $D_{KL}\left ( \bigcdot  || \bigcdot \right )$ is Kullback-Leibler (KL) divergence. Maximizing the above lower bound leads to a reconstruction of  $\left ( s_T, a_T \right )$ with $\left ( c_T, y_T \right )$ as latent factors. 

Further developments are required to meet the transferability and similarity properties. First, we wish to leverage the paired observations, $\left ( s_S, a_S \right )$ and $\left ( s_T, a_T \right )$ to achieve transferability. Instead of using $q_{{\phi}_T}(c_T,y_T|s_T,a_T)$, we propose using  $q_{{\phi}_S}(c_T,y_T|s_S,a_S)$ to approximate $p_{{\theta}_T}(c_T,y_T|s_T,a_T)$. Here the target posterior is approximated by a distribution conditioned on the source control, which should encourage the recognition model to provide more transferable subspaces. Moreover, we expect that explicitly approximating $y_T$ allows the model to better recognize $c_T$. This approximation implies a different variational lower bound, i.e.
\begin{equation}
\begin{split}
{\mathcal{L}}_{S,T} & = \E_{q_{{\phi}_S}}\left [ \log p_{{\theta}_T}\left ( s_T,a_T|c_T,y_T \right ) \right ] \\ 
&- D_{KL}\left (q_{{\phi}_S}\left ( c_T,y_T|s_S,a_S \right )||p_{{\theta}_T}\left ( c_T,y_T \right ) \right )
\label{eqn:2}
\end{split}
\end{equation}
To see this, we consider the KL divergence between the approximate posterior $q_{{\phi}_S}\left ( c_T,y_T|s_S,a_S \right )$ and the true posterior $p_{{\theta}_T}\left ( c_T,y_T|s_T,a_T \right )$. For brevity, we write $\left ( s_S, a_S \right )$ as $o_S$, $\left ( s_T, a_T \right )$ as $o_T$, $\left ( c_T, y_T \right )$ as $z_T$, $q_{{\phi}_S}$ as $q$, and $p_{{\theta}_T}$ as $p$. The KL term then can be expanded as
\begin{equation}
\begin{split}
&D_{KL}\left (q\left ( z_T|o_S \right )||p \left ( z_T|o_T \right ) \right ) = \E_{q}\log \frac{q\left ( z_T|o_S \right )}{p\left ( z_T|o_T \right )} \\
&= \E_{q}\left [{{\log q\left ( z_T|o_S \right )} - \log \frac{p\left( o_T| z_T \right ) p\left( z_T \right )}{p \left( o_T \right)}}\right ]
\label{eqn:3}
\end{split}
\end{equation}
By decomposing the second term in Eq.~\ref{eqn:3}, 
\begin{equation}
\log p \left( o_T \right) = D_{KL}\left (q\left ( z_T|o_S \right )||p \left ( z_T|o_T \right ) \right ) + {\mathcal{L}}_{S,T}
\label{eqn:4}
\end{equation} 
Since $D_{KL}\left ( \bigcdot  || \bigcdot \right ) \ge 0$, we obtain the lower bound in Eq.~\ref{eqn:2},
\begin{equation}
\log p_{{\theta}_T} \left( s_T, a_T \right) \geq {\mathcal{L}}_{S,T}
\label{eqn:5}
\end{equation} 
Maximizing the bound in Eq.~\ref{eqn:2} yields a variational encoder-decoder (VED): the latent factors $\left ( c_T, y_T \right )$ are encoded from $\left( s_S, a_S \right)$, and decoded to  $\left( s_T, a_T \right)$. Thus, $\left ( c_T, y_T \right )$ preserves the transferable information between the two agents. 

The consistency property is still unmet as no terms control the similarity between the distribution of $c_S$ and $c_T$. To address this, we introduce a second, paired VED, that explicitly infer the distribution of $c_S$ in addition to that of $c_T$. The consistency property is then enforced by minimizing the KL divergence between the inferred distributions of $c_T$ and $c_S$. Specifically, with $\left ( c_S, y_S \right )$ as the latent generative factors of $\left ( s_S, a_S \right )$, we use $q_{{\phi}_T}(c_S,y_S|s_T,a_T)$ to approximate $p_{{\theta}_S}(c_S,y_S|s_S,a_S)$. Following similar derivations as those leading to Eq.~\ref{eqn:3}, Eq.~\ref{eqn:4} and Inequality~(\ref{eqn:5}), we obtain a variational lower bound:
\begin{equation}
\begin{split}
{\mathcal{L}}_{T,S} & = \E_{q_{{\phi}_T}}\left [ \log p_{{\theta}_S}\left ( s_S,a_S|c_S,y_S \right ) \right ] \\ 
&- D_{KL}\left (q_{{\phi}_T}\left ( c_S,y_S|s_T,a_T \right )||p_{{\theta}_S}\left ( c_S,y_S \right ) \right )
\label{eqn:6}
\end{split}
\end{equation}
We use different parameterizations for the probability distributions in Eq.~\ref{eqn:2} and Eq.~\ref{eqn:6} due to the morpholigical variations in the two agents. Consistency is imposed by a KL divergence term:
\begin{equation}
{\mathcal{L}}_{C} = - D_{KL}\left (q_{{\phi}_S}\left ( c_T|s_S,a_S \right )||q_{{\phi}_T}\left ( c_S|s_T,a_T \right ) \right )
\label{eqn:7}
\end{equation}
whose maximization promotes the similarity between the distributions of $c_T$ and $c_S$. With this term, we expect $y_S$ and $y_T$ contains the factors that dissimilar between the two agents; thus they are considered as individual factors. 

The model using Eq.~\ref{eqn:2}, Eq.~\ref{eqn:6} and Eq.~\ref{eqn:7} may have degenerated solutions. For example, the encoders may project state-action pairs to zeros in $c_T$ and $c_S$ to satisfy Eq.~\ref{eqn:7}; the encoders would then project all the information to the individual subspaces to maximize Eq.~\ref{eqn:2} and Eq.~\ref{eqn:6}. Such cases are clearly undesirable, as the shared subspace includes little information. To address this issue, we add two terms to facilitate the information to flow in the shared subspaces. We require $q_{{\phi}_S}\left ( c_T|s_S,a_S \right )$ to approximate the true posterior $p_{{\theta}_T^{\prime}}(c_T|s_S,a_S)$, and $q_{{\phi}_T}\left ( c_S|s_T,a_T \right )$ to approximate $p_{{\theta}_S^{\prime}}(c_S|s_T,a_T)$. This leads to two variational lower bounds:
\begin{equation}
\begin{split}
{\mathcal{L}}_{S,S} & = \E_{q_{{\phi}_S}}\left [ \log p_{{\theta}_T^{\prime}}\left ( s_S,a_S|c_T \right ) \right ] \\ 
&- D_{KL}\left (q_{{\phi}_S}\left ( c_T|s_S,a_S \right )||p_{{\theta}_T^{\prime}}\left ( c_T \right ) \right ) \\
{\mathcal{L}}_{T,T} & = \E_{q_{{\phi}_T}}\left [ \log p_{{\theta}_S^{\prime}}\left ( s_T,a_T|c_S \right ) \right ] \\ 
&- D_{KL}\left (q_{{\phi}_T}\left ( c_S|s_T,a_T \right )||p_{{\theta}_S^{\prime}}\left ( c_S \right ) \right )
\label{eqn:8}
\end{split}
\end{equation}

\begin{figure}[t!]
\begin{center}
\centerline{\includegraphics[width=0.9\columnwidth]{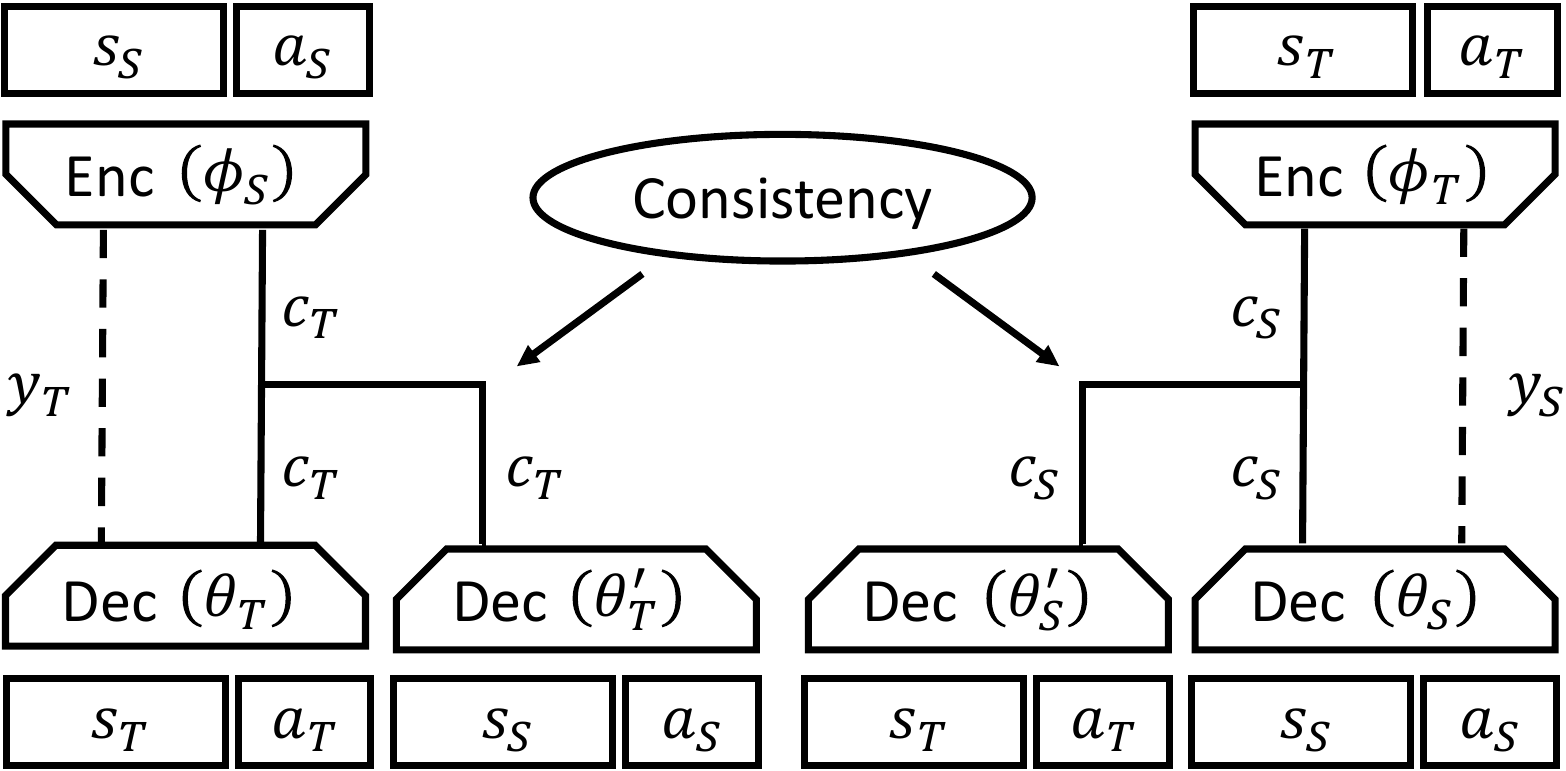}}
\vskip -0.1in
\caption{Illustration of the proposed PVED model. Enc and Dec are abbreviations for encoder and decoder, respectively. A pair of VEDs is formed to infer the distribution of shared and individual factors ($c$ and $y$, respectively). A consistency property is imposed on the inferred distributions of the shared factors between the two VEDs.}
\label{fig1}
\end{center}
\end{figure}

As final objective, we maximize a weighted sum of the terms in Eq.~\ref{eqn:2}, Eq.~\ref{eqn:6}, Eq.~\ref{eqn:7} and Eq.~\ref{eqn:8} , 
\begin{equation}
\begin{split}
{\mathcal{L}} = {\alpha}_{1}\left ( {\mathcal{L}}_{S,T} + {\mathcal{L}}_{T,S} \right )+ {\alpha}_{2}\left ( {\mathcal{L}}_{S,S} + {\mathcal{L}}_{T,T} \right ) + {\alpha}_{3}{\mathcal{L}}_{C}
\label{eqn:9}
\end{split}
\end{equation}
where each individual weight, $\alpha_i$, is a non-negative scalar (see Section~\ref{sec:experiment_settings} for more details on how the weights are selected). An illustration of the resulting PVED model is given in Fig.~\ref{fig1}.

\subsection{Inter-agent skill transfer}
\label{sec:method_transfer}

Upon inferring the shared and individual subspaces using the model in Section~\ref{sec:method_model}, we leverage the shared subspaces for skill transfer between the two MDAs. We use a reward-shaping method that adds a term to the environmental reward. Instead of using the Euclidean distance that is most suited for deterministic subspaces (e.g. \cite{InvariantSubspaceLearning_Gupta_2017}), we employ the KL divergence between stochastic subspaces:
\begin{equation}
R = - \beta D_{KL}\left (q_{{\phi}_S}\left ( c_T|s_S,a_S \right )||q_{{\phi}_T}\left ( c_S|s_T,a_T \right ) \right )
\label{eqn:10}
\end{equation}
Maximizing this term promotes the target agent to perform actions such that the inferred distribution of the shared factor from the target agent matches the inferred distribution from the source agent as much as possible. Thus, skill transfer is achieved through the shared subspace. This approach does not require paired trajectories between source and target agents for the target tasks. This leads the target agent to learn to match the operating rate of the source agent as much as possible. For policy learning, in our experiments, we have employed the proximal policy optimization (PPO) algorithm~\cite{PPO_Schulman_2017}, due to its widely reported efficiency and excellent performance.

\subsection{Theoretical analysis}
\label{sec:method_theoretical_analysis}

We theoretically study the factors that influence the performance of the proposed skill transfer method. Let ${\pi}_S$ be the source policy pretrained on the target task, and ${\pi}_T$ the target policy transferred from ${\pi}_S$ with the proposed method. Due to the morphological differences and transfer error, the reward achieved by ${\pi}_T$ is usually different from that achieved by ${\pi}_S$. Thus, we target the performance gap between target and source policy, in terms of the expected reward and return (i.e. sum of discounted reward). Also, since ${\pi}_S$ can be considered optimal or near-optimal in our settings, the performance gap also provides a performance guarantee for ${\pi}_T$. 

Firstly, we focus on the achieved reward. Let $\left(s_T, a_T\right) \sim {\pi}_{T}$ and $\left(s_S, a_S\right) \sim {\pi}_{S}$. Since we use a KL divergence term for reward shaping to achieve skill transfer (Eq.~\ref{eqn:10}). we study how this KL divergence term influences the achieved reward. Specifically, under the condition $D_{KL}\left (q_{{\phi}_S}\left ( c_T|s_S,a_S \right )||q_{{\phi}_T}\left ( c_S|s_T,a_T \right ) \right ) \leqslant \delta$, we consider the difference in the expected rewards:
\begin{equation}
\left | \E_{s_T, a_T \sim {\pi}_{T}} R_T \left( s_T, a_T \right) - \E_{s_S, a_S \sim {\pi}_{S}} R_S \left( s_S, a_S \right)\right |
\label{eqn:11}
\end{equation}

To derive a bound for Eq.~\ref{eqn:11}, some assumptions need to be made on the reward functions. Recall that, in Section~\ref{sec:problem_statement}, we have assumed that similarities exist in $R_T$ and $R_S$. However, it is difficult to define such similarity with $R_T \left( s_T, a_T \right)$ and $R_S \left( s_S, a_S \right)$. This is because the correspondence between $\left( s_T, a_T \right)$ and $\left( s_S, a_S \right)$ is unknown and hard to obtain, given the fact that the state and action spaces of MHAs may have different dimensions. Luckily, Eq.~\ref{eqn:7} bridges $\left( s_T, a_T \right)$ and $\left( s_S, a_S \right)$ through the shared subspaces $c_S$ and $c_T$. Therefore, we write the reward functions in terms of the shared subspaces.
\newline {\bf Assumption 1}: ${\phi}_S$ defines an injective mapping from $\left( s_S, a_S \right)$ to $c_T$, so does ${\phi}_T$ from $\left( s_T, a_T \right)$ to $c_S$. In other words, the learned PVED model maps each state-action pair to an unique feature representation in the shared subspaces. 

With Assumption 1, ${\theta}_T^{\prime}$ in Eq.~\ref{eqn:8} actually defines an inverse mapping $\left( s_S, a_S \right) = f_{{\theta}_T^{\prime}}\left(c\right)$. Thus, the reward function of the source agent can be written as $R_S\left( s_S, a_S \right) = R_S\left( f_{{\theta}_T^{\prime}}\left(c\right) \right) = R_S^{\prime}\left(c\right)$. Similaly we have $R_T\left( s_T, a_T \right) = R_T\left( f_{{\theta}_S^{\prime}}\left(c\right) \right) = R_T^{\prime}\left(c\right)$. Therefore, we can rewrite Eq.~\ref{eqn:11} as follows:
\begin{equation}
\left | \E_{c \sim {\pi}_{T}} R_T^{\prime} \left( c \right) - \E_{c \sim {\pi}_{S}} R_S^{\prime} \left( c \right)\right |
\label{eqn:12}
\end{equation}
The following theorem gives an upper bound for Eq.~\ref{eqn:12}.
\newline {\bf Theorem 1}. Given $\left(s_T, a_T\right) \sim {\pi}_{T}$ and $\left(s_S, a_S\right) \sim {\pi}_{S}$, such that $D_{KL}\left (q_{{\phi}_S}\left ( c_T|s_S,a_S \right )||q_{{\phi}_T}\left ( c_S|s_T,a_T \right ) \right ) \leqslant \delta$. Given that the source reward function is bounded: $R_S^{\prime}\left(c\right) \in \left[a, b\right]$, where $a,b \in \mathbb{R}$ and $a<b$. Given the similarity of the source and target reward functions: $\left | R_T^{\prime}\left(c\right) - R_S^{\prime}\left(c\right) \right | \leqslant m$, where $c \in \mathcal{C} = \mathcal{C}_{S}\bigcup\mathcal{C}_{T}$, $m \in \mathbb{R}_{\geqslant 0}$. We have the following bound:
\begin{equation}
\left | \E_{c \sim {\pi}_{T}} R_T^{\prime} \left( c \right) - \E_{c \sim {\pi}_{S}} R_S^{\prime} \left( c \right)\right | \leqslant m + \frac{\left(b-a\right)}{2} \sqrt{2\delta}
\label{eqn:13}
\end{equation}
\newline {\it Proof}. See Appendix.

Next, we consider the RL return function $J\left(\tau \right) = \sum_{t}{{\gamma}^t}R^{\prime}\left(c_t\right)$ where $\gamma \in \left[0, 1\right]$ is the discounting factor, $c_t$ is the shared factor at timestep $t$, $\tau = \left(c_0, c_1 ... \right)$. A bound on the gap of expected return can also be established as follows:
\newline {\bf Corollary 1}. Under the same conditions and assumptions as in Theorem 1, we have the following bound:
\begin{equation}
\left | \E_{\tau \sim {\pi}_{T}} J_T \left( \tau \right) - \E_{\tau \sim {\pi}_{S}} J_S \left( \tau \right)\right | \leqslant \left(\frac{1}{1-\gamma}\right)\left[m + \frac{\left(b-a\right)}{2} \sqrt{2\delta}\right]
\label{eqn:14}
\end{equation}
\newline {\it Proof}. See Appendix.

Theorem 1 and Corollary 1 indicate that, with the proposed method for skill transfer, the performance gap between source and target agents is determined by two factors: the agent morphology and the shared subspace. Specifically, The terms with $\delta$ in the right hand side (RHS) of Inequality~(\ref{eqn:13}) and Inequality~(\ref{eqn:14}) show the influence of the shared subspace. The closer the target policy is to the source policy in the shared subspace (i.e. smaller $\delta$), the closer the performance between the target and the source agent. On the other hand, the terms with $m$ in RHS show the influence of both agent morphologies and shared subspace. Explicitly, for the same task, the difference in the reward functions is driven by morphological heterogeneity, hence this term shows the influence of different agent morphologies on the performance of skill transfer. Implicitly, note that $c$ is mapped from $\left(s, a\right)$; thus, $m$ is tighter if the learned mappings are more correct (i.e. highly corresponding $\left(s_S, a_S\right)$ and $\left(s_T, a_T\right)$ are mapped to closer $c_T$ and $c_S$). As a special case of Theorem 1 and Corollary 1, when the source and target agents have the same morphology and the mapping from $\left(s, a\right)$ to $c$ is fully correct (i.e. $m=0$), and when the target policy perfectly matches the source policy in the shared subspace (i.e. $\delta=0$), the target policy is guaranteed to be as good as the source policy (i.e. the RHS of Inequality~(\ref{eqn:13}) and Inequality~(\ref{eqn:14}) equal to 0).

\subsection{Implementation details}
\label{sec:method_implementation_details}

In this subsection, we present the implementation details of the proposed paired variational encoder-decoder (PVED) model. We use normal distributions as the prior distributions for the KL divergence terms in Eq.~\ref{eqn:2}, Eq.~\ref{eqn:6}, and Eq.~\ref{eqn:8}. We model the data distributions in the shared and individual subspaces as multivariate Gaussian distributions with diagonal covariance matrices. We use three-layer neural networks for the recognition models (the encoders) and the generative models (the decoders) in Eq.~\ref{eqn:2}, Eq.~\ref{eqn:6}, and Eq.~\ref{eqn:8}. For the encoders, the weights are shared between the two subspaces except the last layer which outputs the mean and variances for each subspace individually. The decoders learn projections from the shared and individual subspaces separately in their first layers, and they concatenate the output of the first layer as the input for the rest two layers. We use ReLU activations for the encoders, and leaky ReLU for the decoders. For each network, we use 64 hidden units in each layer (128 in the first hidden layer for the decoders parameterized by ${\theta}_S$ and ${\theta}_T$ in Eq.~\ref{eqn:2} and Eq.~\ref{eqn:6}: 64 for the shared subspace and 64 for the individual subspace). The dimensions of the mean and variance of the shared and individual latent subspaces are set to 10. For the objective function in Eq.~\ref{eqn:9}, we tune ${\alpha}_1$, ${\alpha}_2$, ${\alpha}_3$  and $\beta$ for different tasks. The detailed settings are presented in section~\ref{sec:experiment_settings}.

For policy learning with PPO, we use an actor-critic-style implementation. Both actor are critic are modelled as 3-layer neural networks, with 64 hidden unit for each hidden layer and ReLU activations. This includes the policies for the source and target agents on the training tasks, the policies for the source agent in the target task, and the policies learned by skill transfer from the source to the target agent with the added reward term as in Section~\ref{sec:method_transfer}.

\section{Related Work}

{\bf Transfer learning in RL}. Among the extensive work in transfer learning in RL, PVED is closely related to hierarchical models~\cite{HierarchicalRL_Sutton_1999, HierarchicalRL_Kulkarni_2016, HierarchicalRL_Bacon_2017, HierarchicalRL_Vezhnevets_2017, HierarchicalRL_Andreas_2017, HierarchicalRL_Nachum_2018, HierarchicalRL_Tirumala_2019}. Given a task, the high-level goals could be related to the shared factors, and the low-level actions to agent-specific factors. Hierarchical models are able to handle LMH problems by fixing the high-level policies and fine-tuning the low-level policies~\cite{HierarchicalRL_Tirumala_2019}. However, this usually requires to manually define the agent-agnostic and agent-related states as the input to the high-level and low-level policies, respectively.  Meta-learning in RL seeks to exploit patterns across multiple task or dynamic domains to quickly learn new tasks~\cite{MetaLearning_Duan_2016, MetaLearning_Finn_2017, MetaLearning_Mishra_2018, MetaLearning_Yoon_2018, MetaLearning_Grant_2018, MetaLearningDynamics_Nagabandi_2019}. The shared factors in PVED could be interpreted as patterns relating different domains of MDAs. Despite the similarities, most meta-learning models in RL have only considered agents with very similar or fixed morphologies.

Recently, increasing research has focused on LMH problems. Several methods seek a projection relating the state of the source and target agents to enable skill transfer~\cite{ManifoldAlignment_Bocsi_2013, ManifoldAlignment_Ammar_2015, LearningProjection_Raimalwala_2016}. For agents with large morphological differences, such a  mapping between state spaces may not always exist. In~\cite{TransferLearningApprentice_Joshi_2018}, an adaptive policy is introduced to compensate the projection error between the source and target agents. This method assumes known correspondence in the action spaces between the two agents. In~\cite{ModularDRLTrasfer_Devin_2017}, policies are learned with separated modules for tasks and agents to achieve skill transfer among different agents. In~\cite{HardwareEncoding_Chen_2018}, the agent morphology is encoded as a feature vector; the vector is concatenated with the states of agents to learn policies that can generalize to different morphologies. The learning processes  in~\cite{ModularDRLTrasfer_Devin_2017,HardwareEncoding_Chen_2018} are based on multiple agents and/or multiple tasks incurring increased effort required to design agents and tasks. Graph-based policies~\cite{GNNPolicy_Wang_2018} construct a graph representation for the agent, and skills are transferred by sharing policies for the graph nodes with similar physical structures. This policy sharing scheme may be less applicable when the agents have less similar physical structures and/or very different operational functionalities. The method in~\cite{InvariantSubspaceLearning_Gupta_2017} is the most relevant LMH approach. A deterministic latent subspace is inferred from state spaces to capture the information shared between agents. Given the stochasticity in most RL environments and models, a single deterministic latent subspace may be less optimal for varying MDAs and tasks in RL. In contrast, PVED explicitly models a shared subspace and an individual subspace with stochastic models, and infer the two subspaces from joint state and action distributions. 

{\bf Latent feature learning via variational inference}. Variational inference has been widely exploited to learn and disentangle multiple latent generative factors in many research areas~\cite{VAE_DRL_Watter_2015, VAE_LatentSubspaceLearningOtherAreas_Mathieu_2016, VAE_LatentSubspaceLearningOtherAreas_Higgins_2017, VAE_LatentSubspaceLearningOtherAreas_Serban_2017, VAE_LatentSubspaceLearningOtherAreas_Jha_2018, VAE_LatentSubspaceLearningOtherAreas_Achille_2018, VAE_LatentSubspaceLearningOtherAreas_Dupont_2018, VAE_DRL_Nair_2018, VAE_LatentSubspaceLearningOtherAreas_Jeong_2019}. Extensive research has also leveraged variational models to learn latent features to improve the information transfer in DRL~\cite{VAE_DRL_Higgins_2017, TaskTransfer_Hausman_2018, VAE_DRL_Inoue_2018, VAE_DRL_Corneil_2018, VAE_DRL_Ha_2018, DynamicTransfer_Tirinzoni_2018}. However, since most existing methods focus on data with fixed dimensions (images, states of a fixed agent, etc), they are less applicable to MDAs with varying dimensions in the state and action spaces. Moreover, in transfer learning in DRL, most existing methods use a single latent space to model all the data generative factors; few works exploit multiple subspaces to explicitly model and disentangle different factors as in the proposed PVED. Although variational models for modality translations~\cite{ModalityTranslation_Pu_2016,ModalityTranslation_Wang_2017,ModalityTranslation_Su_2018} are able to handle data with different dimensions, these methods focus on learning a direct mapping between two modalities, rather than inferring a shared subspace for policy learning as in the proposed method (note that a mapping between MHAs may not always exist~\cite{InvariantSubspaceLearning_Gupta_2017}). 

\begin{figure*}[t!]
\begin{center}
\subfigure[Robot arms pair 1, training]
{
\includegraphics[height=38mm]{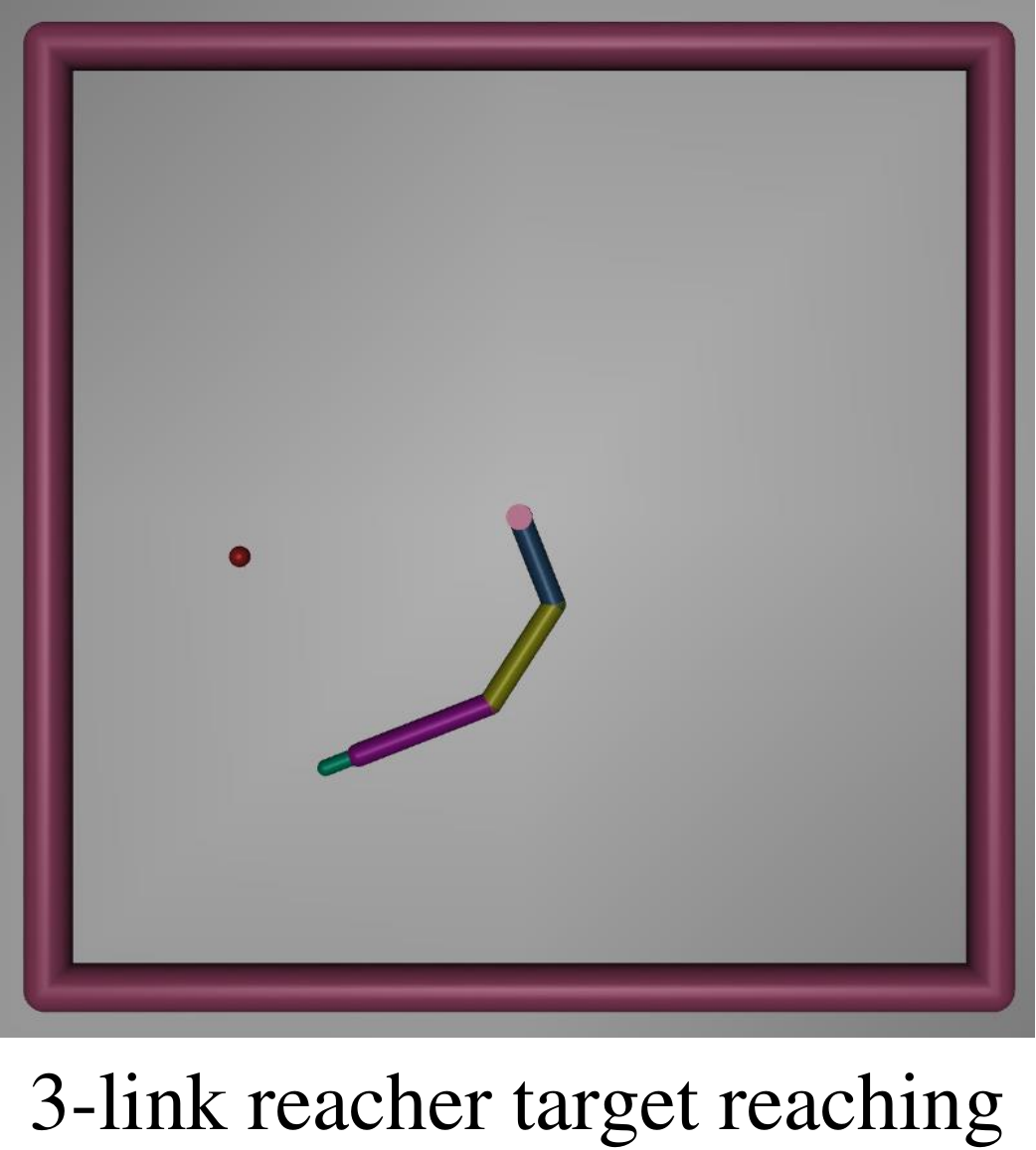}
\label{fig_experi_agent_env_reacher_reaching}
}
\hspace{-0.1in}
\subfigure[Robot arms pair 1, target]
{
\includegraphics[height=38mm]{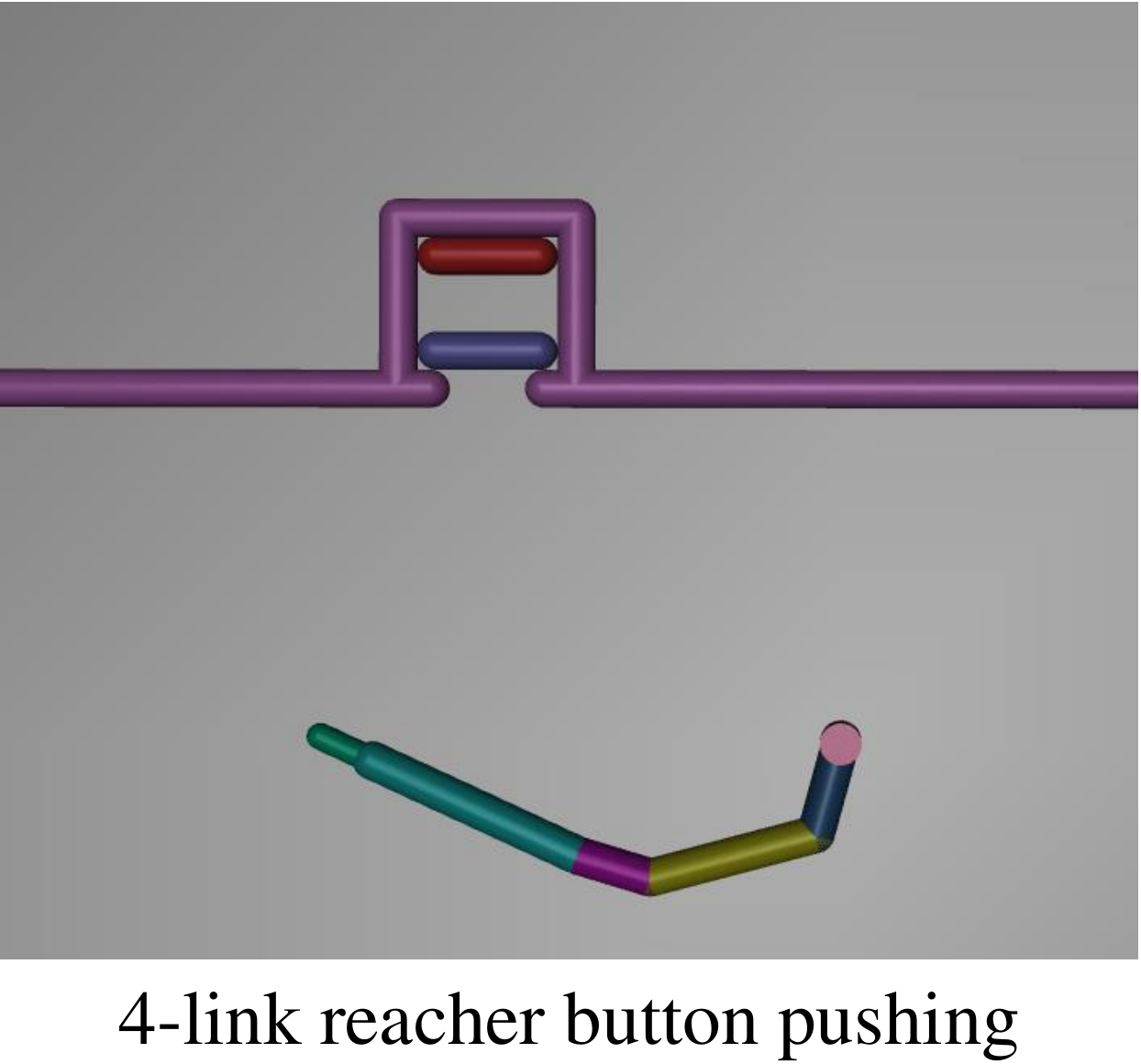}
\label{fig_experi_agent_env_reacher_button_pushing}
}
\hspace{-0.1in}
\subfigure[Robot arms pair 2, training]
{
\includegraphics[height=38mm]{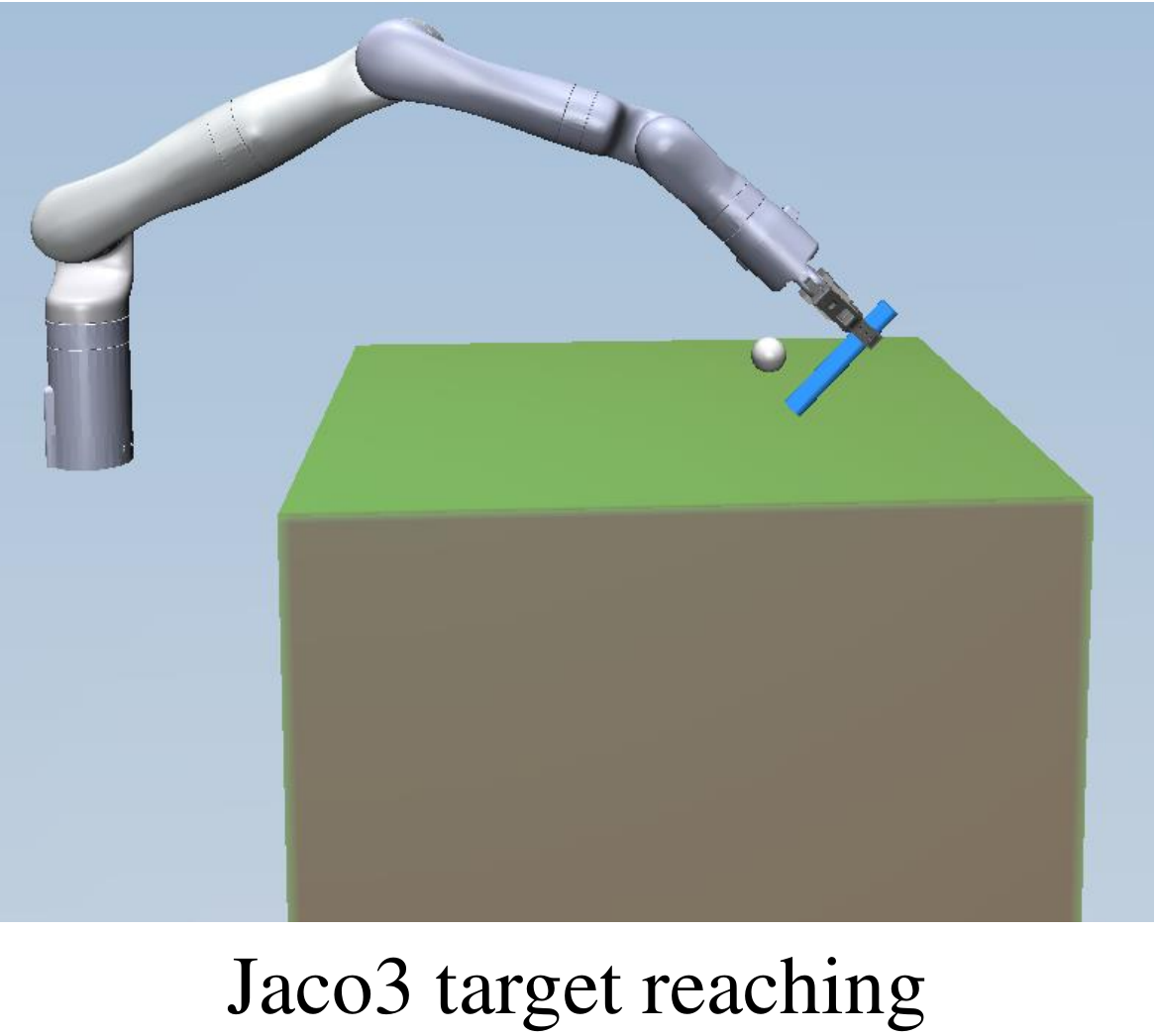}
\label{fig_experi_agent_env_robot_arms_reaching}
}
\hspace{-0.1in}
\subfigure[Robot arms pair 2, target]
{
\includegraphics[height=38mm]{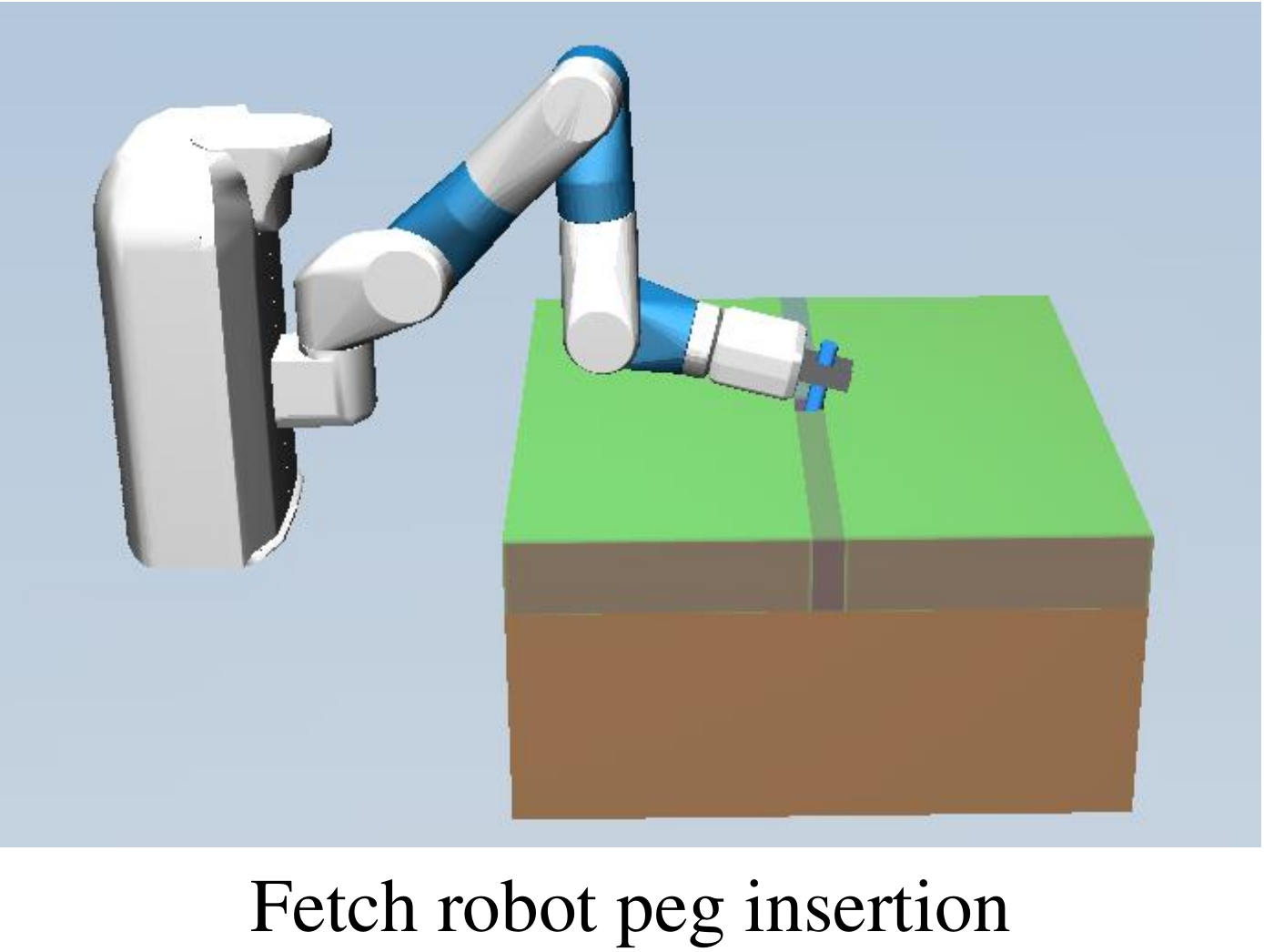}
\label{fig_experi_agent_env_robot_arms_peg_insertion}
}
\hspace{-0.1in}
\subfigure[Locomotion robots pair 1, training]
{
\includegraphics[height=38mm]{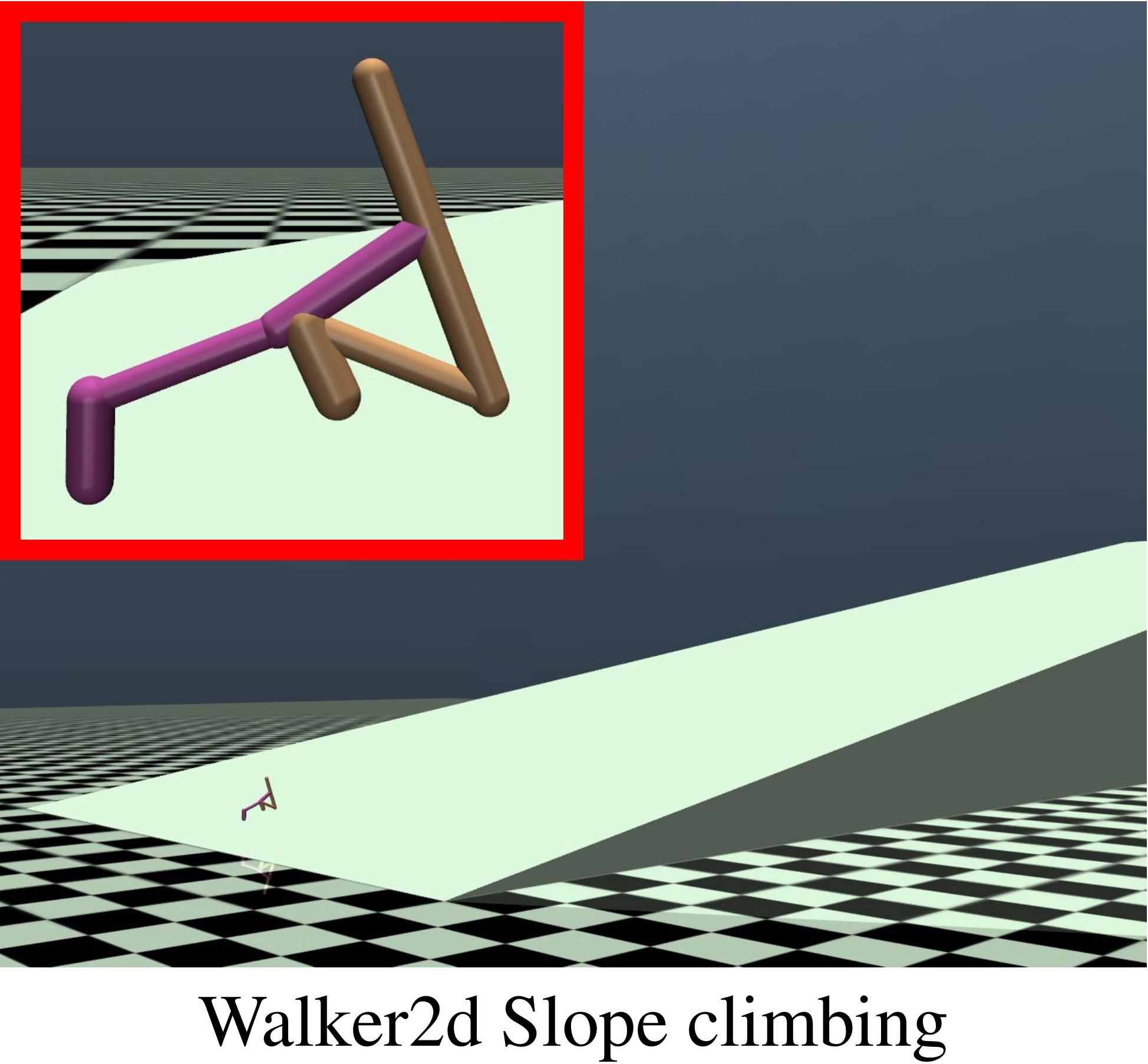}
\label{fig_experi_agent_env_lmrp1_hc}
}
\hspace{-0.1in}
\subfigure[Locomotion robots pair 1, target]
{
\includegraphics[height=38mm]{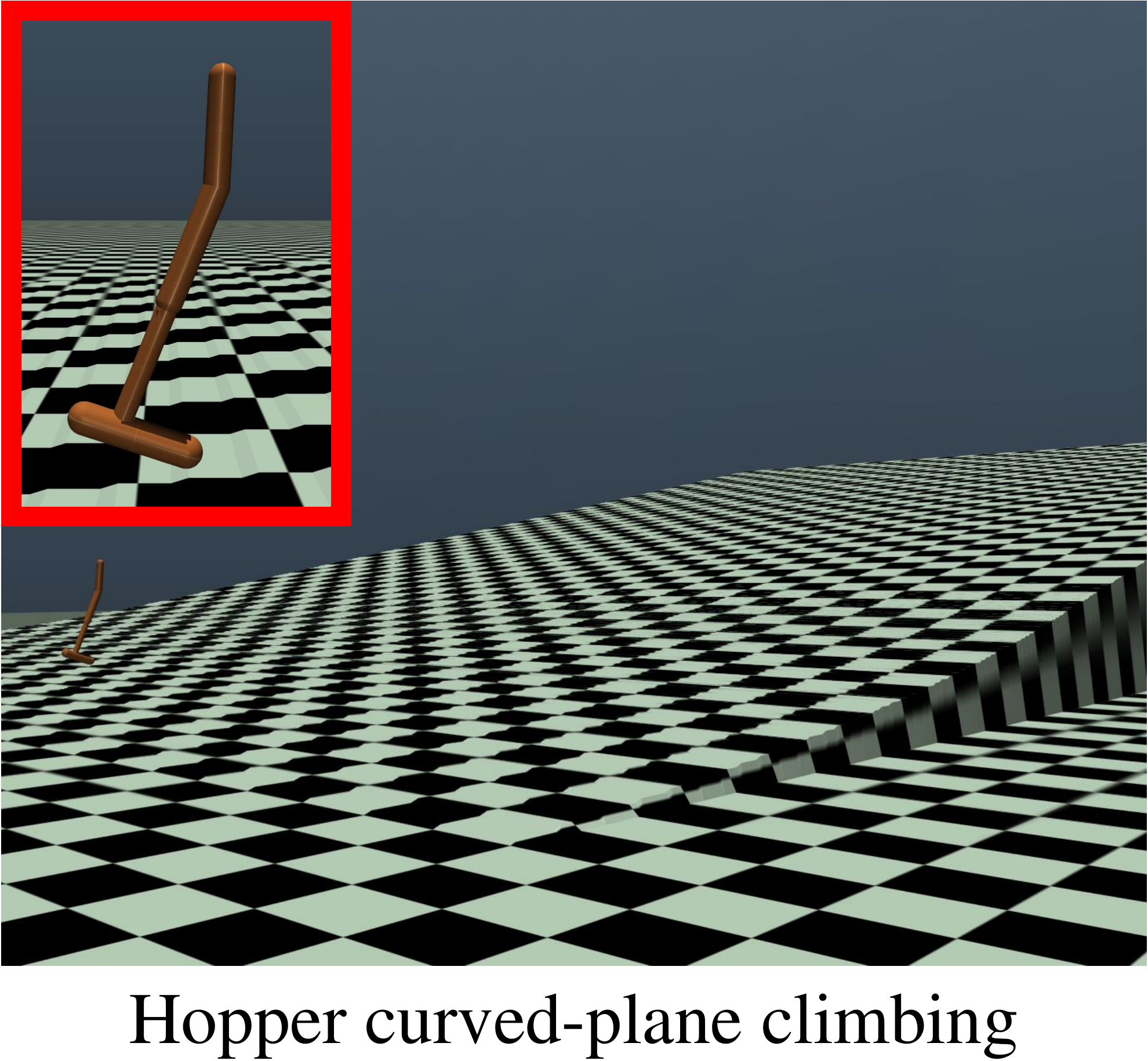}
\label{fig_experi_agent_env_lmrp1_cpc}
}
\hspace{-0.1in}
\subfigure[Locomotion robots pair 2, training]
{
\includegraphics[height=38mm]{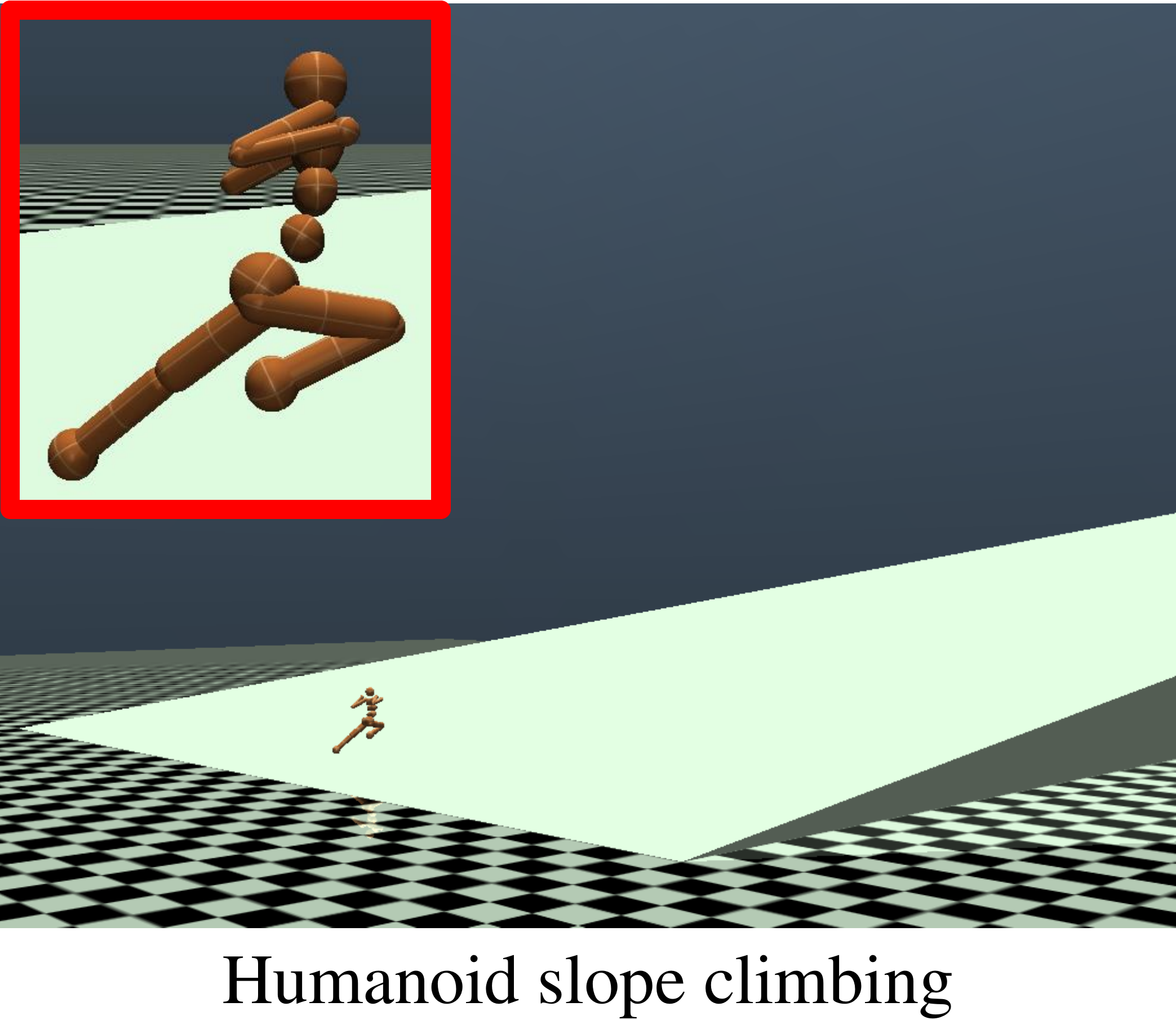}
\label{fig_experi_agent_env_lmrp2_hc}
}
\hspace{-0.1in}
\subfigure[Locomotion robots pair 2, target]
{
\includegraphics[height=38mm]{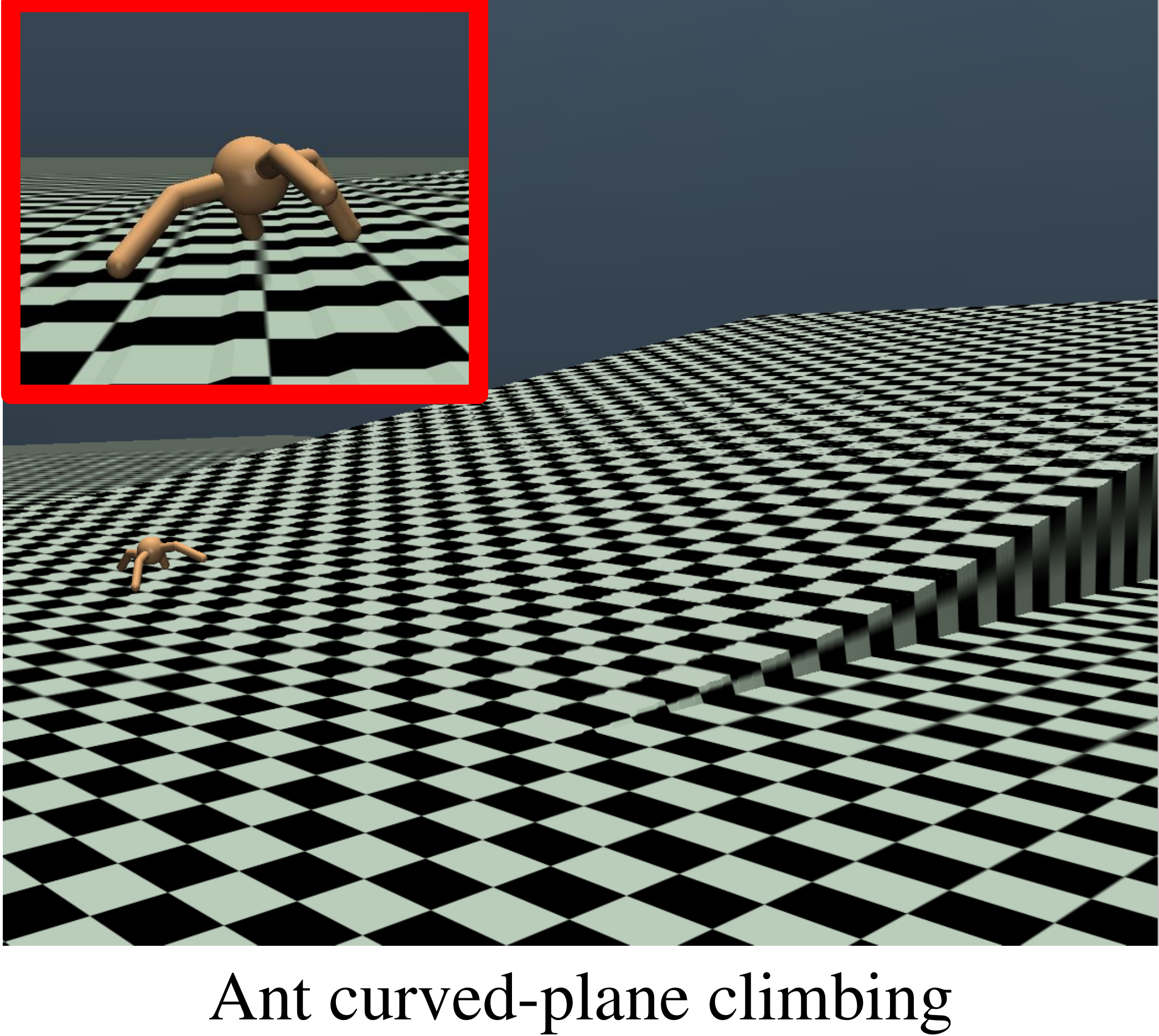}
\label{fig_experi_agent_env_lmrp2_cpc}
}
\caption{Illustrations of the agents and environments of training/target tasks used in our  experiments. We show the source agent in the training task and the target agent in the target task.}
\label{fig_experi_example_agent_env}
\end{center}
\end{figure*}

\section{Experimental results} \label{sec:experiment}

\subsection{Experimental settings} \label{sec:experiment_settings}

{\bf Agents and environments}. We study PVED on two types of agents: robot arms with similar shapes and locomotion robots with very different morphologies requiring quite different control mechanisms. We create our agents and environments using the Mujoco physics engine~\cite{Mujoco_Todorov_2012} and OpenAI's  Gym~\cite{Brockman_OpenAIGym_2016}. Fig.~\ref{fig_experi_example_agent_env} shows examples of these agents and environments. The training and target tasks have been designed such that the target tasks need similar, but further developed skills in addition to the training tasks. For the robot arms, we use two pairs of agents. Fig.~\ref{fig_experi_agent_env_reacher_reaching} and Fig.~\ref{fig_experi_agent_env_reacher_button_pushing} show the first pair of agents. We use a 3-link reacher robot as the source agent, and a 4-link reacher robot as the target agent. The link lengths vary in each robot. The training task is target reaching, where each robot is required to move its end effector to a target position that is randomly generated in each episode. The target task is button pushing. This task requires the robots to push a button along a channel with the button location randomly generated around a point. This button pushing task is considered completed if (1) the button reaches the goal location (red bar in Fig.~\ref{fig_experi_agent_env_reacher_button_pushing}) and (2) the end effector of the robot moves little after (1). For both agents on the source tasks, the reward is based on the distance between the end effector and the target location. For the source agent on the target task, the reward is based on the distances between the end effector and the button, and between the button and the goal. We use simpler reward for these tasks, as they are assumed to be pre-learned in our problem setting (see section~\ref{sec:problem_statement}). For the target agent on the target task, we use a difficult reward which is based on the negative distance between the button and the goal position. This reward does not encourage the target agent to approach the button. Therefore, for the target agent, skills must be transferred from the source agent to complete the target task. 

The second pair of robot arms are based on the MuJoCo models for two real robot arms. As shown in Fig.~\ref{fig_experi_agent_env_robot_arms_reaching} and Fig.~\ref{fig_experi_agent_env_robot_arms_peg_insertion}, we use a Jaco3 arm with 7 degrees of freedom (DoF) as the source agent, and we use the Fetch robot in the OpenAI Gym as the target agent. The original Fetch robot has 7 DoF. To increase difficulty, we disable 3 of its joints to modify it to a 4 DoF arm. The training task is target reaching. The target task is peg insertion, which requires the robot to insert a peg into a hole on a table. For both agents on the source tasks, the reward is based on the distance between the end of the peg and the target location. For the source agent on the target task, the reward is based on the distances between the end of the peg and the target location in the middle of the hole. For the target agent on the target task, the agent receives reward only if the end of the peg is in the hole.

For the locomotion robots, we also use two pairs of agents. In the first pair of agents, we use the Walker2d in the MuJoCo environments from OpenAI Gym as the source agent, and the Hopper in this library as the target agent, as shown in Fig.~\ref{fig_experi_agent_env_lmrp1_hc} and Fig.~\ref{fig_experi_agent_env_lmrp1_cpc}. The two agents have not only very different morphologies, but also distinguished operational functionalities (walking vs. hopping). The training task for the two agents is slope-climbing. It requires the two agents to climb a slope as fast as possible without falling. The slope angle is randomly generated between 2 to 10 degrees in each episode. The target task is curved-plane climbing where the agents need to climb a curved surface with gradual increasing tangential angle up to 10 degrees in 12 meters, after which the tangential angle does not change. The target task requires the agents to adjust gait to handle the changing slopes in different parts of the surface to move fast without falling. For both agents on the source task and the source agent on the target task, the reward is mainly based on the velocity alone positive $x$-direction and a bonus for not falling. For the target agent on the target task, the reward only includes a bonus for not falling. No rewards exist to encourage the target agent to move forward. Therefore, in the target task, the target agent must learn from the source agent to move forward quickly on a curved surface without falling.

Fig.~\ref{fig_experi_agent_env_lmrp2_hc} and Fig.~\ref{fig_experi_agent_env_lmrp2_cpc} illustrate the second pair of locomotion robots. We use the Humanoid in the MuJoCo environments from OpenAI Gym as the source agent, and the Ant in this library as the target agent. In addition to very different morphologies and distinguished operational functionalities, the two agents have complex state and action spaces, imposing more challenges on the inference of the shared and individual subspaces. The training/target tasks and reward settings are similar to the ones used for the first pair of locomotion robots. The only difference is that, for the target agent on the target task, we still include the velocity alone positive $x$-direction into the reward. This is because we find that the complexity of this pair of agents makes it challenging to transfer the skill of moving forward without a specific reward. With this setting, we investigate if the model can transfer skills from a highly complex source agent, to benefit the learning of a very different and complex target agent.

\begin{table}
\caption{Parameter settings for all environments.}
\label{tab:parameter_setting}
\begin{center}
\begin{tabular}{|c|c|c|c|c|}
\hline
Agents & $\alpha_1$ & $\alpha_2$ & $\alpha_3$ & $\beta$\\
\hline
Reachers (Fig.~\ref{fig_experi_agent_env_reacher_reaching}, Fig.~\ref{fig_experi_agent_env_reacher_button_pushing}) & 1.0 & 1.0 & 10.0 & 0.1\\
Jaco3-Fetch (Fig.~\ref{fig_experi_agent_env_robot_arms_reaching}, Fig.~\ref{fig_experi_agent_env_robot_arms_peg_insertion}) & 1.0 & 1.0 & 10.0 & 0.01\\
Walker2d-Hopper (Fig.~\ref{fig_experi_agent_env_lmrp1_hc}, Fig.~\ref{fig_experi_agent_env_lmrp1_cpc}) & 1.0 & 1.0 & 10.0 & 0.02\\
Humanoid-Ant (Fig.~\ref{fig_experi_agent_env_lmrp2_hc}, Fig.~\ref{fig_experi_agent_env_lmrp2_cpc}) & 2.5 & 2.5 & 15.0 & 0.01\\
\hline
\end{tabular}
\end{center}
\end{table}

{\bf Training of PVED}. The proposed PVED model assumes paired $\left ( s_S, a_S \right )$ and $\left ( s_T, a_T \right )$ obtained from the training tasks. To generate paired trajectories, we train individual policies for each agent on the corresponding training tasks with PPO. For the training tasks of robot arms, we generate paired trajectories as follows: (1) we set the same root location, similar random initialization and the same random target position for both source and target agents; (2) we run the trained policies of each agent simultaneously and collect the trajectories for each agent for an episode; (3) we align the trajectories between the two agent by subsampling the longer trajectories along the timesteps. We repeat the above steps over multiple episodes to collect a dataset of paired trajectories. For the locomotion robots, we follow a similar procedure with the exept that in step (1) we impose that the angles of the random slopes are the same for both agents in the training tasks. Experimentally, we found that the alignment of trajectories obtained by a computationally inexpensive and fast subsampling procedure is sufficient to achieve good performance. In all our experiments, the PVED model is trained using backpropagation with the reparameterization trick~\cite{VAE_Kingma_2014}, and using the Adam optimizer~\cite{AdamSGD_Kingma_2015} with batch size of $256$ and initial learning rate of $1\mathrm{e}{-4}$.

{\bf Parameter-tuning and performance evaluation}. Due to the known issue of DRL sensitivity to initialization and hyper-parameters, we use separated seeds in the DRL stage in Section~\ref{sec:method_transfer} for parameter tuning and performance evaluation. For parameter tuning, the seeds are 2000, 4000, 6000, 8000, 10000. We select the parameters based on the performance of the target agent in the target tasks. For performance evaluation, the seeds are 20000, 40000, 60000, 80000, 100000. For each seed, we perform skill transfer with 3 different source policies (i.e. the policies of the source agent are trained with 3 different random seeds). It leads to $3 \times 5 = 15$ runs for each method. Note that, we find that different initialization for the PVED model may vary the optimal setting of $\beta$ in Eq.~\ref{eqn:10}. To reduce the parameter tuning burden, we firstly determine $\beta$ with one PVED model with a specific random initialization, and we apply the same $\beta$ to two additional PVED models learned with different random initialization. The model with the best performance on the training seeds is selected. This procedure applies to all the comparison methods later described in this paper. The final parameter settings for the PVED model is reported in Tab.~\ref{tab:parameter_setting}.

As performance measures, we use the success rate over $500$ episodes for the button pushing and peg insertion tasks, and the mean distance traveled along the $x$-axis in $1000$ timesteps over $100$ episodes for the curved-plane climbing task.

\subsection{Performance analysis and comparisons}
\label{sec:experiment_analysis_cmp}
A number of ablation studies have been carried out to understand the role and importance of the different components required by the PVED model. We consider four variants of PVED. {\bf (1) PVED learned from state distribution only (referred to as PVED{\it -state only})}: in Section~\ref{sec:method_repre_operation_patterns} we advocate using the joint distribution of states and actions to represent the control mechanisms; this approach differs from related work as other algorithms typically use the states only. In order to study the effect of learning from the joint state and action distributions, we remove the action space from PVED whilst keeping the other components fixed. {\bf(2) PVED without the individual subspace (refered to as PVED{\it  -shared subspace only})}: as stated in Section~\ref{sec:method_repre_operation_patterns}, we expect that the explicit modeling of the agent-specific factors in the individual subspace increases the model's ability to recognize  the shared factors. We justify this by examining the performance of PVED with the individual subspace removed and the other components fixed. {\bf(3) PVED with a deterministic shared subspace (referred to as PVED{\it -det shared subspace})}: we study the benefits of adopting a stochastic model for the shared subspaces against the simpler alternative of deterministic subspaces; accordingly, we change the shared subspace in PVED to a real value vector with the other components of the model fixed, and change the KL divergence in Eq.~\ref{eqn:10} to Euclidean distance for skill transfer. {\bf(4) PVED with traditional VAE posterior approximation (referred to as PVED{\it -VAE})}: as stated in Section~\ref{sec:method_model}, PVED approximates the true posterior distribution of one agent from a different agent, in order to better seek more transferable subspaces; this is different from traditional VAE models that approximates true posterior distribution of one agent from itself; to study the effect of this difference, we derive the models in Eq.~\ref{eqn:2}, Eq.~\ref{eqn:6}, Eq.~\ref{eqn:7}, Eq.~\ref{eqn:8} from traditional VAE approximations. Furthermore, we compare PVED against a baseline (PPO without any skill transfer) and a state-of-the-art approach \cite{InvariantSubspaceLearning_Gupta_2017}, which has been shown to be superior to other projection methods for skill transfer with MDAs.

\begin{figure*}[t!]
\begin{center}
\subfigure[4-link reacher button pushing]
{
\includegraphics[width=0.85\columnwidth]{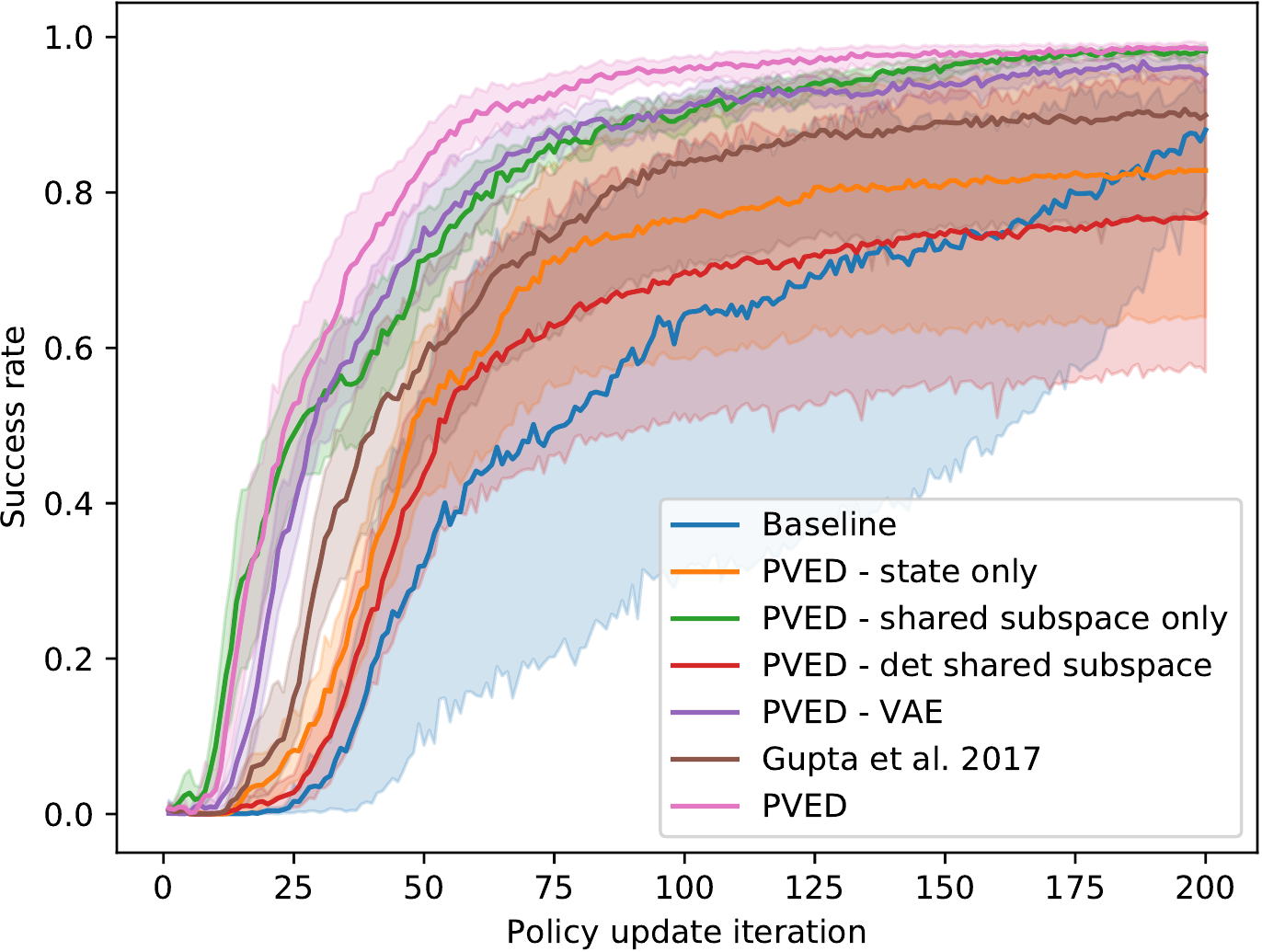}
\label{fig_rst_cmp_button_pushing}
}
\subfigure[Fetch robot peg insertion]
{
\includegraphics[width=0.85\columnwidth]{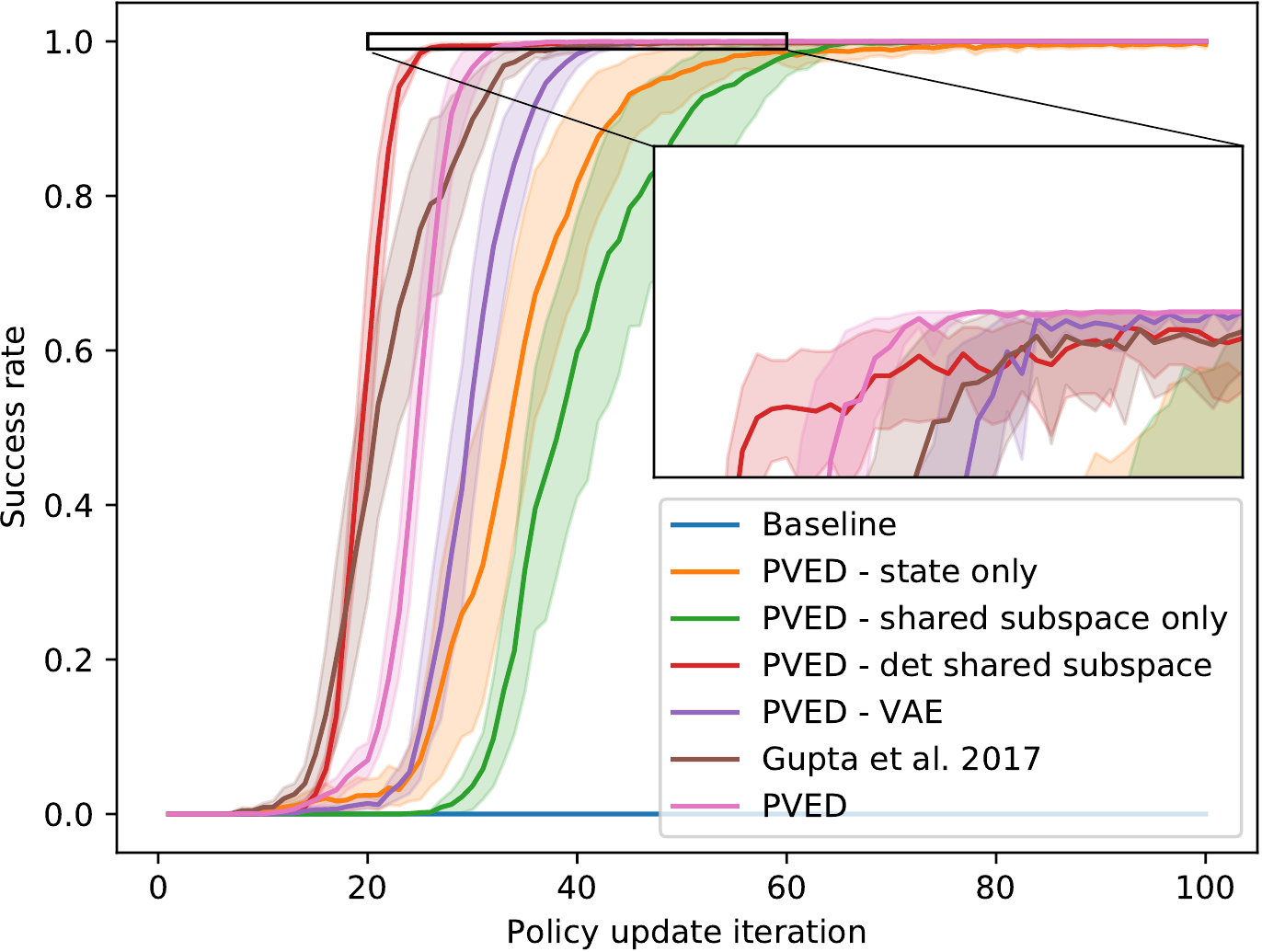}
\label{fig_rst_cmp_gf4dof_peg_insert}
}
\subfigure[Hopper curved-plane climbing]
{
\includegraphics[width=0.85\columnwidth]{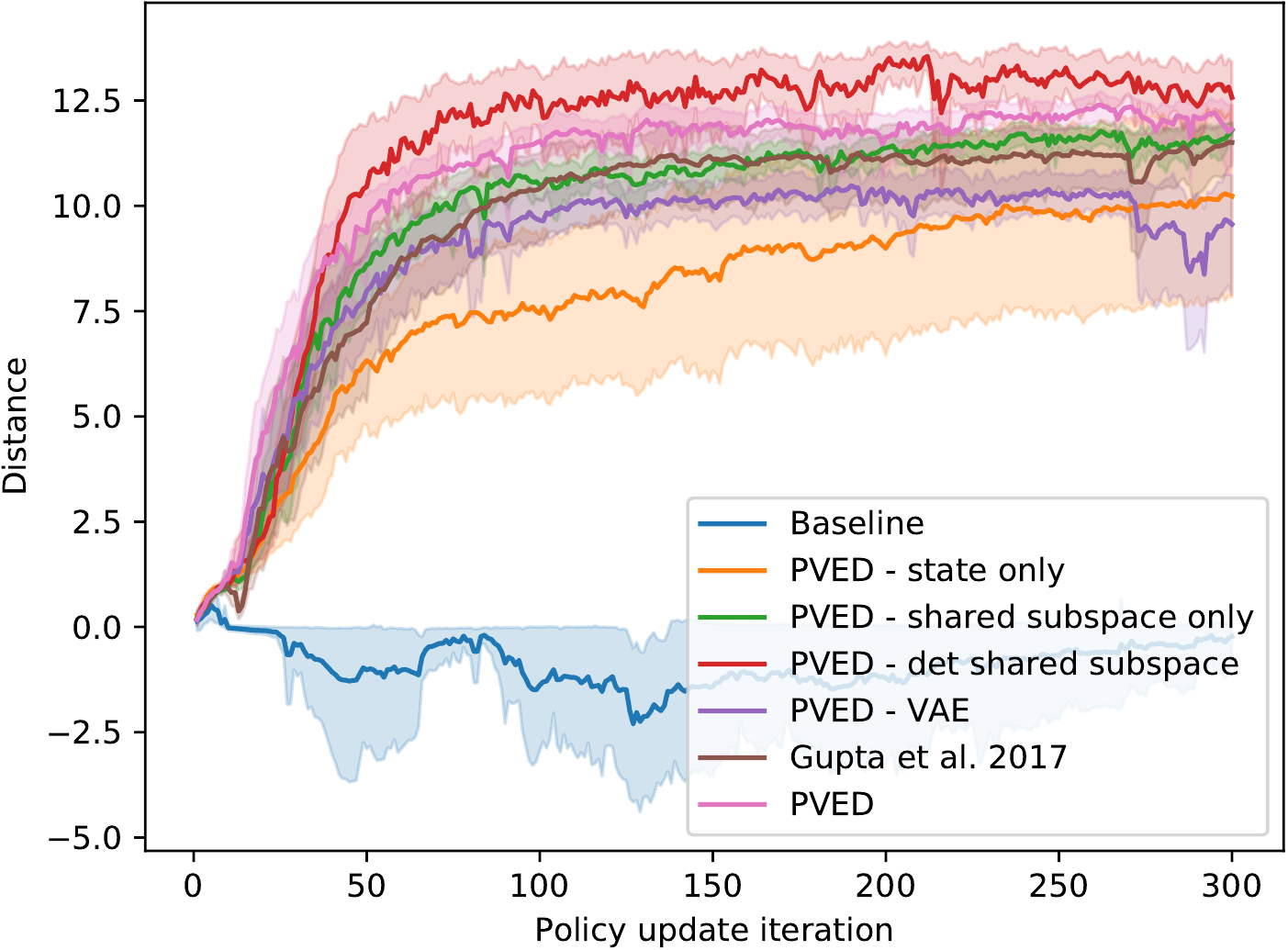}
\label{fig_rst_cmp_hopper_curve_plane}
}
\subfigure[Ant curved-plane climbing]
{
\includegraphics[width=0.85\columnwidth]{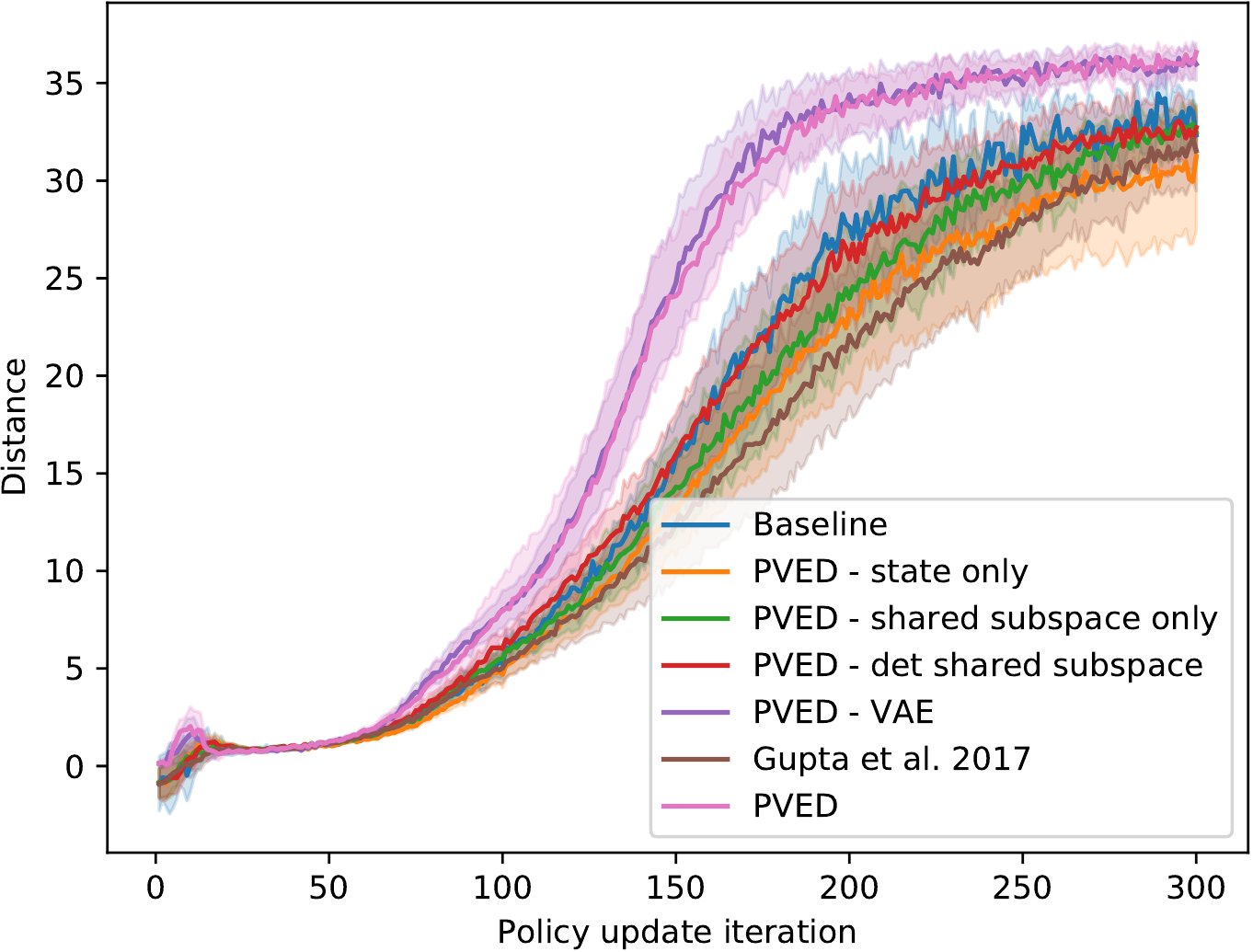}
\label{fig_rst_cmp_ant_curve_plane}
}
\caption{Performance of comparison algorithms on the target agent on the target tasks.}
\label{fig_experi_cmp}
\end{center}
\end{figure*}

Fig.~\ref{fig_experi_cmp} reports the performance of all the algorithms on the four target tasks. Several observations can be made here. First, as expected, we find that PVED achieves better performance compared to the baseline. Notably, Fig.~\ref{fig_rst_cmp_button_pushing} shows that the baseline is still able to learn the target task for an easier task (4-link reacher button pushing), but fails completely for the more challenging tasks in the second and third environments (Fetch robot peg insertion and Hopper curved-plane climbing tasks) as seen in Fig.~\ref{fig_rst_cmp_gf4dof_peg_insert} and Fig.~\ref{fig_rst_cmp_hopper_curve_plane}. In contrast, the benefits of skill transfer through PVED can be seen in all tasks, either with sparse/partial reward (Fig.~\ref{fig_rst_cmp_button_pushing}, Fig.~\ref{fig_rst_cmp_gf4dof_peg_insert}, Fig.~\ref{fig_rst_cmp_hopper_curve_plane}) or with complete reward (Fig.~\ref{fig_rst_cmp_ant_curve_plane}). Second, PVED achieves improved performance compared to the method in~\cite{InvariantSubspaceLearning_Gupta_2017} (Gupta et al 2017). Third, as shown in Fig.~\ref{fig_rst_cmp_ant_curve_plane}, only PVED and PVED{\it -VAE} show advantage compared to the baseline with the most complex pair of agents. Only these two methods are able to correctly identify the shared control factors between very complex MHAs.

Moreover, we observe that PVED performs consistently better than PVED{\it -state only} on all tasks. This demonstrates the advantage of learning subspaces from joint state and action distributions. Interestingly, PVED{\it -state only} is observed to perform worse compared to the method in~\cite{InvariantSubspaceLearning_Gupta_2017}. A possible reason is that PVED may not work properly with states only. With the stochastic subspaces, Eq.~\ref{eqn:7} may promote the encoders to map some information to zeros, while the reconstruction loss can still be minimized by tweaking the decoders (analogously to a posterior collapse). When this happens, information loss lowers the performance. Adding actions could alleviate this issue, as it increase the difficulty of the reconstruction from random latent prior. In addition, PVED outperforms PVED{\it -shared subspace only} on all tasks, which supports our belief that explicitly modelling agent-specific factors can improve the estimation of the shared factors.

We have observed that the comparative performance between PVED and PVED{\it -det shared subspace} varies on different environments. PVED performs consistently better on the Reacher button pushing and the Ant curved-plane climbing tasks. On the Fetch robot peg insertion task, PVED{\it -det shared subspace} learns faster initially, but PVED catches up and learns faster when the performance is near-optimal. In the Hopper curved-plane climbing task, PVED{\it -det shared subspace} performs better. Overall, we observe that PVED achieves competitive performance and is more stable compared to PVED{\it -det shared subspace}. Increased flexibility and stability stem from the stochastic shared subspace model.  

Finally, we have found that PVED outperfoms PVED{\it -VAE} on three out of the four tasks, while the two methods perform similarly on the Ant curved-plane climbing task (Fig.~\ref{fig_rst_cmp_ant_curve_plane}). This observation justifies our assumption that approximating the true posterior distribution of one agent from a different agent better seeks transferable subspaces.

\begin{figure*}[t!]
\begin{center}
\subfigure[t-SNE embedding, Shared subspace]
{
\includegraphics[width=0.75\columnwidth]{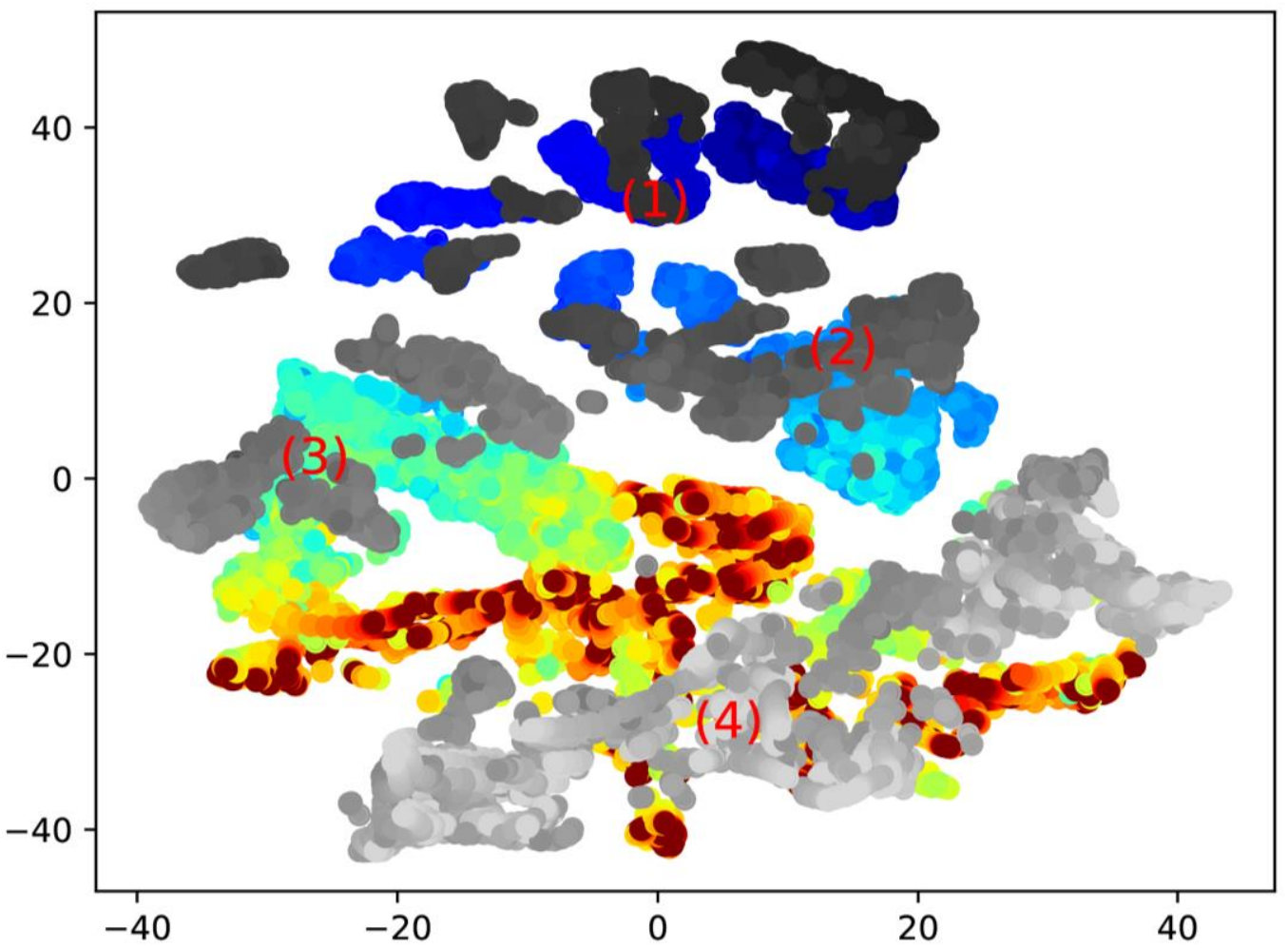}
\label{fig_experi_tsne_button_pushing_common}
}
\subfigure[t-SNE embedding, Individual subspace]
{
\includegraphics[width=0.75\columnwidth]{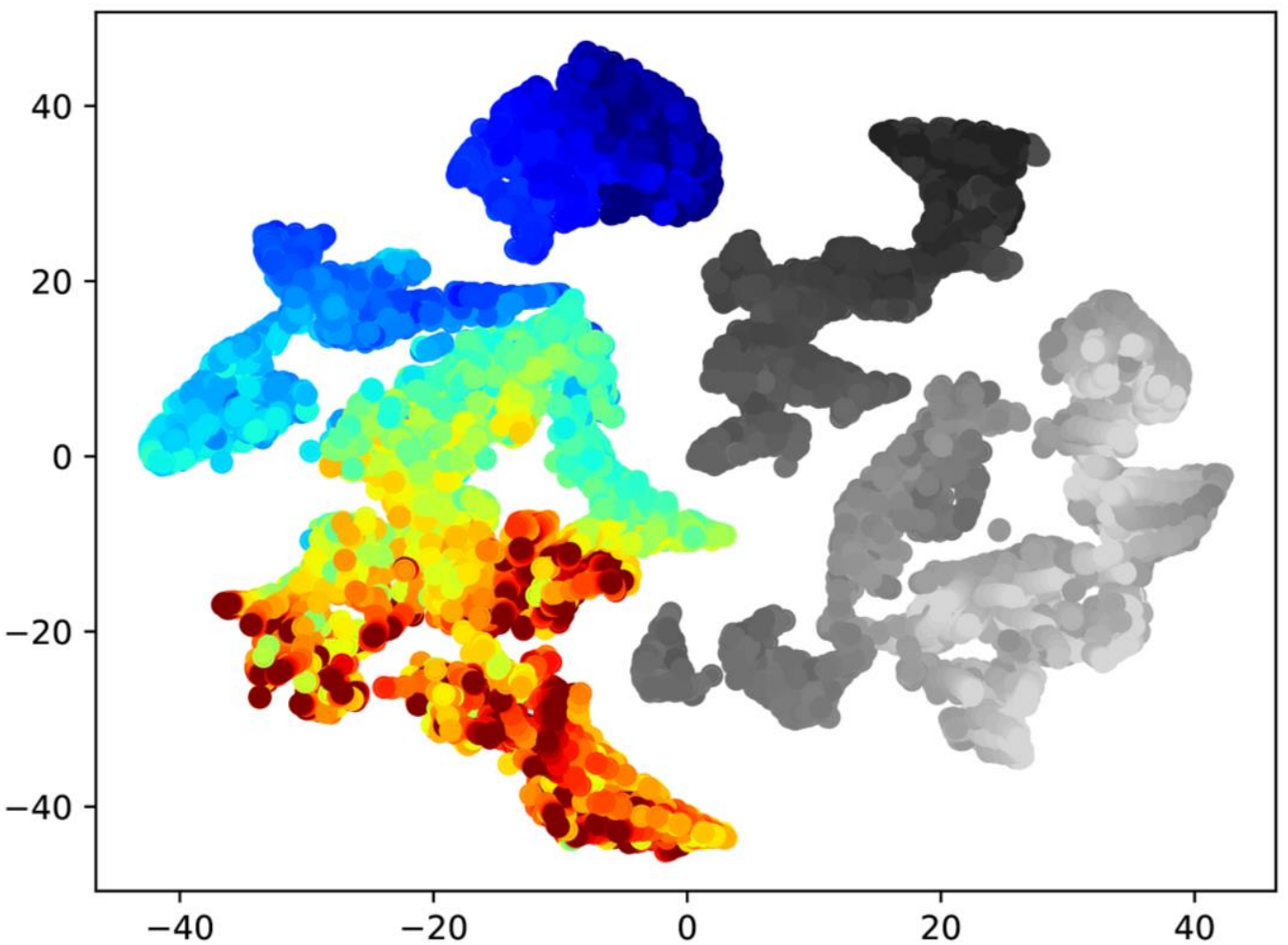}
\label{fig_experi_tsne_button_pushing_specific}
}
\subfigure[Color maps]
{
\includegraphics[width=0.32\columnwidth]{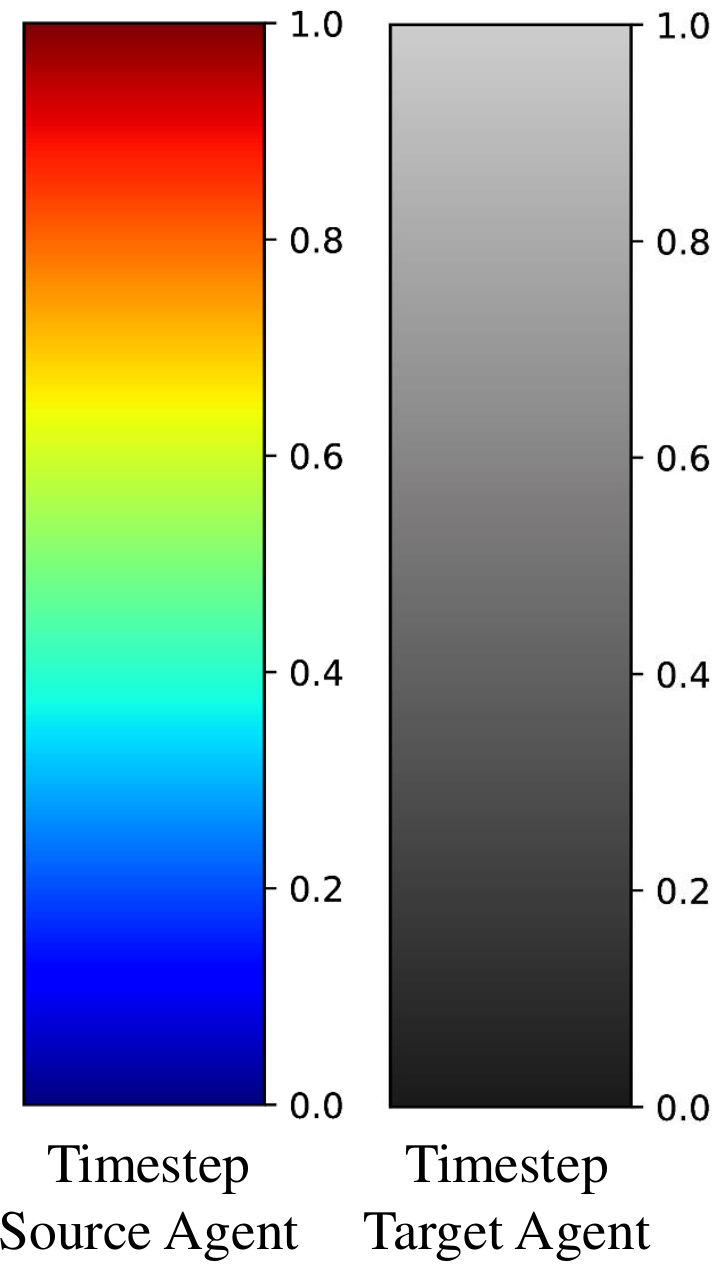}
\label{fig_experi_tsne_button_pushing_color_map}
}
\linebreak
\subfigure[Agents nearest to (1)]
{
\includegraphics[width=0.4\columnwidth]{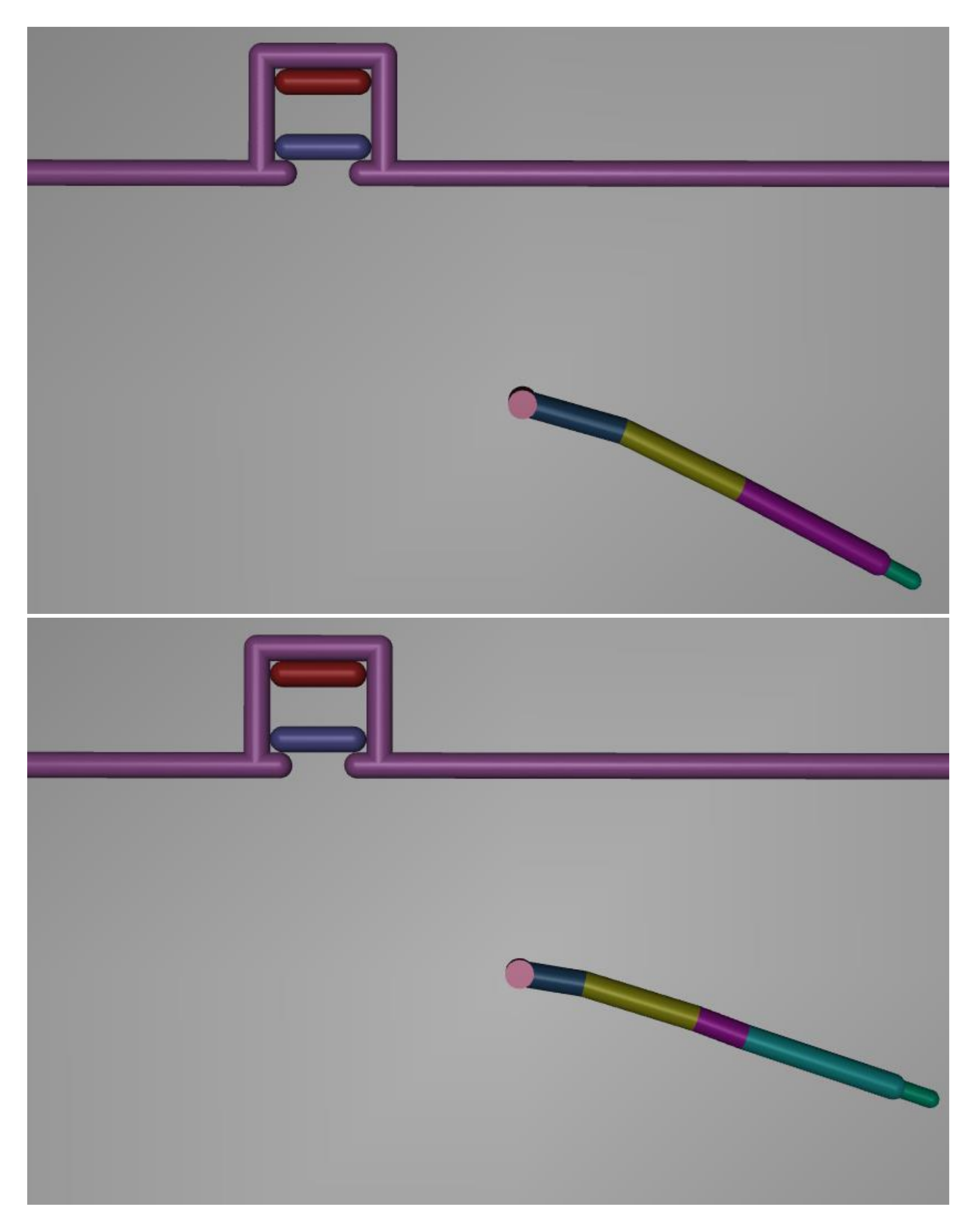}
\label{fig_experi_tsne_button_pushing_common_example_1}
}
\subfigure[Agents nearest to (2)]
{
\includegraphics[width=0.4\columnwidth]{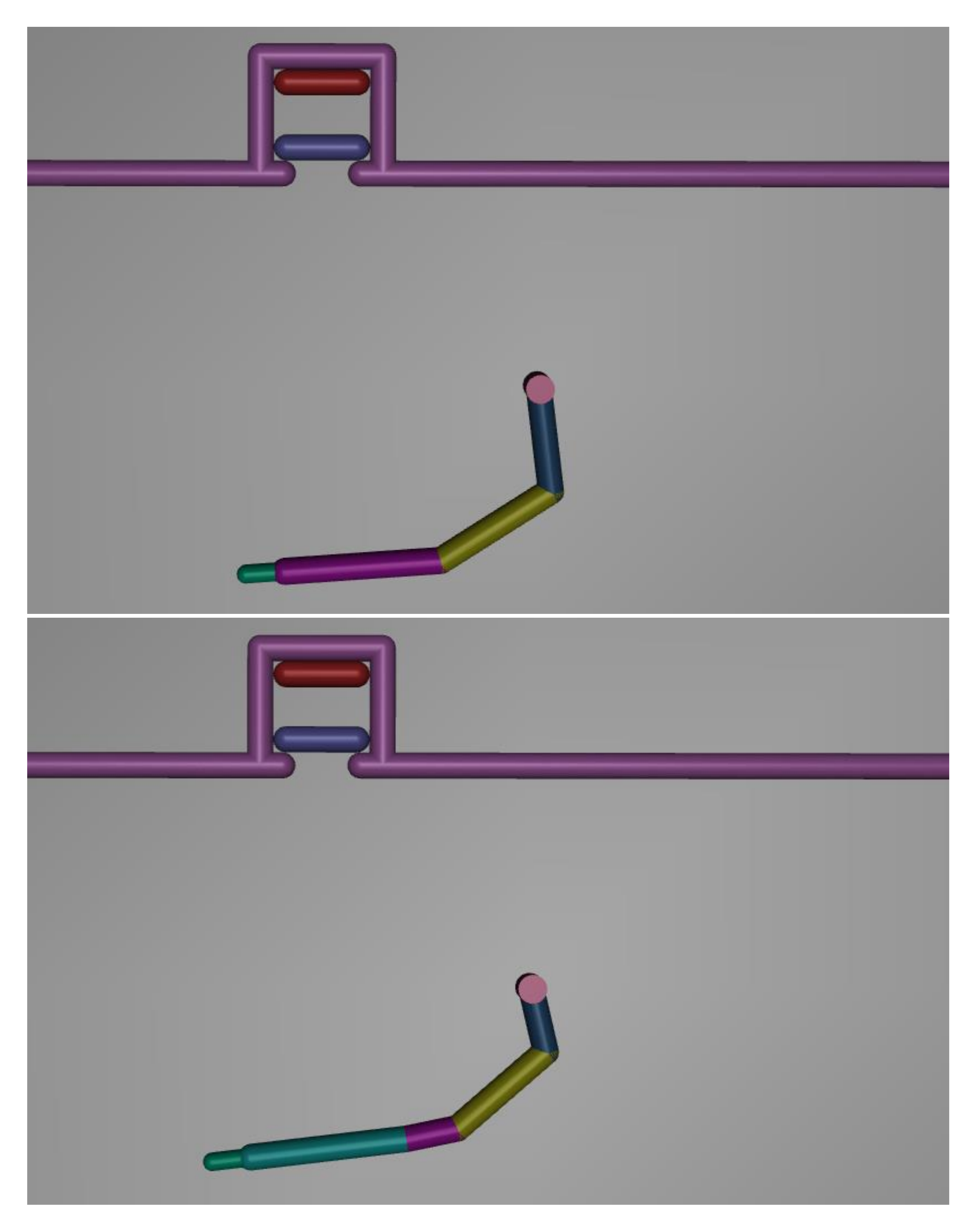}
\label{fig_experi_tsne_button_pushing_common_example_2}
}
\subfigure[Agents nearest to (3)]
{
\includegraphics[width=0.4\columnwidth]{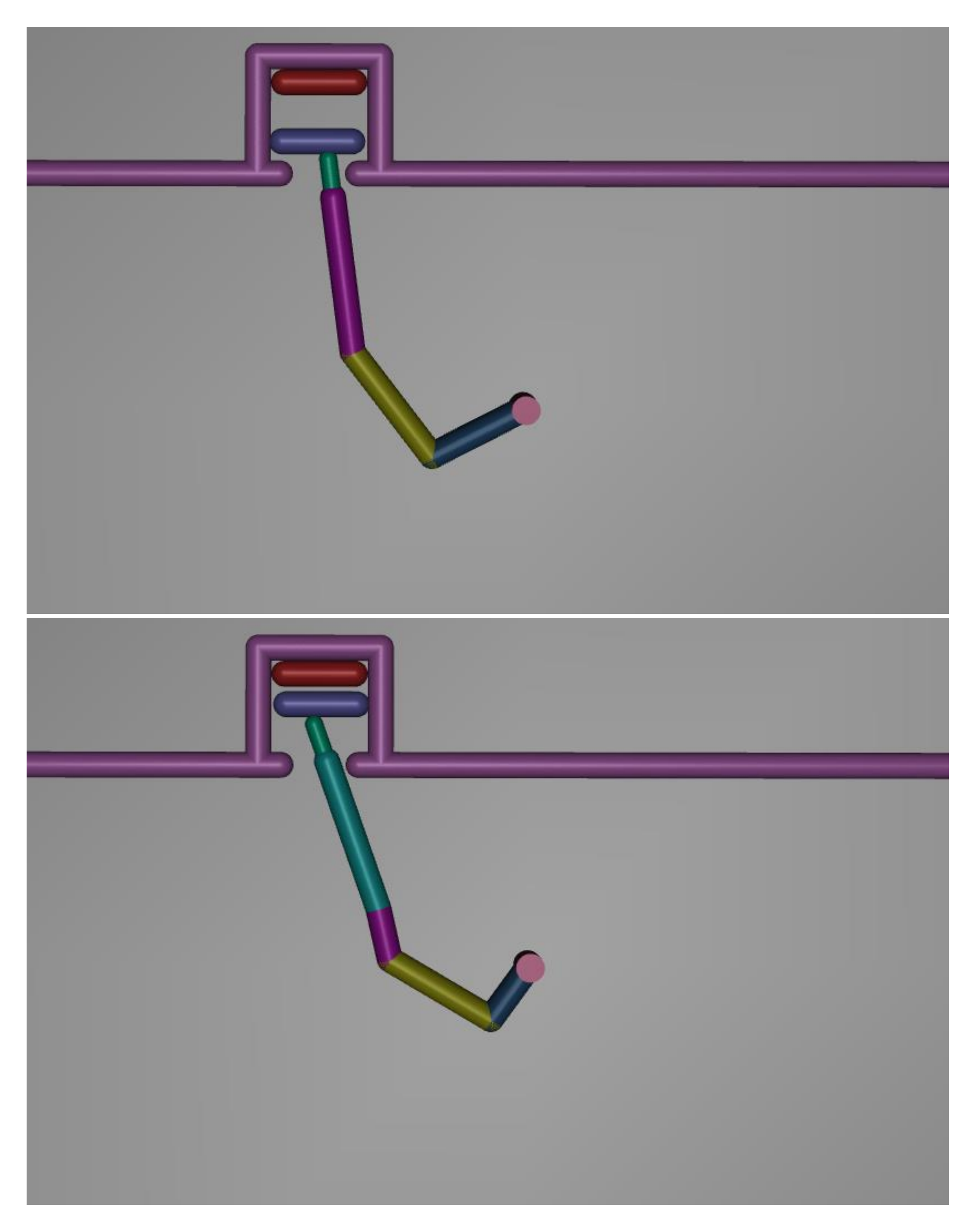}
\label{fig_experi_tsne_button_pushing_common_example_3}
}
\subfigure[Agents nearest to (4)]
{
\includegraphics[width=0.4\columnwidth]{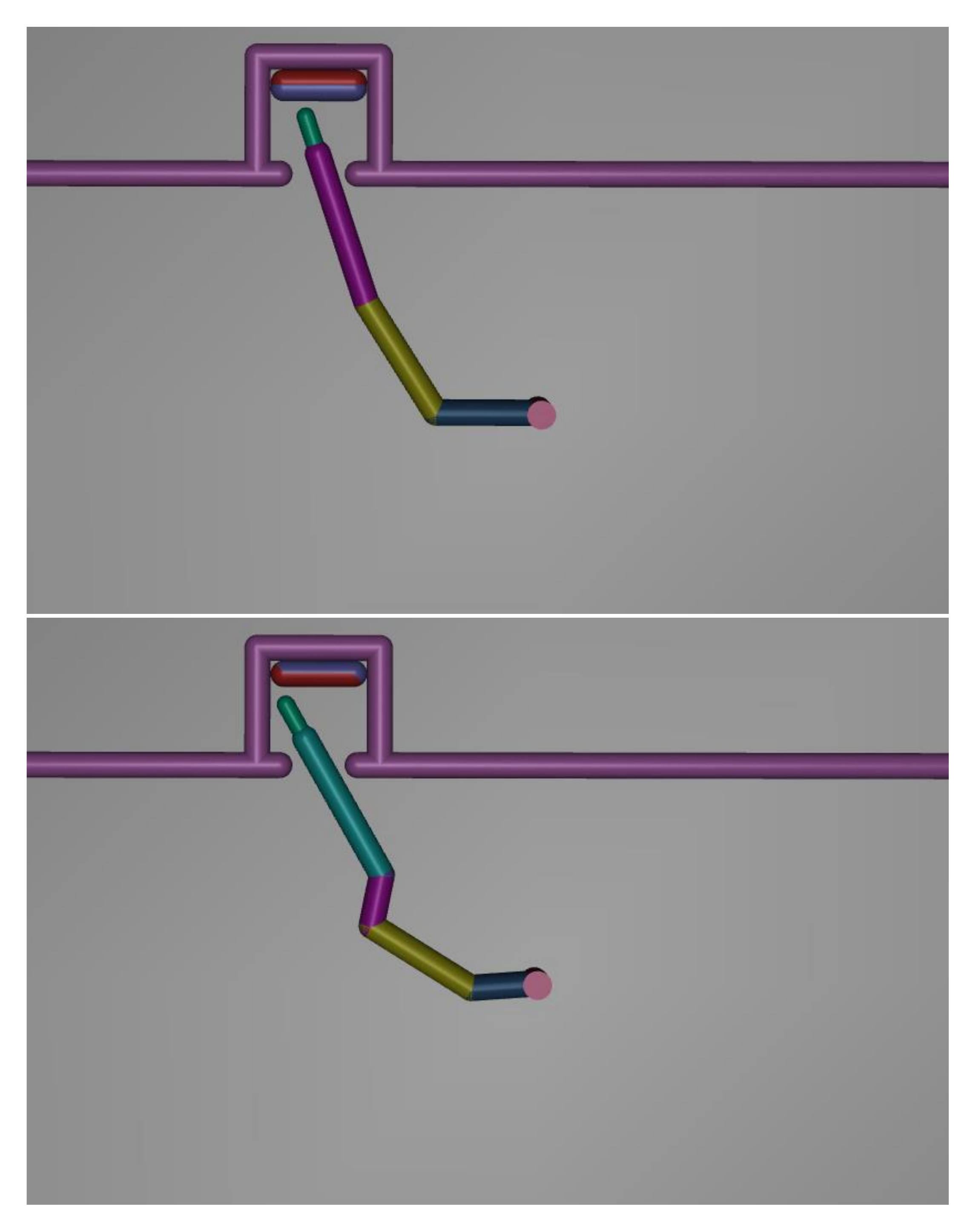}
\label{fig_experi_tsne_button_pushing_common_example_4}
}
\caption{Analysis of t-SNE embeddings on the button pushing tasks with 3-link and 4-link Reachers. (c): The timesteps are normalized to $\left [ 0,1 \right ]$, with different color mappings for each agent; (d) - (g): the top images show the source agent; the bottom ones show the target agent.}
\label{fig_experi_tsne_button_pushing}
\end{center}
\end{figure*}

\begin{figure*}[t!]
\begin{center}
\subfigure[t-SNE embedding, Shared subspace]
{
\includegraphics[width=0.75\columnwidth]{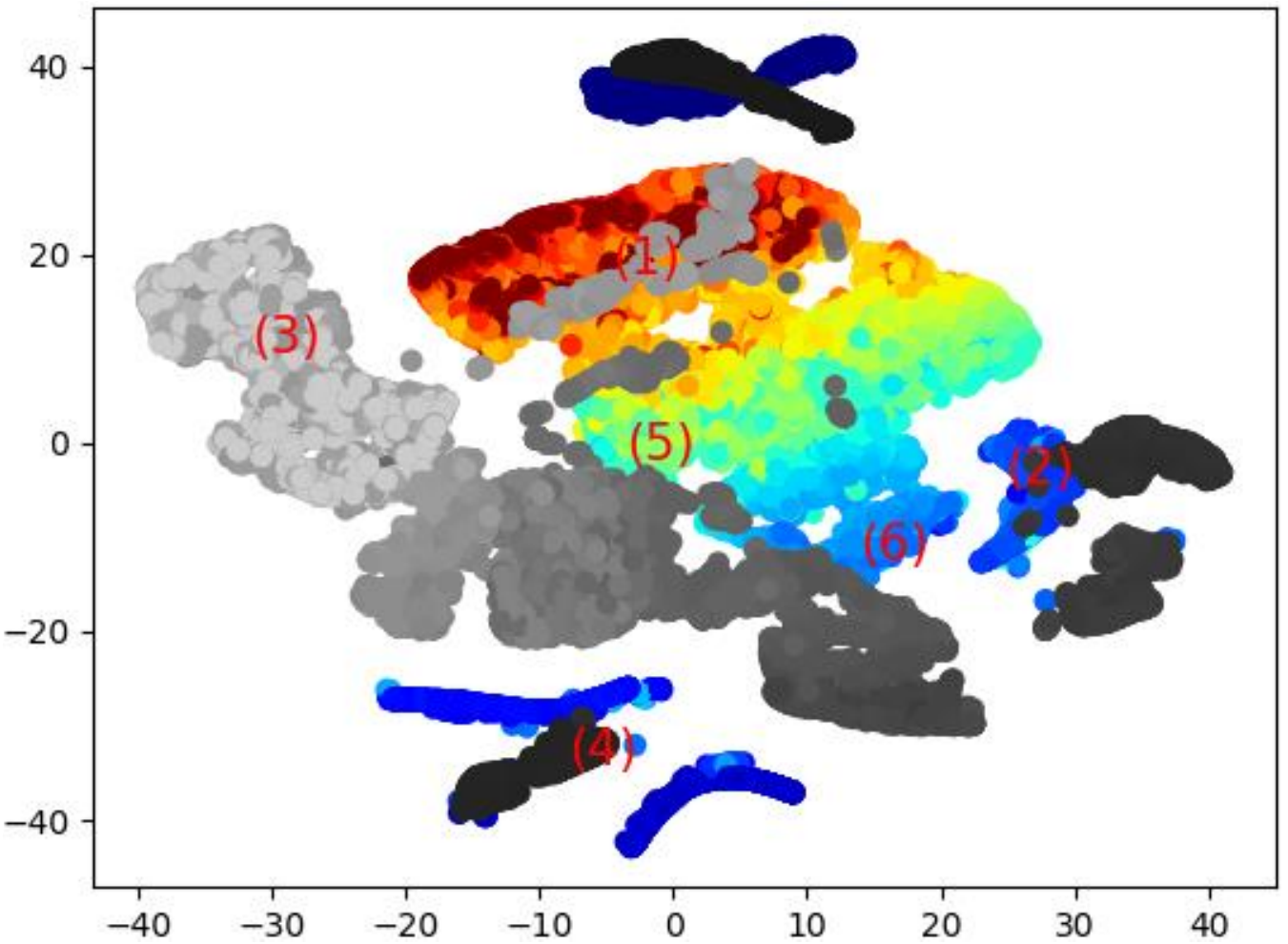}
\label{fig_experi_tsne_peg_insertion_common}
}
\subfigure[t-SNE embedding, Individual subspace]
{
\includegraphics[width=0.75\columnwidth]{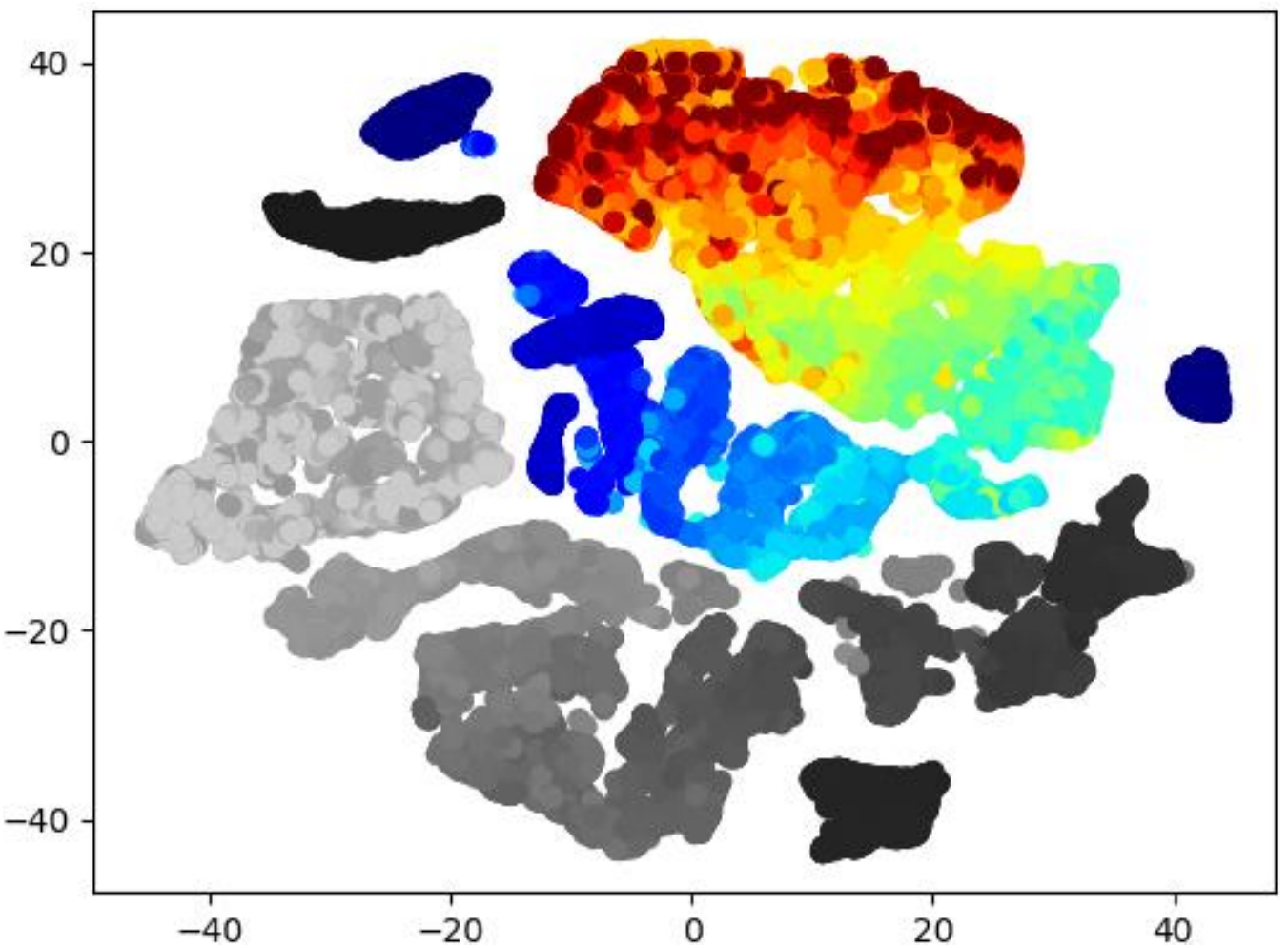}
\label{fig_experi_tsne_peg_insertion_specific}
}
\subfigure[Color maps]
{
\includegraphics[width=0.32\columnwidth]{time_step_bars}
\label{fig_experi_tsne_peg_insertion_color_map}
}
\linebreak
\subfigure[Agents nearest to (1)]
{
\includegraphics[height=34.5mm]{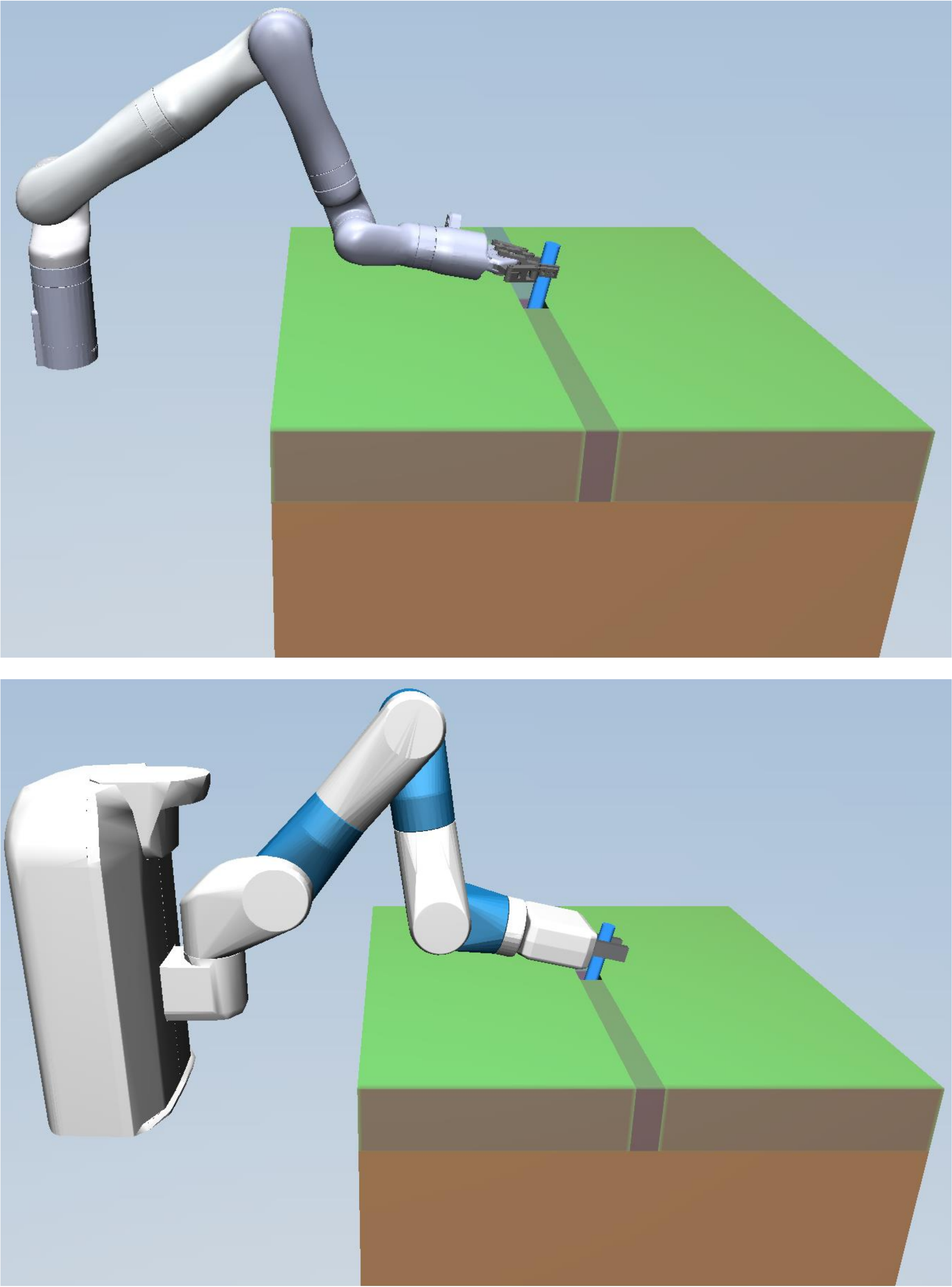}
\label{fig_experi_tsne_peg_insertion_common_example_1}
}
\hspace{-0.12in}
\subfigure[Agents nearest to (2)]
{
\includegraphics[height=34.5mm]{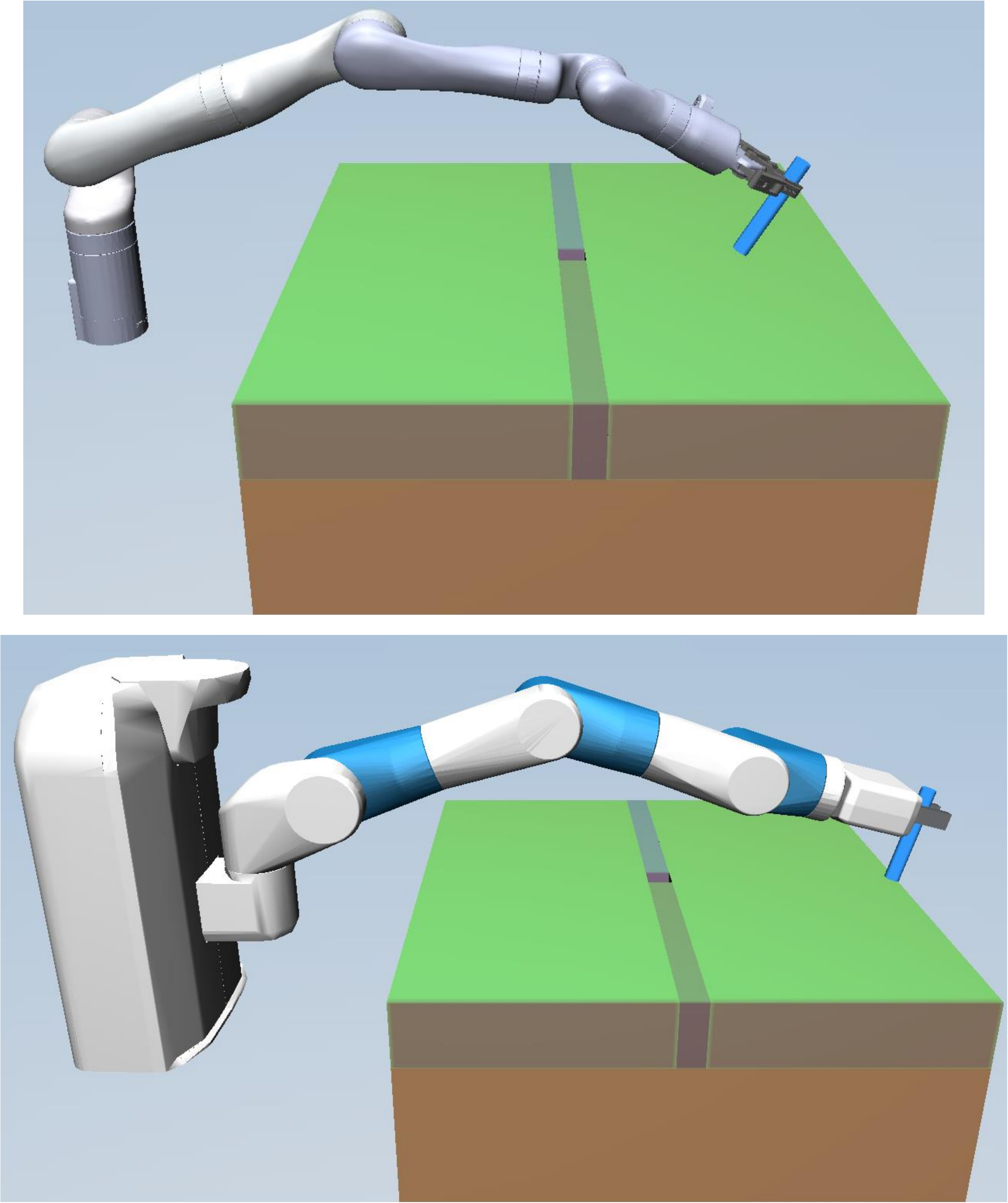}
\label{fig_experi_tsne_peg_insertion_common_example_2}
}
\hspace{-0.12in}
\subfigure[Agents nearest to (3)]
{
\includegraphics[height=34.5mm]{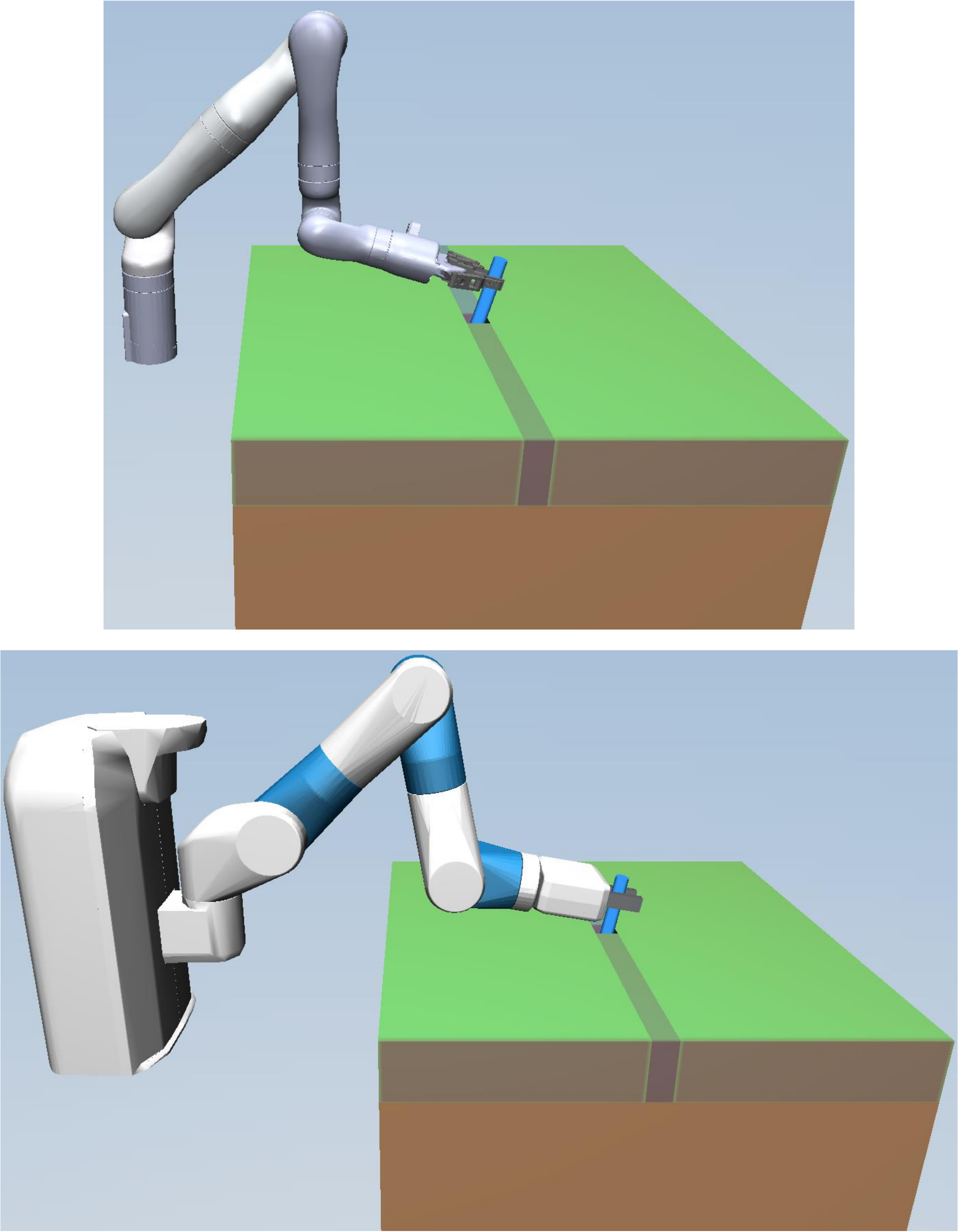}
\label{fig_experi_tsne_peg_insertion_common_example_3}
}
\hspace{-0.12in}
\subfigure[Agents nearest to (4)]
{
\includegraphics[height=34.5mm]{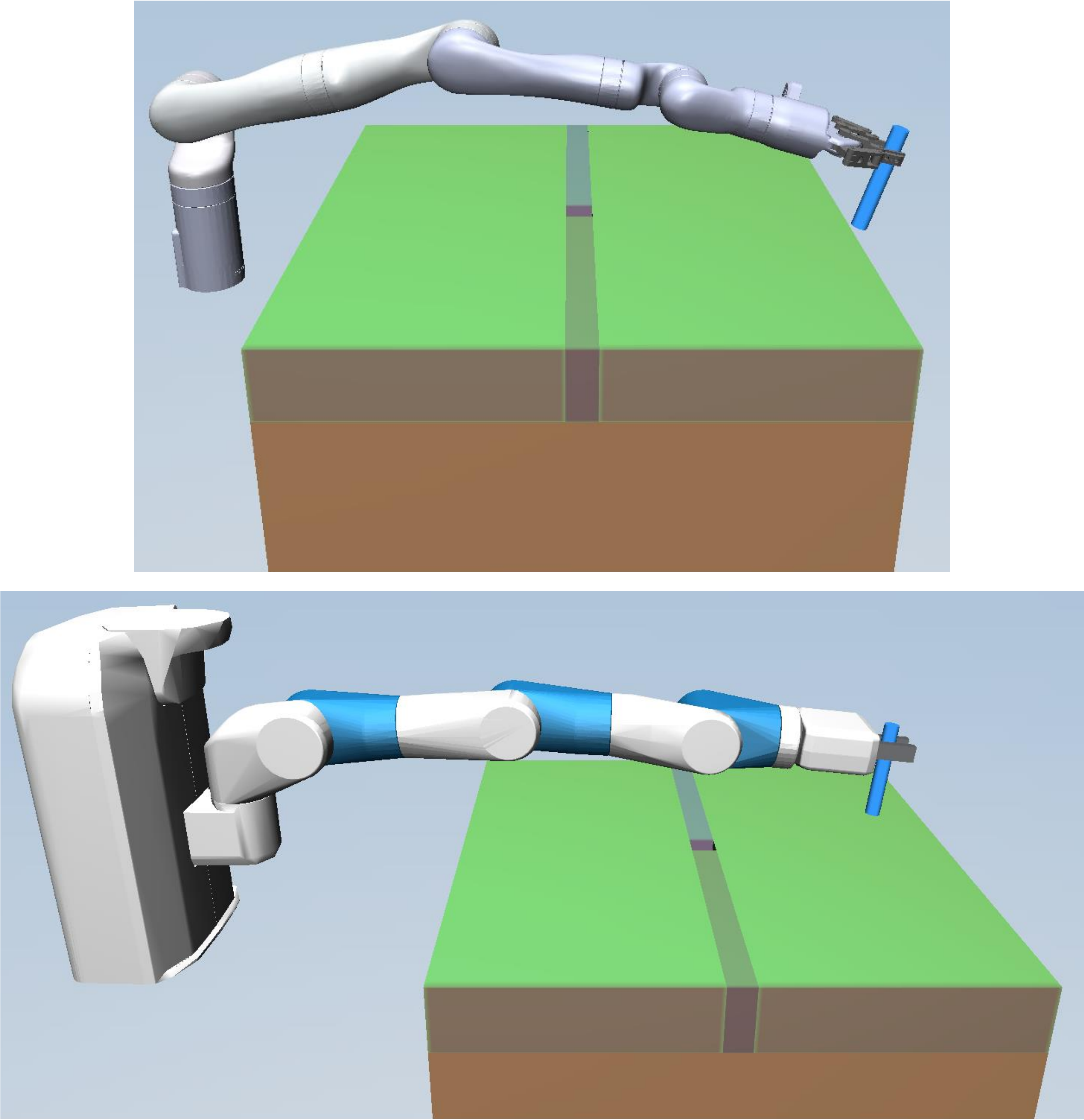}
\label{fig_experi_tsne_peg_insertion_common_example_4}
}
\hspace{-0.12in}
\subfigure[Agents nearest to (5)]
{
\includegraphics[height=34.5mm]{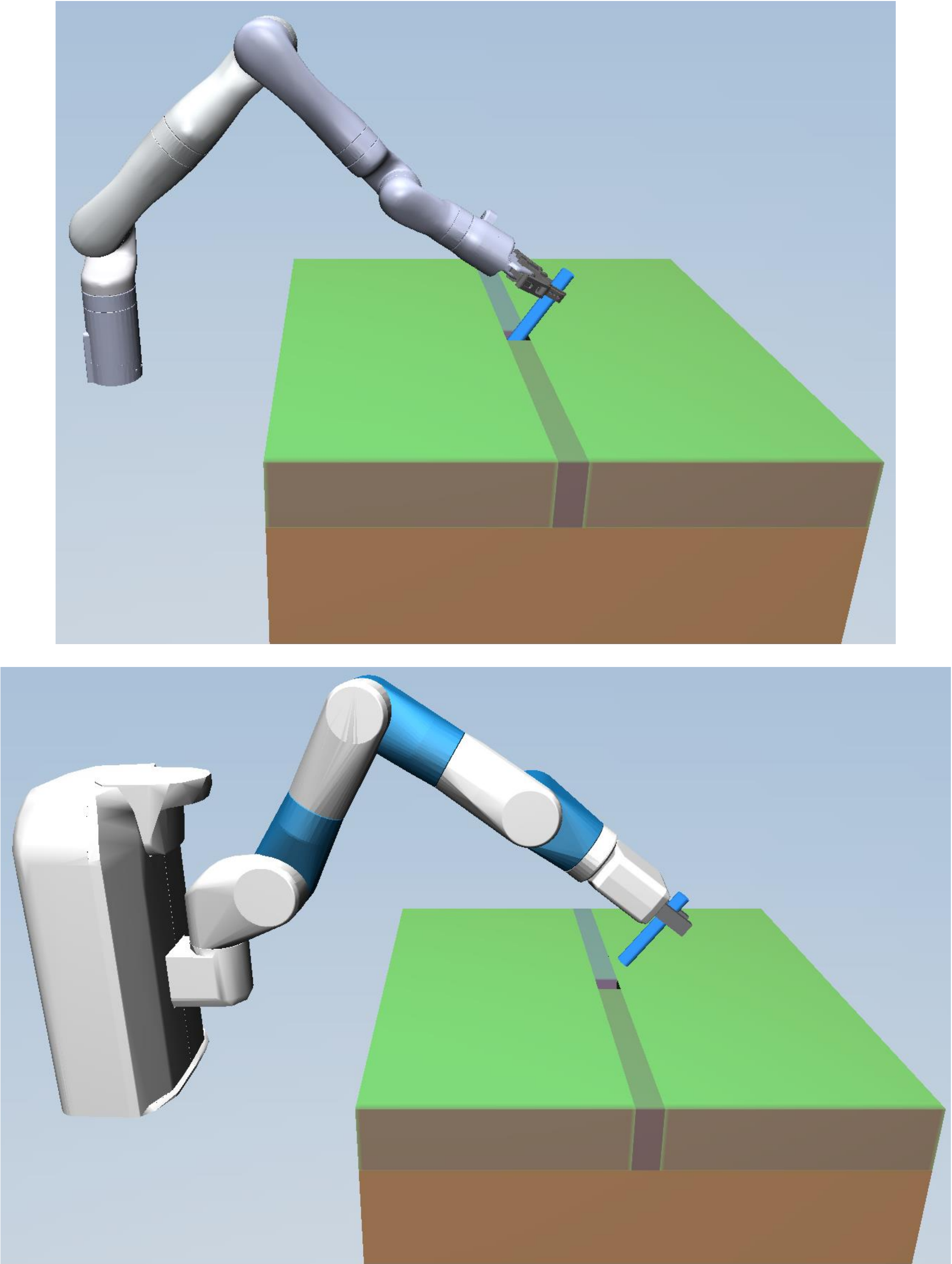}
\label{fig_experi_tsne_peg_insertion_common_example_5}
}
\hspace{-0.12in}
\subfigure[Agents nearest to (6)]
{
\includegraphics[height=34.5mm]{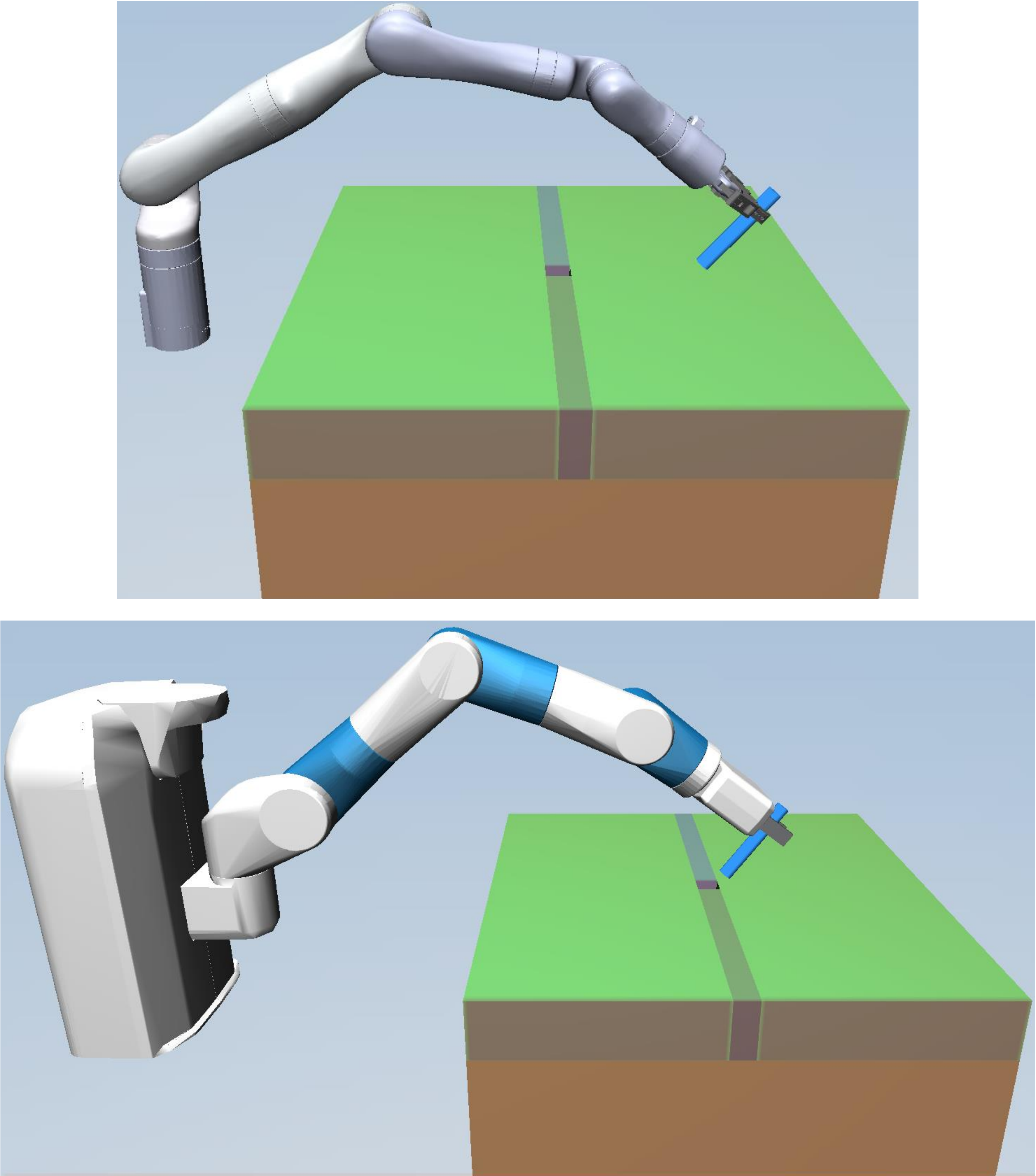}
\label{fig_experi_tsne_peg_insertion_common_example_6}
}
\caption{Analysis of t-SNE embeddings on the peg insertion tasks with the Jaco3 and Fetch robot. (c): The timesteps are normalized to $\left [ 0,1 \right ]$, with different color mappings for each agent; (d) - (i): the top images show the source agent; the bottom ones show the target agent.}
\label{fig_experi_tsne_peg_insertion}
\end{center}
\end{figure*}

\begin{figure*}[t!]
\begin{center}
\subfigure[t-SNE embedding, Shared subspace]
{
\includegraphics[width=0.75\columnwidth]{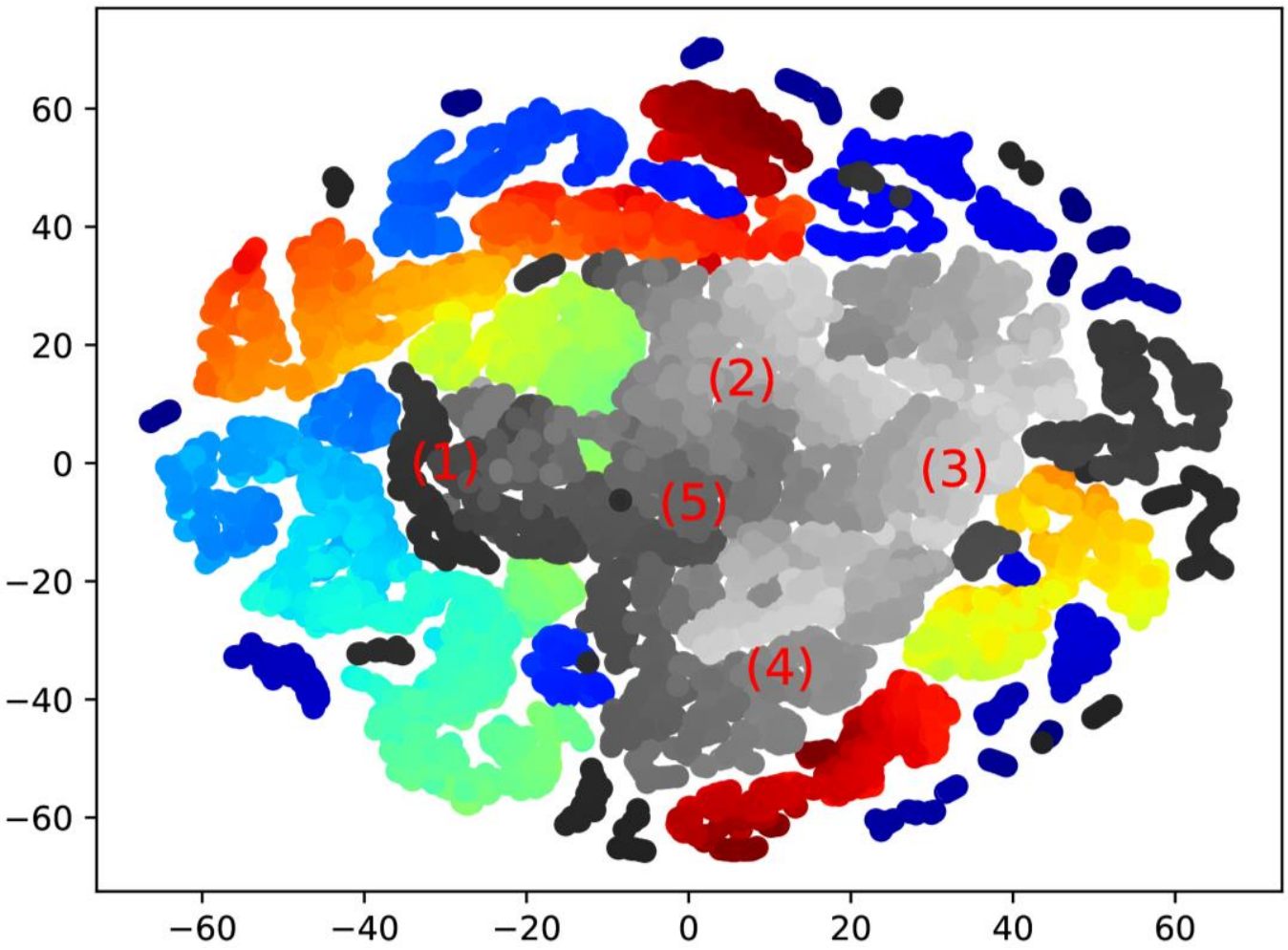}
\label{fig_experi_tsne_curve_plane_common}
}
\subfigure[t-SNE embedding, Individual subspace]
{
\includegraphics[width=0.75\columnwidth]{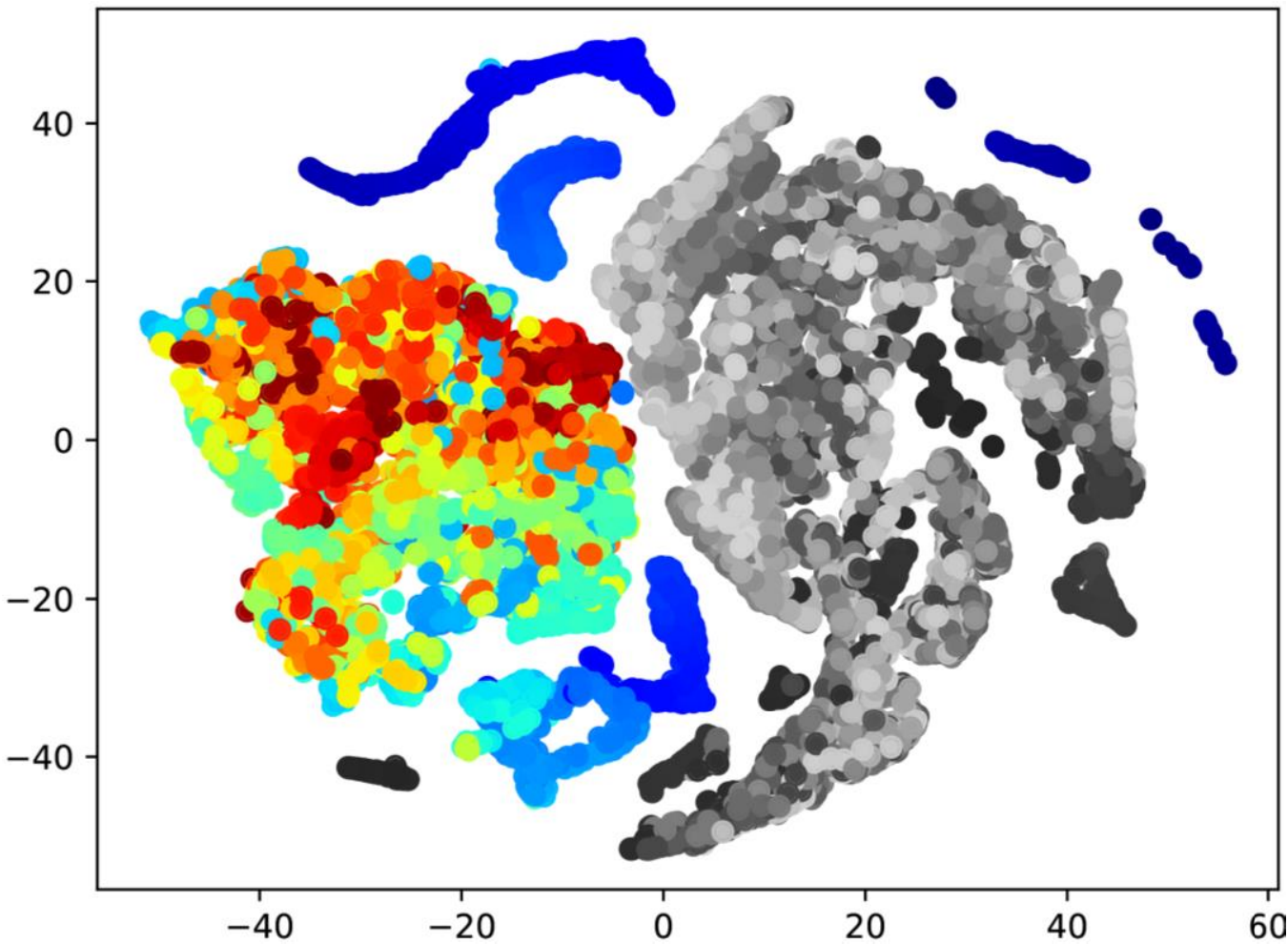}
\label{fig_experi_tsne_curve_plane_specific}
}
\subfigure[Color maps]
{
\includegraphics[width=0.32\columnwidth]{time_step_bars}
\label{fig_experi_tsne_curve_plane_color_map}
}
\linebreak
\subfigure[Agents nearest to (1)]
{
\includegraphics[width=0.37\columnwidth]{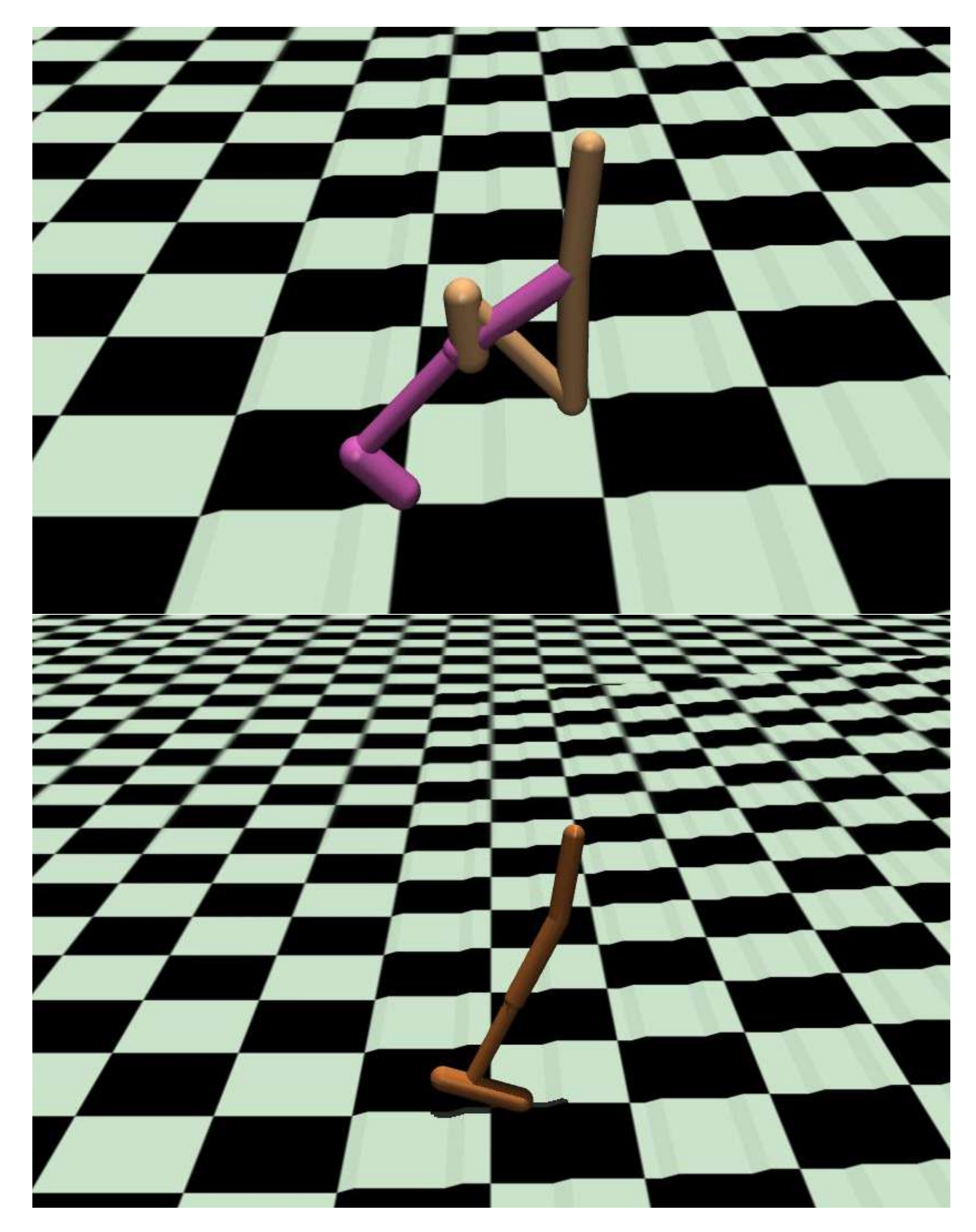}
\label{fig_experi_tsne_curve_plane_common_example_1}
}
\subfigure[Agents nearest to (2)]
{
\includegraphics[width=0.37\columnwidth]{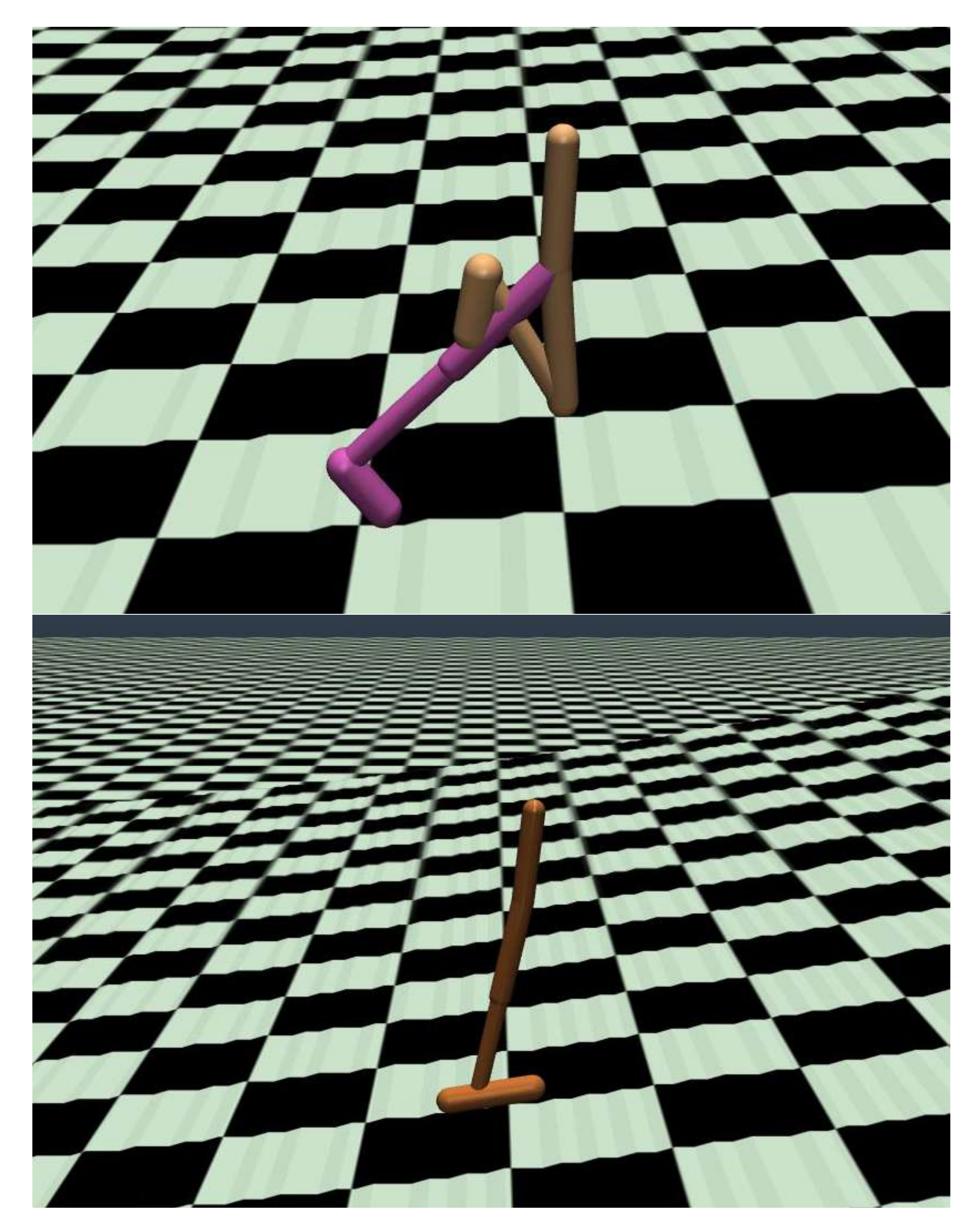}
\label{fig_experi_tsne_curve_plane_common_example_2}
}
\subfigure[Agents nearest to (3)]
{
\includegraphics[width=0.37\columnwidth]{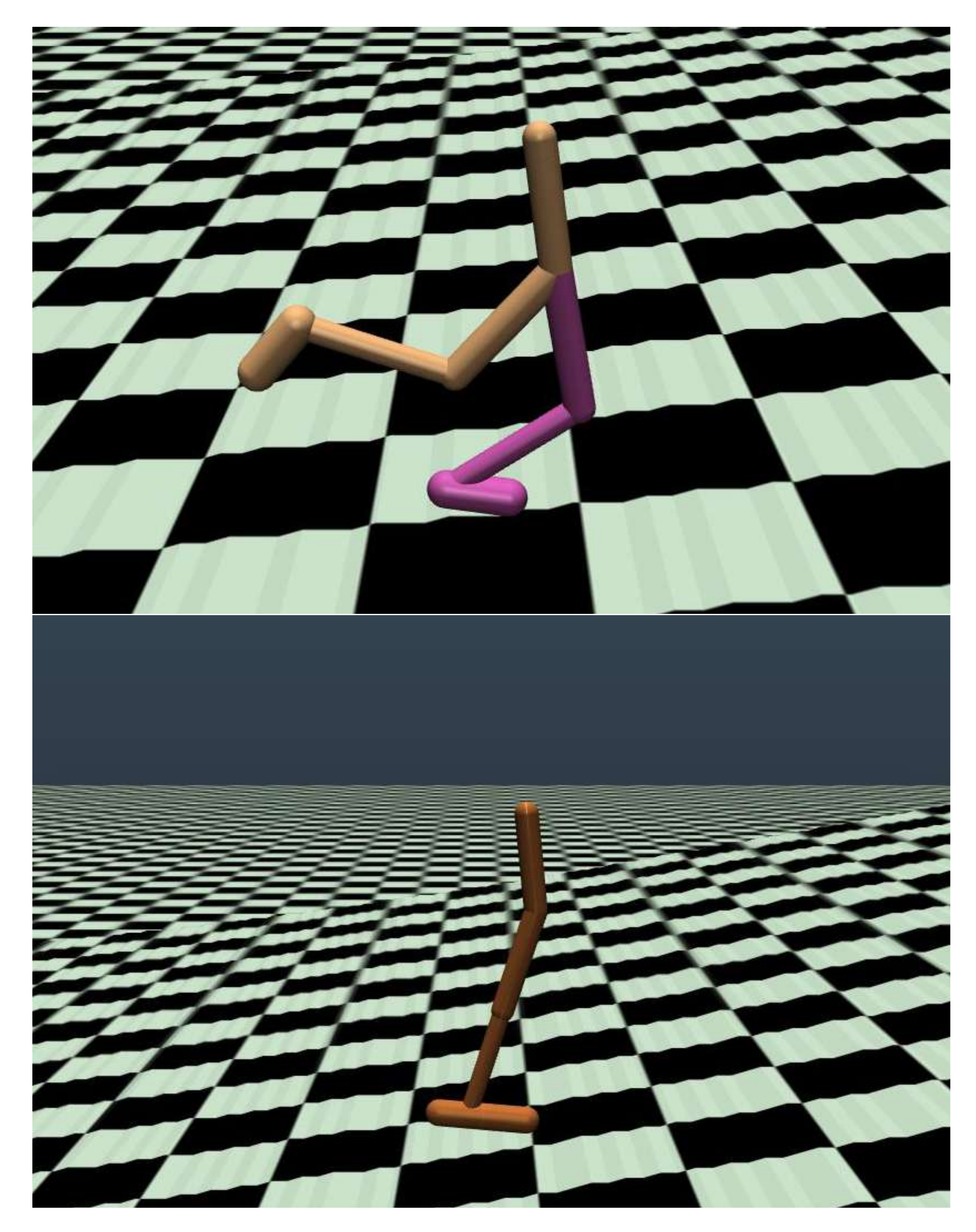}
\label{fig_experi_tsne_curve_plane_common_example_3}
}
\subfigure[Agents nearest to (4)]
{
\includegraphics[width=0.37\columnwidth]{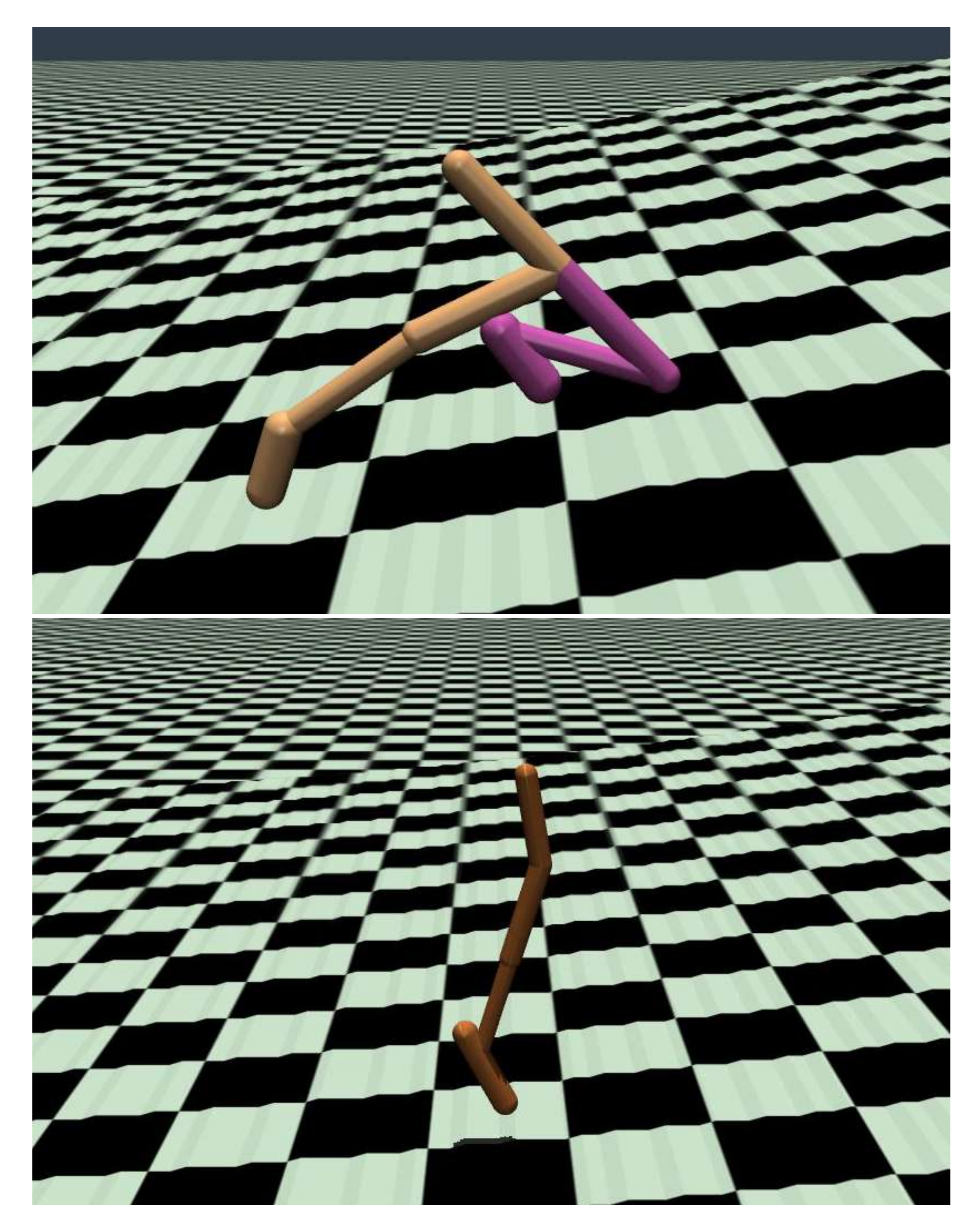}
\label{fig_experi_tsne_curve_plane_common_example_4}
}
\subfigure[Agents nearest to (5)]
{
\includegraphics[width=0.37\columnwidth]{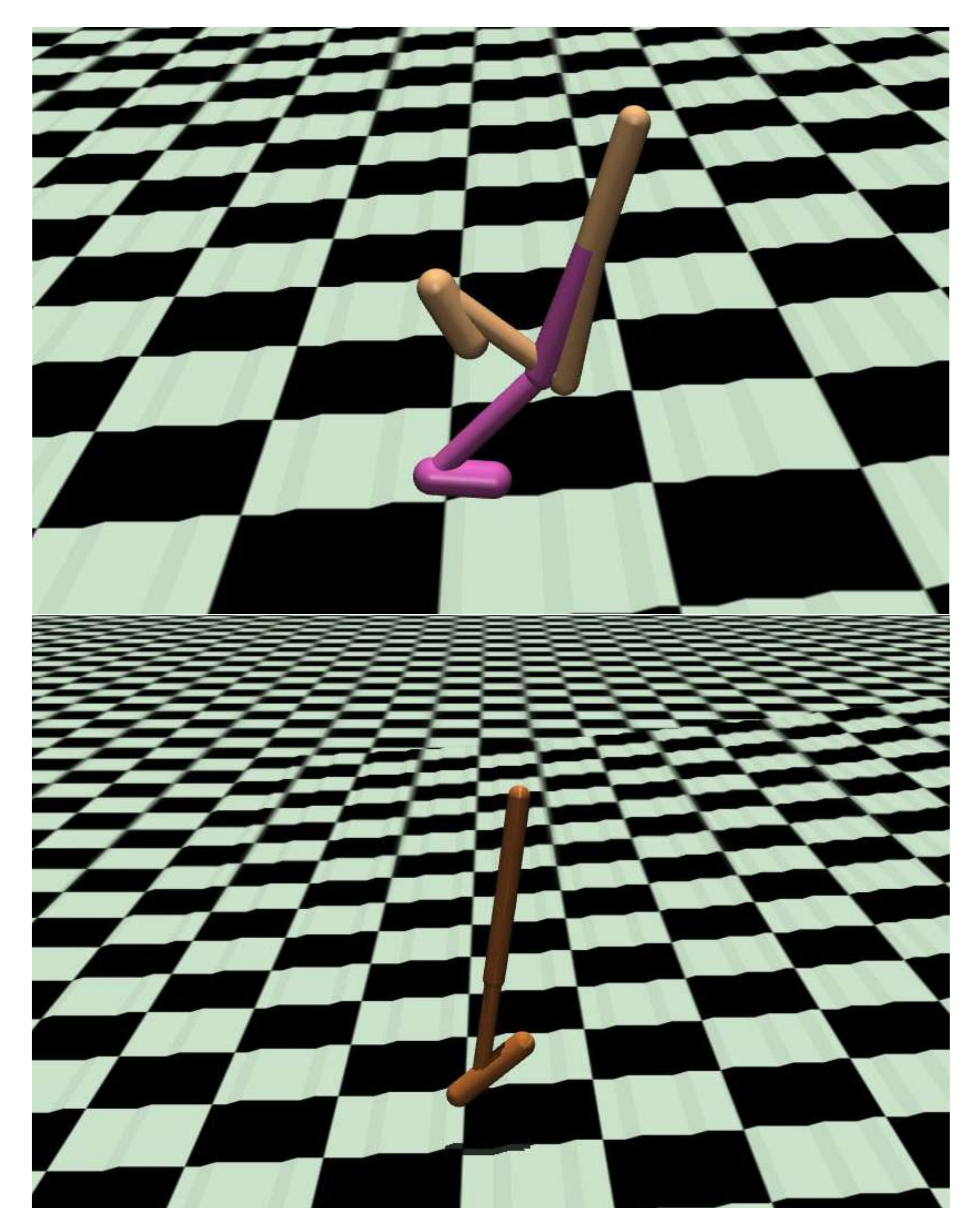}
\label{fig_experi_tsne_curve_plane_common_example_5}
}
\caption{Analysis of t-SNE embeddings on the curved plane climbing tasks with Walker2d and Hopper. (c): The timesteps are normalized to $\left [ 0,1 \right ]$, with different color mappings for each agent; (d) - (h): the top images show the source agent; the bottom ones show the target agent.}
\label{fig_experi_tsne_curve_plane}
\end{center}
\end{figure*}

\begin{figure*}[t!]
\begin{center}
\subfigure[t-SNE embedding, Shared subspace]
{
\includegraphics[width=0.75\columnwidth]{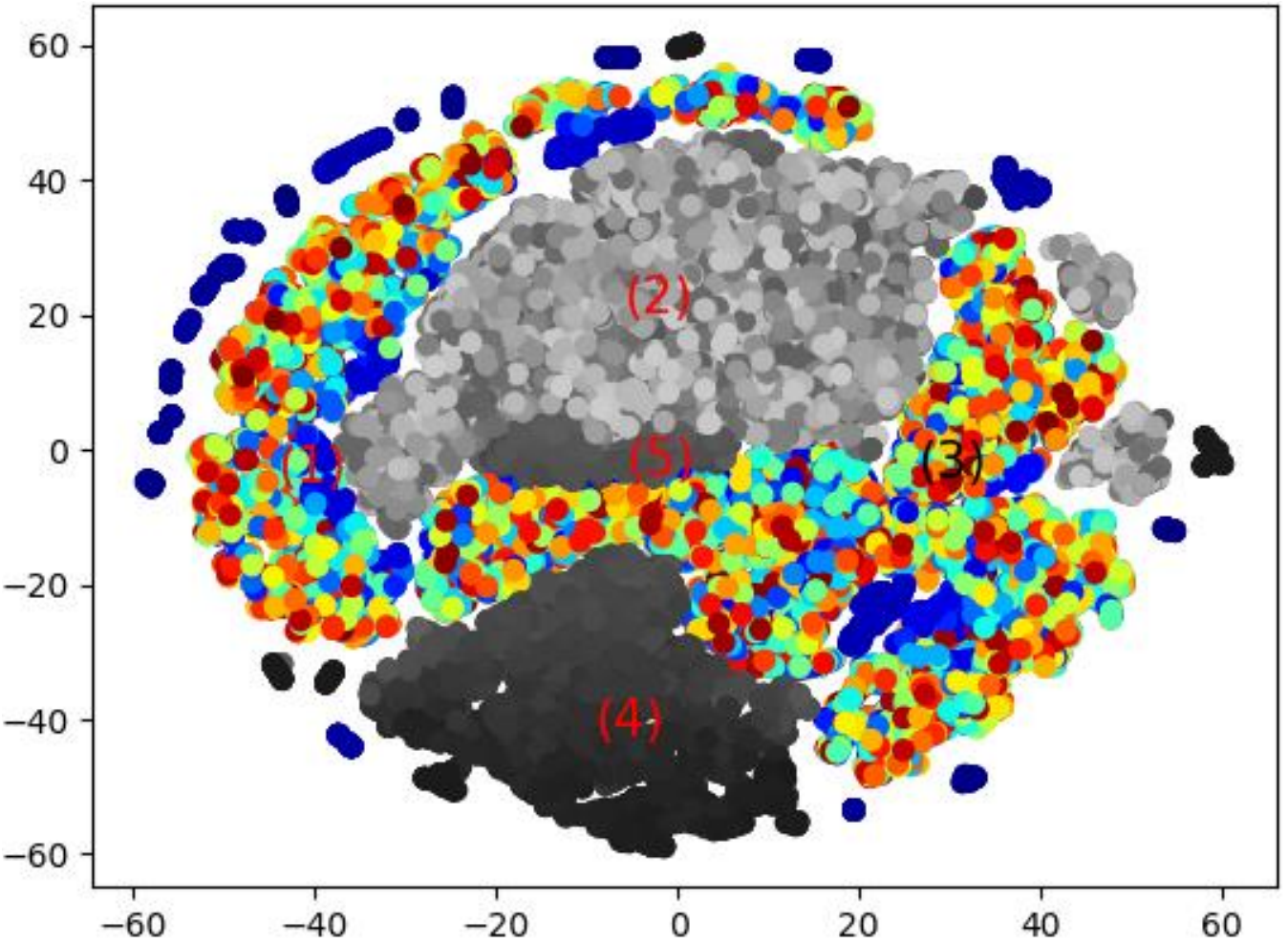}
\label{fig_experi_tsne_ha_curve_plane_common}
}
\subfigure[t-SNE embedding, Individual subspace]
{
\includegraphics[width=0.75\columnwidth]{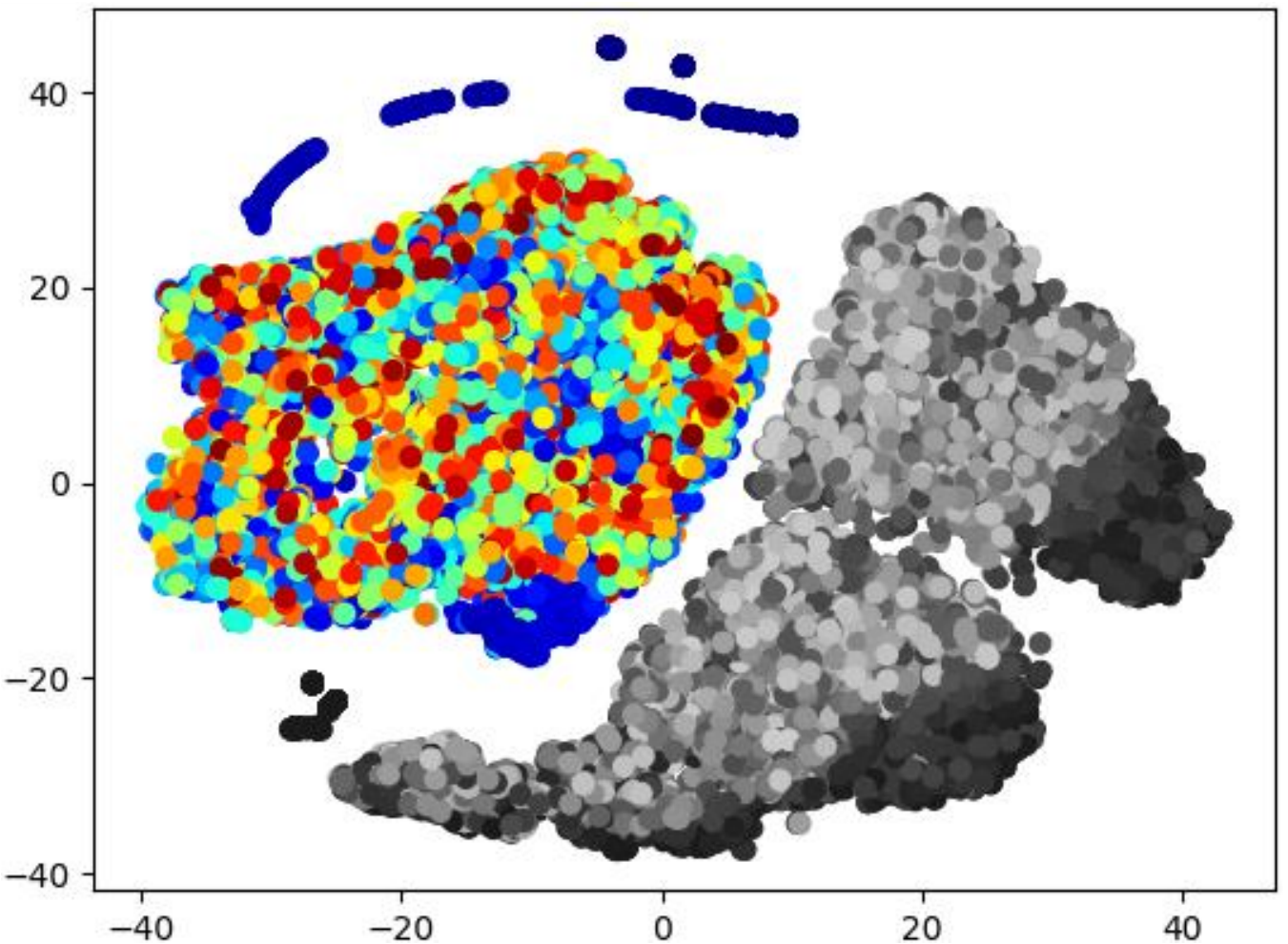}
\label{fig_experi_tsne_ha_curve_plane_specific}
}
\subfigure[Color maps]
{
\includegraphics[width=0.32\columnwidth]{time_step_bars}
\label{fig_experi_tsne_ha_curve_plane_color_map}
}
\linebreak
\subfigure[Agents nearest to (1)]
{
\includegraphics[width=0.37\columnwidth]{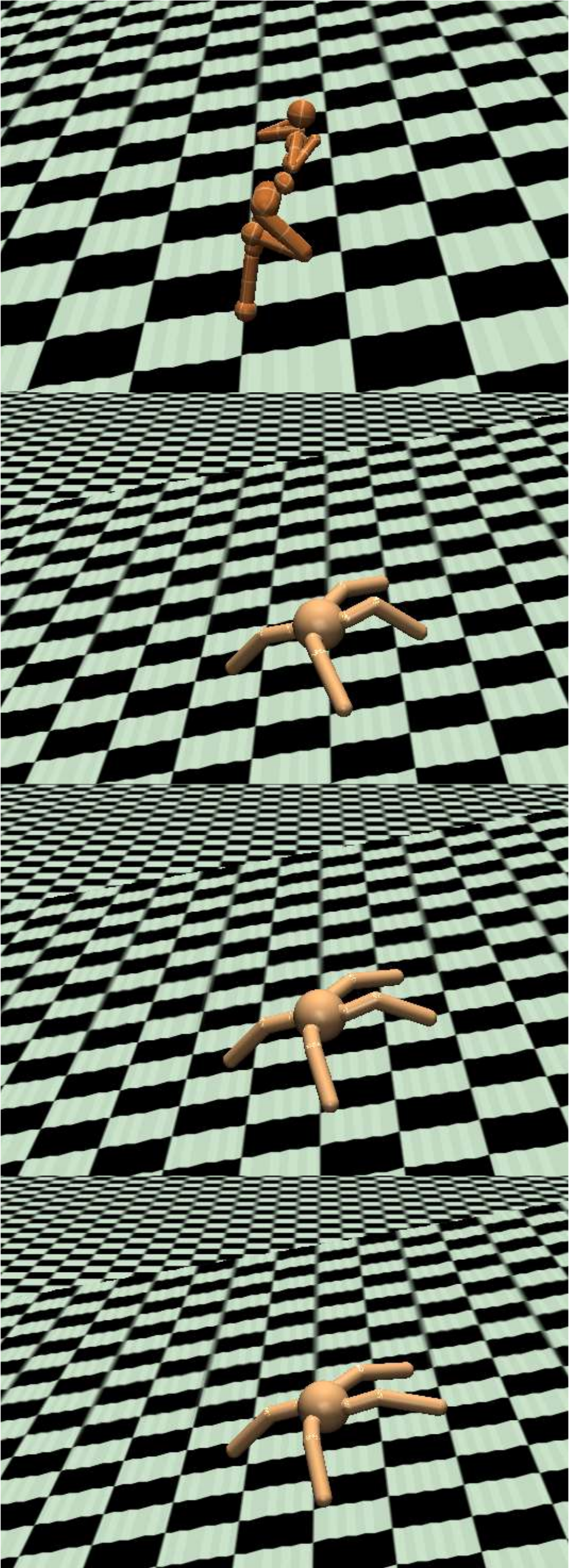}
\label{fig_experi_tsne_ha_curve_plane_common_example_1}
}
\subfigure[Agents nearest to (2)]
{
\includegraphics[width=0.37\columnwidth]{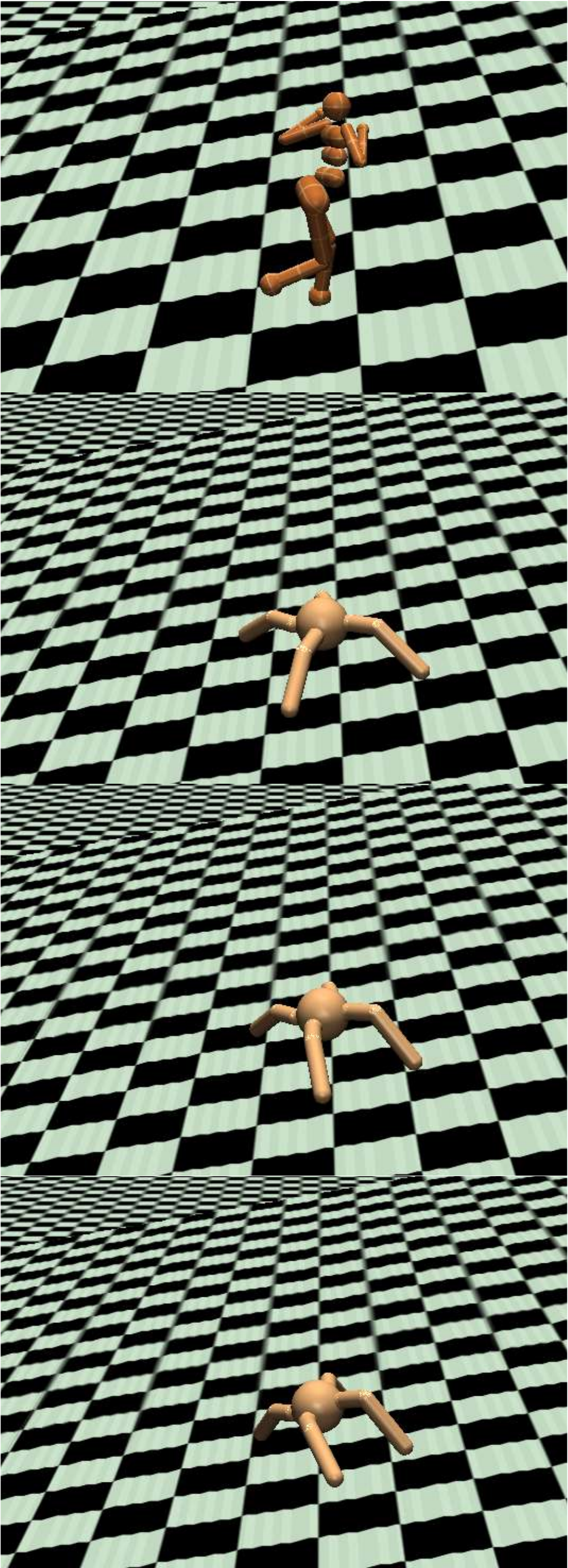}
\label{fig_experi_tsne_ha_curve_plane_common_example_2}
}
\subfigure[Agents nearest to (3)]
{
\includegraphics[width=0.37\columnwidth]{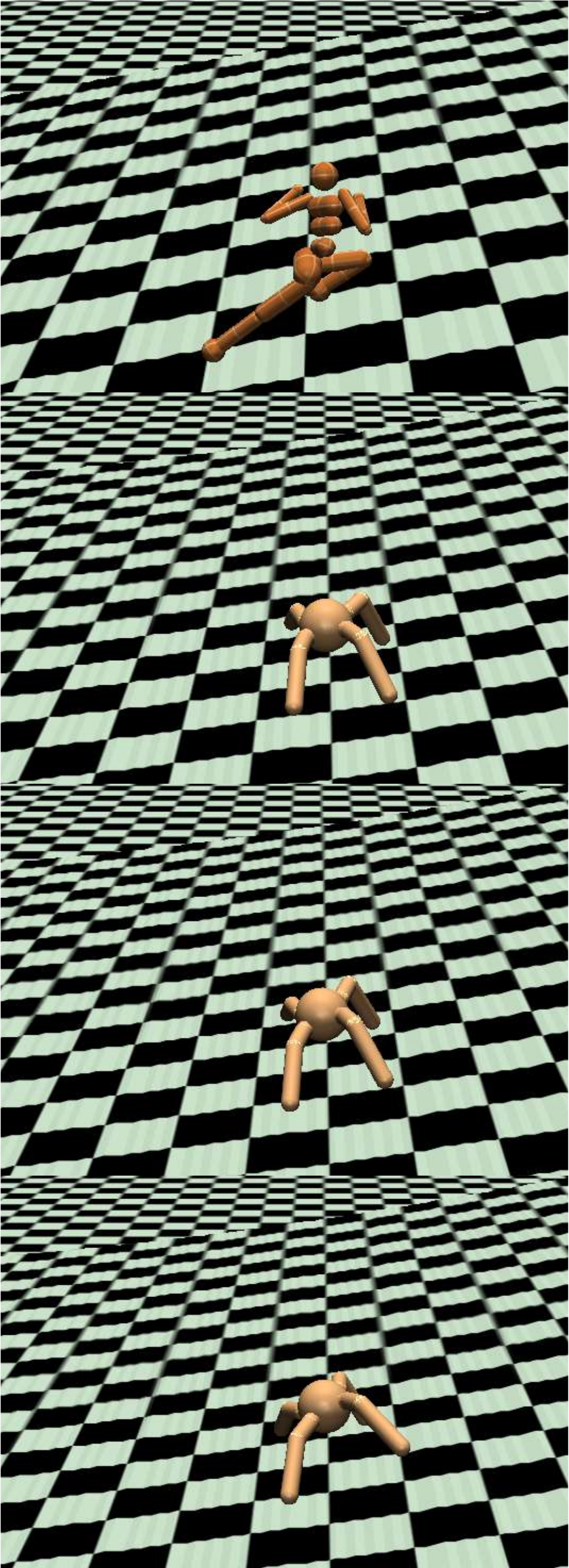}
\label{fig_experi_tsne_ha_curve_plane_common_example_3}
}
\subfigure[Agents nearest to (4)]
{
\includegraphics[width=0.37\columnwidth]{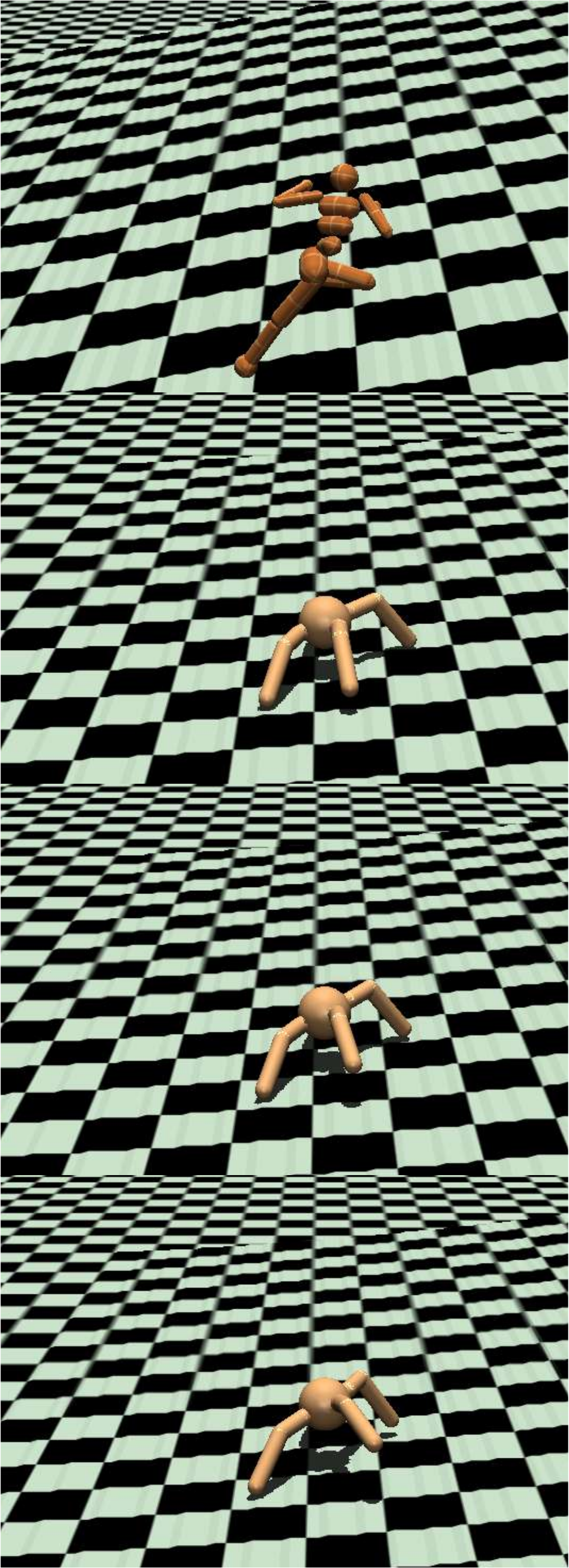}
\label{fig_experi_tsne_ha_curve_plane_common_example_4}
}
\subfigure[Agents nearest to (5)]
{
\includegraphics[width=0.37\columnwidth]{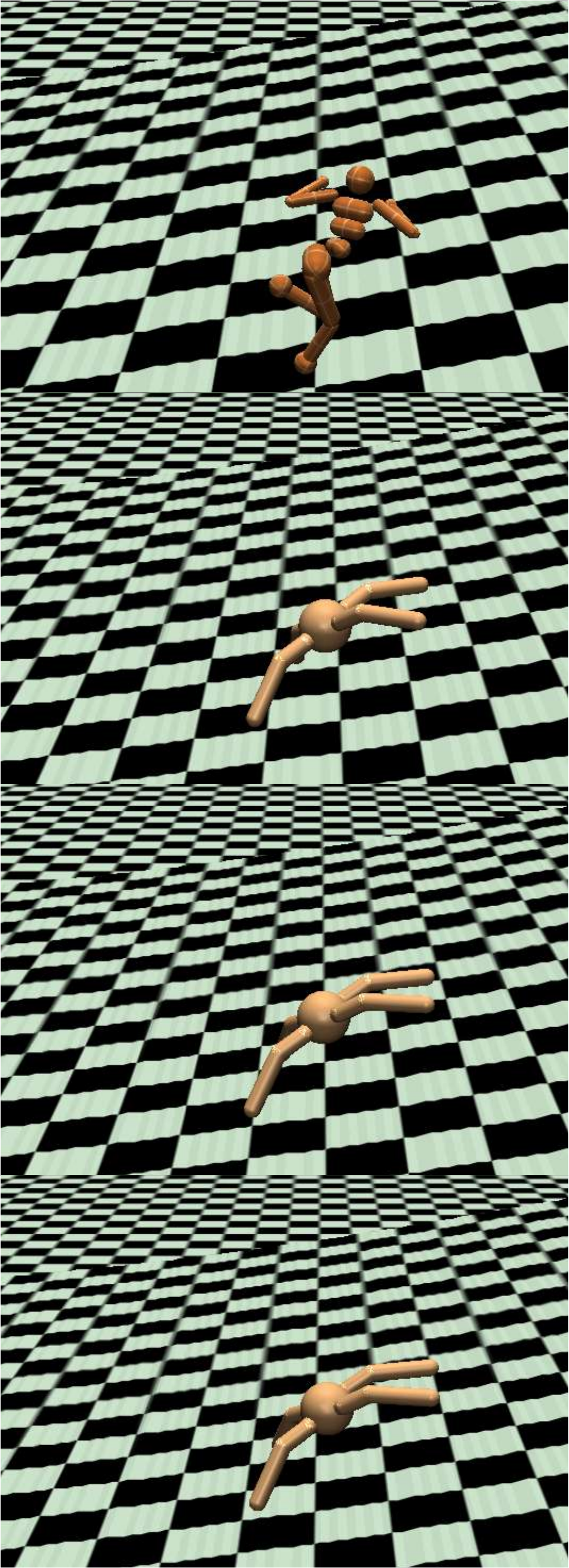}
\label{fig_experi_tsne_ha_curve_plane_common_example_5}
}
\caption{Analysis of t-SNE embeddings on the curved plane climbing tasks with Humanoid and Ant. (c): The timesteps are normalized to $\left [ 0,1 \right ]$, with different color mappings for each agent; (d) - (h): the top images show the source agent; the second images shows the target agent; the bottom two rows show two future frames following the second images.}
\label{fig_experi_tsne_ha_curve_plane}
\end{center}
\end{figure*}

\begin{figure}[t!]
\begin{center}
\includegraphics[width=0.85\columnwidth]{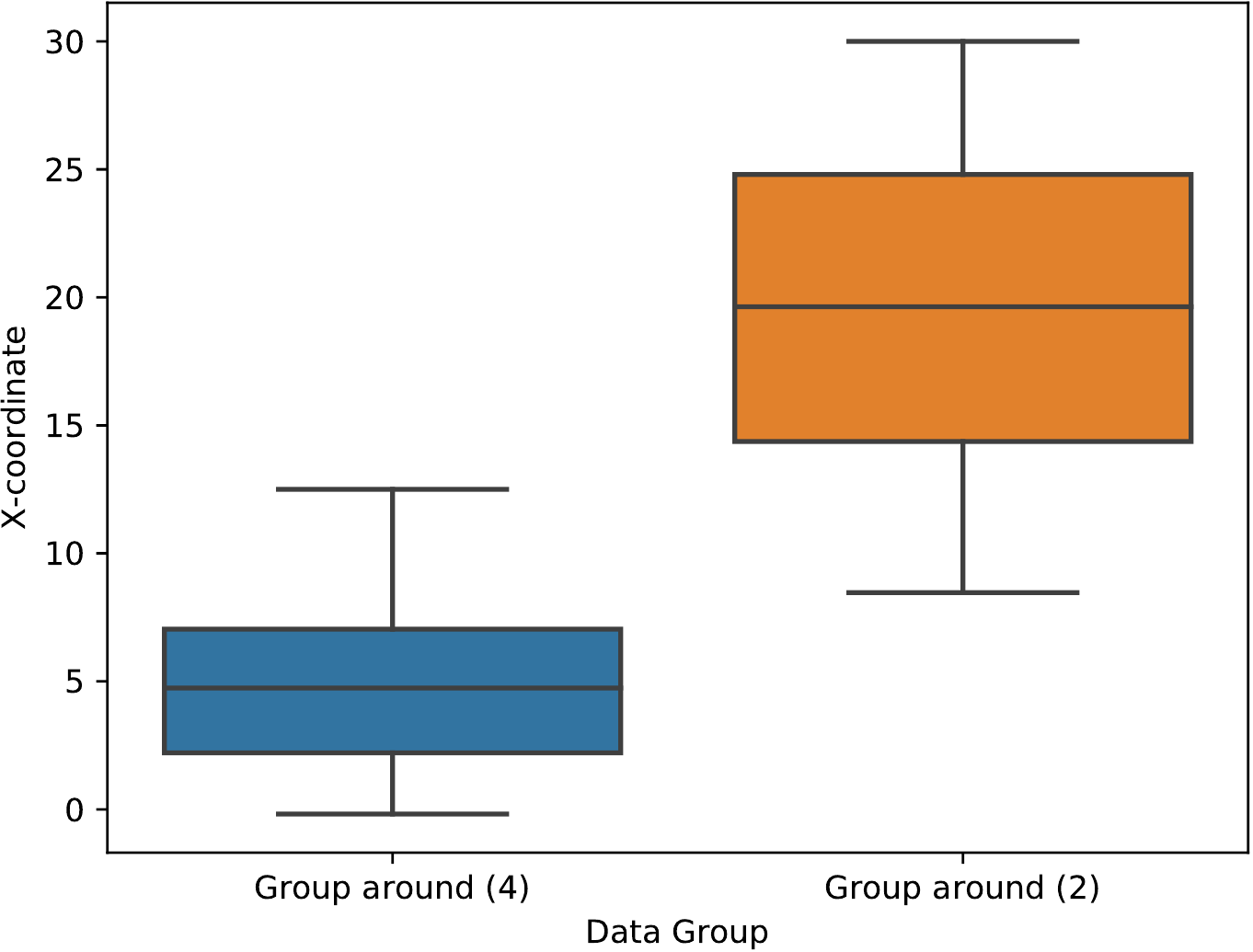}
\caption{Statistics for the grayscale groups around the markers (4) and (2) in Fig.~\ref{fig_experi_tsne_ha_curve_plane_common}}
\label{fig_experi_tsne_ha_curve_plane_common_statistics}
\end{center}
\end{figure}

\subsection{Interpretation of the learned subspaces}  

Our experimental results strongly support the benefits of using both shared and individual factors to achieve skill transfer. In this subsection we investigate whether the factors estimated by PVED can be interpreted in the context of our four environments. We resort to a non-linear dimensionality reduction approach based on t-SNE embeddings~\cite{tSNE_Maaten_2008} as a means to visualise the subspaces. Given a target task, we run multiple episodes using the learned policies of the source and target agents; then, for each subspace, we compute two-dimensional t-SNE embeddings for both agents. 

{\bf Reachers in button pushing tasks}. Fig.~\ref{fig_experi_tsne_button_pushing_common} shows the t-SNE embeddings for the shared subspace in this task. As expected, we observe a degree of overlap between the two agents' embeddings that is indicative of factors shared between agents. By color-coding the embeddings to represent the time domain, we are able to observe the presence of smooth patterns along the timesteps indicating a correspondence between the two agents across different stages of the task. To validate this assessment, we select four positions sufficiently spread out in the embedding, marked as (1), (2), (3) and (4) in Fig.~\ref{fig_experi_tsne_button_pushing_common}, and observe the source and target agents nearest to those marks; these are shown in Fig.~\ref{fig_experi_tsne_button_pushing_common_example_1} to Fig.~\ref{fig_experi_tsne_button_pushing_common_example_4}. By visual assessment, we conclude that these four locations roughly correspond to different stages in the control of button pushing: in position (1), which corresponds to an early timestep, the agents begin swinging towards the button location (Fig.~\ref{fig_experi_tsne_button_pushing_common_example_1}); then in (2) they adjust their poses and prepare to approach the button (Fig.~\ref{fig_experi_tsne_button_pushing_common_example_2}); finally, in (3) and (4), they  push the button (Fig.~\ref{fig_experi_tsne_button_pushing_common_example_3} and Fig.~\ref{fig_experi_tsne_button_pushing_common_example_4}). A certain similarity between the states of the two agents can also be observed at each selected location. 

On the other hand, the embeddings related to the individual subspace, shown in Fig.~\ref{fig_experi_tsne_button_pushing_specific}, indicate the presence of agent-dependent clusters. Furthermore, each cluster appears smooth along the timesteps still capturing the different sequential stages of the button pushing process. The fact that the agent-specific clusters are distinct and well separated provides some empirical validation to support the existence of individual latent factors.

{\bf Jaco3 and Fetch robot in peg insertion tasks}.  Fig.~\ref{fig_experi_tsne_peg_insertion_common} and Fig.~\ref{fig_experi_tsne_peg_insertion_specific} show the t-SNE embeddings for the shared subspace and the individual subspace, respectively. As before, we select 6 locations covering different areas in the shared embedding, marked as (1) to (6), as seen in Fig.~\ref{fig_experi_tsne_peg_insertion_common}. The source and target agents nearest to these marks are shown in  Fig.~\ref{fig_experi_tsne_peg_insertion_common_example_1} to Fig.~\ref{fig_experi_tsne_peg_insertion_common_example_6}. Similar observations to the Reachers button pushing tasks can be observed on the shared and individual subspaces. Note that, in Fig.~\ref{fig_experi_tsne_peg_insertion_common_example_1} to Fig.~\ref{fig_experi_tsne_peg_insertion_common_example_6}, the poses of robot arms are highly similar, but this is not the case for table locations. This is because we only consider robot-related states to infer the shared and individual subspaces in this task.

{\bf Walker2d and Hopper in curved-plane climbing tasks}.  Fig.~\ref{fig_experi_tsne_curve_plane_common} shows the t-SNE embeddings of the shared subspace in this task. We select five positions within this embedding, marked as (1) to (5) in Fig.~\ref{fig_experi_tsne_curve_plane_common}. Correspondingly, in Fig.~\ref{fig_experi_tsne_curve_plane_common_example_1} to Fig.~\ref{fig_experi_tsne_curve_plane_common_example_5}, we show the source and target agents nearest to those marks. These figures retain the same interpretability seen in the previous tasks. In Fig.~\ref{fig_experi_tsne_curve_plane_common_example_1} and Fig.~\ref{fig_experi_tsne_curve_plane_common_example_2}, the source agent leans forward to accelerate itself. Correspondingly, the target agent also leans forward; this pose also requires it to accelerate to keep the balance. In Fig.~\ref{fig_experi_tsne_curve_plane_common_example_3}, the source agent seems to be in moderate pace, with its torso (top part) leaning back to keep its center of mass back to preserve the balance; the target agent shows similar pose between torso and body with similar effect on its center of mass. In Fig.~\ref{fig_experi_tsne_curve_plane_common_example_4}, the source agent is taking off and floating with the reaction force generated from the contact between the brown leg and the ground, while the target agent is also jumping up which, similarly, requires the reaction force from the contact between the leg and the ground. Finally, in Fig.~\ref{fig_experi_tsne_curve_plane_common_example_5}, the source agent is in a transition phase before another round of force generation to move forward, and the target agent is about to land before another round of hopping forward. Based on these observations, there seems to be sufficient evidence to indicate that the shared subspace has captured the corresponding stages in the motion gait control between the two agents. The reason for observing smooth patterns along the timestep may be due to the gait adjustments required by the increasing tangential angles of the surface as the agents move forward, which is a common factor between them.

As in the previous tasks, in Fig.~\ref{fig_experi_tsne_curve_plane_specific}, we find distinct and well-separated clusters for each agent in the embeddings corresponding to the individual subspaces, which reflect the morphological heterogeneity. Again, this indicates that the model learns to encode morphology-specific motion control as individual factors. Moreover, there is a certain degree of overlap between early and late timesteps, possibly due to the occurrence of individual walking/hopping cycles. 

{\bf Humanoid and Ant in curved-plane climbing tasks}.
Fig.~\ref{fig_experi_tsne_ha_curve_plane_common} shows the t-SNE embeddings of the shared subspace in this task. We select five positions within this embedding, shown as marks (1) to (5) in Fig.~\ref{fig_experi_tsne_ha_curve_plane_common}. The source and target agents nearest to those marks are shown in Fig.~\ref{fig_experi_tsne_ha_curve_plane_common_example_1} to Fig.~\ref{fig_experi_tsne_ha_curve_plane_common_example_5}. Differently from the previous tasks, we show three frames for Ant: the first frame is the one nearest to the marks in Fig.~\ref{fig_experi_tsne_ha_curve_plane_common}, the other two frames are two future frames following the first frame. This is because we find the movement of Ant is not clear with a single frame. These figures show interesting correspondence between the two agents in the shared subspace. In Fig.~\ref{fig_experi_tsne_ha_curve_plane_common_example_1}, the Humanoid is about to stretch the shank of one leg forward, while the Ant is also stretching the shank of one leg forward. In Fig.~\ref{fig_experi_tsne_ha_curve_plane_common_example_2}, the Humanoid is moving the shank of one leg from the back of the body to the front of the body, and the Ant is moving one of its back legs forward. In Fig.~\ref{fig_experi_tsne_ha_curve_plane_common_example_3} and  Fig.~\ref{fig_experi_tsne_ha_curve_plane_common_example_4}, the Humanoid is about to move and land one of its feet, while the Ant is also moving and landing one of its feet. In Fig.~\ref{fig_experi_tsne_ha_curve_plane_common_example_5}, the overall pose of the Ant is similar to the pose of Humanoid, with two back legs of the Ant corresponding to the legs of the Humanoid and the two front legs of the ant corresponding to the arms of the Humanoid.

Interestingly, we can see that the shared embedding of the Ant can be roughly categorised into two groups around mark (2) and mark (4) in Fig.~\ref{fig_experi_tsne_ha_curve_plane_common}. We assume this is because the Ant learns separate strategies to move for the
curved plane before 12 meters and for the slope after 12 meters. This is validated by the summary shown in Fig.~\ref{fig_experi_tsne_ha_curve_plane_common_statistics}. On the other hand, we observe in Fig.~\ref{fig_experi_tsne_ha_curve_plane_common} that the group around (4) show smooth embeddings alone the timestep, and the group around (2) has heavy overlaps between early and late time steps. This may indicate that the Ant learns to adjust its pose for the increasing tangential angles of the curved plane before 12 meters, while it learns a fixed walking cycle for the slope after that. In contrast, the embedding of the Humanoid does not show any smooth pattern along the timestep. This may mean that the Humanoid is able to learn a universal strategy to handle different tangential angles.

In terms of the individual subspace in Fig.~\ref{fig_experi_tsne_ha_curve_plane_specific}, we see distinguished clusters indicating the difference in morphologies, and overlapping between early and later timesteps, similarly to the Walker2d and Hopper in curved-plane climbing tasks.

\section{Conclusion}

In this paper we have proposed a paired variational encoder-decoder model, PVED, specifically designed to  address the LMH problem. The model explicitly accounts for shared and individual hidden factors underpinning the control mechanisms of two MDAs. These latent factors are jointly modelled in order to better disentangle their distributions, which results in superior skill transfer capability in the context of DRL. We have theoretically indicated how the performance of the proposed skill transfer method depends on the agent morphologies and shared subspaces. We have reported on experimental studies showing the potential performance gains that can be achieved by PVED in four simulation environments. We have visualised and interpreted the subspaces learned by the model. PVED enables skill transfer under morphological heterogeneity. A possible direction for further improvement consists in learning feature representations for the agent morphologies to be incorporated into the model for improved subspace learning. Another possible direction is to divide the shared factors into more detailed categories, designing a model to identify and emphasize the categories that are more important for skill transfer to achieve better performance. 


\appendix
\section*{Proof of Theorem 1}
Note that $c$ is given by $q_{{\phi}_S}\left ( c_T|s_S,a_S \right )$ and $q_{{\phi}_T}\left ( c_S|s_T,a_T \right )$ under policies ${\pi}_S$ and ${\pi}_T$. In the rest of the proof we simplify the notation and use $q_S$ and $q_T$ instead of $q_{{\phi}_S}$ and $q_{{\phi}_T}$. By expanding the left hand side of Inequality~(\ref{eqn:13}), we have:
\begin{align}
&\left | \E_{c \sim {\pi}_{T}} R_T^{\prime} \left( c \right) - \E_{c \sim {\pi}_{S}} R_S^{\prime} \left( c \right)\right | \nonumber \\
= &\left | \sum_{c}{{q_T\left( c \right)}{R_T^{\prime} \left( c \right)}} - \sum_{c}{{q_S\left( c \right)}{R_S^{\prime} \left( c \right)}} \right | \nonumber \\
= &\left | \sum_{c,s_T,a_T}{{q_T\left( c,s_T,a_T\right)}{R_T^{\prime} \left( c \right)}} - \sum_{c,s_S,a_S}{{q_S\left( c,s_S,a_S \right)}{R_S^{\prime} \left( c \right)}} \right | \nonumber \\
= & \left| \sum_{c,s_T,a_T}{{q_T\left( c|s_T,a_T\right)}{q_T\left( s_T,a_T\right)}{R_T^{\prime} \left( c \right)}} \sum_{s_S,a_S}{q_S\left( s_S,a_S\right)} \right.\nonumber \\
& -  \left. \sum_{c,s_S,a_S}{{q_S\left( c|s_S,a_S\right)}{q_S\left( s_S,a_S\right)}{R_S^{\prime} \left( c \right)}} \sum_{s_T,a_T}{q_T\left( s_T,a_T\right)} \right | \nonumber \\
= & \left | \sum_{\substack{c,s_T,a_T\\s_S,a_S}}{{q_T\left( s_T,a_T\right)}{q_S\left( s_S,a_S\right)}} \left[ {q_T\left( c|s_T,a_T\right)}{R_T^{\prime} \left( c \right)} \right. \right.  \nonumber \\
&- \left. \left. {q_S\left( c|s_S,a_S\right)}{R_S^{\prime} \left( c \right)} \right] \vphantom{\sum_{\substack{c,s_T,a_T\\s_S,a_S}}} \right| \nonumber \\
\leqslant & \sum_{\substack{c,s_T,a_T\\s_S,a_S}}{{q_T\left( s_T,a_T\right)}{q_S\left( s_S,a_S\right)}} \left | {q_T\left( c|s_T,a_T\right)}{R_T^{\prime} \left( c \right)} \right.  \nonumber \\
&- \left. {q_S\left( c|s_S,a_S\right)}{R_S^{\prime} \left( c \right)}  \vphantom{q_T\left( s_T,a_T\right)} \right|
\label{eqn:15}
\end{align}
By using $R_T^{\prime} \left( c \right) = R_T^{\prime} \left( c \right) + R_S^{\prime} \left( c \right) - R_S^{\prime} \left( c \right)$ and $\left| R_T^{\prime} \left( c \right) - R_S^{\prime} \left( c \right) \right| \leqslant m$, we have
\begin{align}
& \left|{q_T\left( c|s_T,a_T\right)}{R_T^{\prime} \left( c \right)}  
- {q_S\left( c|s_S,a_S\right)}{R_S^{\prime} \left( c \right)}\right|\nonumber \\
=& \left|\left[ {q_T\left( c|s_T,a_T\right)} - {q_S\left( c|s_S,a_S\right)} \right] R_S^{\prime} \left( c \right)\right. \nonumber\\
&+\left. \left[ R_T^{\prime} \left( c \right) -  R_S^{\prime} \left( c \right) \right]{q_T\left( c|s_T,a_T\right)}\right|\nonumber \\
\leqslant& \left|\left[ {q_T\left( c|s_T,a_T\right)} - {q_S\left( c|s_S,a_S\right)} \right] R_S^{\prime} \left( c \right)\right| \nonumber\\
&+\left| \left[ R_T^{\prime} \left( c \right) -  R_S^{\prime} \left( c \right) \right]{q_T\left( c|s_T,a_T\right)}\right| \nonumber\\
\leqslant& \left|\left[{q_T\left( c|s_T,a_T\right)}-{q_S\left( c|s_S,a_S\right)}\right] {R_S^{\prime} \left( c \right)}\right| + m{q_T\left( c|s_T,a_T\right)}
\label{eqn:16}
\end{align}
By substituting Inequality~(\ref{eqn:16}) into Inequality~(\ref{eqn:15}), we have
\begin{align}
& \left | \E_{c \sim {\pi}_{T}} R_T^{\prime} \left( c \right) - \E_{c \sim {\pi}_{S}} R_S^{\prime} \left( c \right)\right | \nonumber \\ 
\leqslant & \sum_{\substack{c,s_T,a_T\\s_S,a_S}}{{q_T\left( s_T,a_T\right)}{q_S\left( s_S,a_S\right)}} \left | {q_T\left( c|s_T,a_T\right)}{R_T^{\prime} \left( c \right)} \right.  \nonumber \\
&- \left. {q_S\left( c|s_S,a_S\right)}{R_S^{\prime} \left( c \right)}  \vphantom{q_T\left( s_T,a_T\right)} \right|  \nonumber \\
\leqslant& m + \sum_{\substack{c,s_T,a_T\\s_S,a_S}}\left |{{q_T\left( s_T,a_T\right)}{q_S\left( s_S,a_S\right)}} \left[ {q_T\left( c|s_T,a_T\right)} \right. \right.  \nonumber \\
&- \left. { \left.{q_S\left( c|s_S,a_S\right)}\right]{R_S^{\prime} \left( c \right)}} \vphantom{q_T\left( c|s_T,a_T\right)} \right|
\label{eqn:17}
\end{align}
Let $q_{T,S} = {q_T\left( s_T,a_T\right)}{q_S\left( s_S,a_S\right)}$. Note that the following equation holds for any $k \in \mathbb{R}$:
\begin{align}
\sum_{\substack{c,s_T,a_T\\s_S,a_S}}{kq_{T,S}} {q_T\left( c|s_T,a_T\right)} = \sum_{\substack{c,s_T,a_T\\s_S,a_S}}{kq_{T,S}} {q_S\left( c|s_S,a_S\right)}
\label{eqn:18}
\end{align}
Thus, the second term in the last line of Inequality~(\ref{eqn:17}) can be rewritten:
\begin{align}
&\sum_{\substack{c,s_T,a_T\\s_S,a_S}} \left|{q_{T,S}} \left[  {q_T\left( c|s_T,a_T\right)} -  {q_S\left( c|s_S,a_S\right)} \right] {R_S^{\prime} \left( c \right)} \right| \nonumber \\
=& \sum_{\substack{c,s_T,a_T\\s_S,a_S}}\left|{q_{T,S}} \left[  {q_T\left( c|s_T,a_T\right)} -  {q_S\left( c|s_S,a_S\right)} \right] {\left[R_S^{\prime} \left( c \right) - k\right]} \right| \nonumber \\
\leqslant& \max_{c}\left|{R_S^{\prime} \left( c \right)}-k\right|\sum_{\substack{c,s_T,a_T\\s_S,a_S}}{ q_{T,S}\left| {q_T\left( c|s_T,a_T\right)} - {q_S\left( c|s_S,a_S\right)}\right| }
\label{eqn:19}
\end{align}
Note that:
\begin{align}
&\sum_{\substack{c,s_T,a_T\\s_S,a_S}}{q_{T,S}\left|{q_T\left( c|s_T,a_T\right)} - {q_S\left( c|s_S,a_S\right)}\right| }\nonumber \\
= & \sum_{\substack{s_T,a_T\\s_S,a_S}}q_{T,S}\sum_{c}\left| {q_T\left( c|s_T,a_T\right)} - {q_S\left( c|s_S,a_S\right)}\right|
\label{eqn:20}
\end{align}
According to Pinsker's inequality and using $D_{KL}\left (q_{S}\left ( c_T|s_S,a_S \right )||q_{T}\left ( c_S|s_T,a_T \right ) \right ) \leqslant \delta$:
\begin{align}
& \sum_{c} \left| {q_T\left( c|s_T,a_T\right)} - {q_S\left( c|s_S,a_S\right)} \right| \nonumber\\
\leqslant& \sqrt{2D_{KL}\left (q_{S}\left ( c|s_S,a_S \right )||q_{T}\left ( c|s_T,a_T \right ) \right )} \nonumber\\
\leqslant& \sqrt{2\delta}
\label{eqn:21}
\end{align}
Substituting Inequality~(\ref{eqn:21}) into Eq.~\ref{eqn:20}, we have
\begin{align}
\sum_{\substack{c,s_T,a_T\\s_S,a_S}}{q_{T,S}\left|{q_T\left( c|s_T,a_T\right)} - {q_S\left( c|s_S,a_S\right)}\right| } \leqslant  \sqrt{2\delta}
\label{eqn:22}
\end{align}
On the other hand, we note that the upper bound in Inequality~(\ref{eqn:19}) holds for all $k \in \mathbb{R}$. Recall that $R_S^{\prime} \left( c \right) \in \left[a, b\right]$. The upper bound is therefore minimized when $k = \left(a+b\right)/2$. In this case we have $\max_{c}\left|{R_S^{\prime} \left( c \right)}-k\right| \leqslant \left(b-a\right)/2$. By substituting Inequality~(\ref{eqn:22}) and $\max_{c}\left|{R_S^{\prime} \left( c \right)}-k\right| \leqslant \left(b-a\right)/2$ into  Inequality~(\ref{eqn:19}), we obtain:
\begin{align}
&\sum_{\substack{c,s_T,a_T\\s_S,a_S}}\left|{q_{T,S}\left[ {q_T\left( c|s_T,a_T\right)} - {q_S\left( c|s_S,a_S\right)}\right]{R_S^{\prime} \left( c \right)}}\right| \nonumber \\
\leqslant& \frac{\left(b-a\right)}{2} \sqrt{2\delta}
\label{eqn:23}
\end{align} 

Finally, with Inequality~(\ref{eqn:23}) and Inequality~(\ref{eqn:17}), we derive
\begin{align}
\left | \E_{c \sim {\pi}_{T}} R_T^{\prime} \left( c \right) - \E_{c \sim {\pi}_{S}} R_S^{\prime} \left( c \right)\right | \leqslant & m + \frac{\left(b-a\right)}{2} \sqrt{2\delta}
\label{eqn:24}
\end{align}
\section*{Proof of Corollary 1}
Starting with finite timesteps, we expand the left hand side of Inequality~(\ref{eqn:14}): 
\begin{align}
&\left | \E_{\tau \sim {\pi}_{T}} J_T \left( \tau \right) - \E_{\tau \sim {\pi}_{S}} J_S \left( \tau \right)\right | \nonumber \\
=&\left | \sum_{\tau}{q_{T}\left(\tau \right)} \sum_{t=0}^{n}{{\gamma}^t R_T^{\prime}\left(c_t\right)} -  \sum_{\tau}{q_{S}\left(\tau \right)} \sum_{t=0}^{n}{{\gamma}^t R_S^{\prime}\left(c_t\right)}\right |
\label{eqn:25}
\end{align}
Since $\sum_{\tau}{q\left(\tau \right)} \sum_{t=0}^{n}{{\gamma}^t R^{\prime}\left(c_t\right)} = \sum_{t=0}^{n} {\gamma}^t \sum_{\tau} {q\left(\tau \right)} {R^{\prime}\left(c_t\right)} = \sum_{t=0}^{n} {\gamma}^t \sum_{c_t}q\left(c_t\right) {R^{\prime}\left(c_t\right)} $, we can derive from Eq.~\ref{eqn:25}:
\begin{align}
&\left | \E_{\tau \sim {\pi}_{T}} J_T \left( \tau \right) - \E_{\tau \sim {\pi}_{S}} J_S \left( \tau \right)\right | \nonumber \\
=&\left | \sum_{t=0}^{n} {\gamma}^t \sum_{c_t}q_T\left(c_t\right) {R^{\prime}_T \left(c_t\right)} - \sum_{t=0}^{n} {\gamma}^t \sum_{c_t}q_S\left(c_t\right) {R^{\prime}_S \left(c_t\right)} \right |  \nonumber \\
=&  \left| \sum_{t=0}^{n}{{\gamma}^t \E_{c_t \sim {\pi}_{T}} {R^{\prime}_T \left(c_t\right)}} - \sum_{t=0}^{n}{{\gamma}^t \E_{c_t \sim {\pi}_{S}} {R^{\prime}_S \left(c_t\right)}} \right| \nonumber \\
=&  \left| \sum_{t=0}^{n}{{\gamma}^t} \left[ \E_{c_t \sim {\pi}_{T}} {R^{\prime}_T \left(c_t\right)} - \E_{c_t \sim {\pi}_{S}} {R^{\prime}_S \left(c_t\right)} \right] \right| \nonumber \\
\leqslant& \sum_{t=0}^{n}{{\gamma}^t \left|  \E_{c_t \sim {\pi}_{T}} {R^{\prime}_T \left(c_t\right)} - \E_{c_t \sim {\pi}_{S}} {R^{\prime}_S \left(c_t\right)} \right|} \nonumber \\
\leqslant& \left[m + \frac{\left(b-a\right)}{2} \sqrt{2\delta} \right] \sum_{t=0}^{n}{{\gamma}^t}
\label{eqn:26}
\end{align}
Theorem 1 is applied to derive the last step. As $t \rightarrow \infty$, we have
\begin{align}
&\left | \E_{\tau \sim {\pi}_{T}} J_T \left( \tau \right) - \E_{\tau \sim {\pi}_{S}} J_S \left( \tau \right)\right | \leqslant \left(\frac{1}{1-\gamma}\right) \left[m + \frac{\left(b-a\right)}{2} \sqrt{2\delta} \right]
\label{eqn:27}
\end{align}
%





\ifCLASSOPTIONcaptionsoff
  \newpage
\fi



\bibliographystyle{IEEEtran}
\bibliography{bare_jrnl_arxiv}
\end{document}